\documentclass[11pt]{article}
\usepackage{amsmath}
\usepackage{graphicx}
\usepackage{enumerate}
\usepackage{natbib}
\usepackage{url} 
\usepackage{hyperref}
\hypersetup{
	colorlinks=true,
	linkcolor=blue,
	citecolor=blue,
}

\newcommand{\blind}{0}

\usepackage{amsmath,bm}
\usepackage{amssymb}
\usepackage{amsfonts}
\usepackage{multirow}
\usepackage{amsthm}
\usepackage{algorithm,algorithmic}

\usepackage{caption}
\usepackage{subcaption}

\newcommand{\pr}[1]{\left(#1\right)}

\def\E{{\mathbb E}}
\def\P{{\mathbb P}}
\def\x{{\bf x}}
\def\D{{\mathcal D}}
\def\Z{{\mathbb Z}}

\def\N{{\mathcal N}}
\def\CN{{\mathcal C\mathcal N}}

\def\I{{\mathbb{I}}}
\def\R{{\mathbb{R}}}

\newtheorem{theorem}{Theorem}
\newtheorem{lemma}{Lemma}

\newtheorem{proposition}{Proposition}
\theoremstyle{definition}
\newtheorem{definition}{Definition}

\newtheorem{remark}{Remark}
\newtheorem{assump}{Assumption}

\DeclareMathOperator*{\argmin}{arg\,min}

\allowdisplaybreaks

\begin{document}

\def\spacingset#1{\renewcommand{\baselinestretch}%
{#1}\small\normalsize} \spacingset{1}


{
  \title{\bf  Consistency for Large Neural Networks: Regression and Classification}
  \author{Haoran Zhan\thanks{
    haoran.zhan@u.nus.edu}\hspace{.2cm}\\
    Department of Data Science and Statistics, \\
    National University of Singapore\\
    and \\
    Yingcun  Xia\\
    Department of Data Science and Statistics,\\
     National University of Singapore
    }
    \date{}
  \maketitle
}

\if\blind
{
  \bigskip
  \bigskip
  \bigskip
  \begin{center}
    {\LARGE\bf Non-asymptotic Properties of Generalized Mondrian
Forests in Statistical Learning}
\end{center}
  \medskip
} \fi

\bigskip
\begin{abstract}
Although overparameterized models have achieved remarkable practical success, their theoretical properties—particularly their generalization behavior—remain incompletely understood. The well known double descents phenomenon suggests that the test error curve of neural networks decreases monotonically as model size grows and eventually converges to a non-zero constant. This work aims to explain the theoretical mechanism underlying this tail behavior and study the statistical consistency of  deep overparameterized neural networks in many different learning tasks including regression and classification. Firstly, we prove that as the number of parameters increases, the approximation error decreases monotonically, while explicit or implicit regularization (e.g., weight decay) keeps the generalization error existing but bounded. Consequently, the overall error curve eventually converges to a constant determined by the bounded generalization error and the optimization error. Secondly, we prove that deep overparameterized neural networks are statistical consistency across multiple learning tasks if regularization technique is used. Our theoretical findings coincide with  numerical experiments and provide a perspective for understanding the generalization behavior of overparameterized neural networks.

\end{abstract}

\noindent%

\spacingset{1.0} 
\section{Introduction}
\label{sec:intro}

The field of machine learning has experienced a significant surge in the development and application of overparameterized neural networks, particularly in deep learning; see, for example, \cite{vaswani2017attention} and \cite{goodfellow2020generative}. These models, which have comparable parameters with training examples, have become central to modern machine learning; see Table \ref{TTab1}.
\begin{table}[H]
\centering
\caption{Comparison of Key Data of GPT Family Models}\label{TTab1}
\begin{tabular}{|c|c|c|c|}
\hline
Model & Release Time & Parameter Count & Training Data Volume \\
\hline
GPT-1 & June 2018 & 117 million & About 5GB \\
\hline
GPT-2 & February 2019 & 1.5 billion & 40GB \\
\hline
GPT-3 & May 2020 & 175 billion & 17GB \\
\hline
GPT-4 & March 2023 & 1.8 trillions & 45GB \\
\hline
\end{tabular}
\end{table}
\noindent Despite their widespread use and impressive success in practice, understanding the theoretical properties of overparameterized networks remains an active area of research. In this paper, we focus on  their statistical consistency.

One of the key questions in the study of overparameterized networks is whether they can achieve good predictions and generalize effectively. Traditional learning theory suggests that overparameterization could lead to  either  poor generalization or good generalization. Let $\N\N_k$ be the deep network class with $k$ parameters and consider the least squares regression below
$$
   \mathcal{S}:=\argmin_{f\in \N\N_k} \frac{1}{n}\sum_{i=1}^{n}(Y_i-f(X_i))^2.
$$
When $k>n$, it is known that $\mathcal{S}$ has many different networks. On the one hand, \cite{lin2025generalization} proved that some network  in  $\mathcal{S}$  does not converge to the true regression function even if each $Y_i$ contains no noise;  On the other hand, \cite{lin2025generalization} shows that some network  in  $\mathcal{S}$  behaves very well and  achieve the best consistency rate. Unfortunately, we do not how to characterize these two types of networks in  $\mathcal{S}$. To avoid the overfitting problem, \cite{neyshabur2017exploring} summarized several approaches to measure the generalization error of overparameterized networks, with one of the effective methods being regularization techniques. Additionally, \cite{soudry2018implicit} demonstrated through numerical studies that gradient descent in overparameterized networks tends to converge to minima with low training and generalization error, suggesting that implicit regularization plays a significant role. Extensive numerical experiments by \cite{zhang2021understanding} and \cite{arora2018optimization} have further shown that even shallow overparameterized networks (with just two layers) do not necessarily overfit in the traditional sense and perform exceptionally well in image classification tasks, often exhibiting implicit regularization properties leading to good generalization. Thus, despite their large model size, these networks tend to avoid overfitting due to the effects of implicit regularization. It is worth noting that even when explicit regularization is applied during training, the use of a finite number of iterations before convergence (early stopping) acts as an implicit regularization, contributing to the success of the training process; see, for example, \cite{Prechelt1998}.

Although regularization is essential for the performance of both large neural networks and traditional statistical models, the mathematical details involved differ significantly. Traditionally, the number of neurons has been used to measure the generalization error of neural networks (e.g., see \cite{SchmidtHieber2020} and \cite{kohler2021rate}). However, when applied to large networks, this method results in overly large generalization error bounds, making the traditional approach is not suitable for studying the consistency of large networks.

On the other hand, the consistency of large neural networks relates closely to another research topic, the phenomenon of double descent appearing in the error curve of neural networks. In detail, this curve has two descent times rather than one global minimal point in the traditional U curve. Many papers have studied this phenomenon and contributed its appearance to  different reasons encompassing both theoretical and computational aspects; see \cite{belkin2019reconciling}. For instance, \cite{hastie2022surprises} argued that the occurrence of double descent is closely linked to variance reduction within the framework of linear regression. In such cases, the error curve can become a monotone decreasing function if the penalty is properly chosen. Additionally, \cite{schaeffer2023double} highlighted that this phenomenon only arises when certain mathematical relationships between the training and testing data are satisfied. Moreover, \cite{curth2024u} suggested that in many machine learning problems, double descent is a direct consequence of transitioning between two distinct mechanisms for increasing the total number of model parameters along two independent axes.

\begin{figure}[h!]
  \centering
  \includegraphics[width=1.0\linewidth]{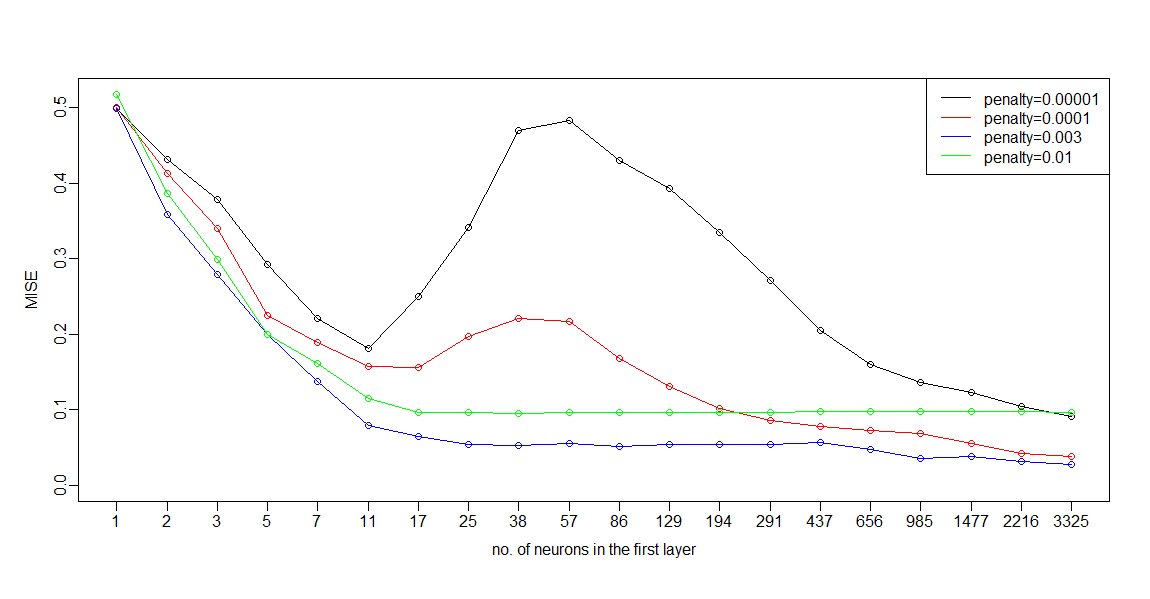}
  \caption{Simulation results for data generated from model $ y = \sqrt{\x_1^2 + \x_2^2} + \cos(\pi(X_3+X_{4})) + 0.2\varepsilon $, where $ X = (\x_1, ..., \x_d)  \sim unif( \sqrt{2}\mathbb{S}^{d} ) $, where $ unif( \sqrt{2}\mathbb{S}^{d}) $ is uniform distribution on the $d$-dimensional sphere with  radius $\sqrt{2}$, and  $ \varepsilon \sim N(0, 1) $ are independent, with $ d = 32 $ and sample size $ n = 1024$, using R package \texttt{nnet} or \texttt{keras} respectively. For both packages, we choose the number of iterations, i.e. maxit and echo, big enough to avoid the additional regularization due to early stopping. The X-axis is the number of neurons, $K$, following a geometric sequence with a common ratio of 1.5,  in the hidden layer(s) of $NN(K, max(2, [K/2]), 1)$, and Y-axis is the generalization MISE with different regularization error.}\label{fig:2}
\end{figure}

Here, we give an least squares regression example in Figure \ref{fig:2}, the MISE curve (shown in blue) clearly exhibits this double descent. However, different levels of regularization can result in double descent or other possible patterns. It can be observed  in Figure \ref{fig:2} that the error curve  is divided into two parts, namely the first part for small neural networks and the second part for large neural networks. The curve of the  first part can be the traditional U type which is known to be the bias and variance trade-off, or the decreasing type which is affected by the regularized technique; see the discussion in \cite{shiyan2023}.  As suggested above, it is difficult to fully understand the double descent phenomenon. In this paper, we are interesting the tail part of curves in Figure \ref{fig:2}. In fact, it is natural  to ask the question below.

 \begin{center}
  \textit{ How to understand the tail part of curve (namely, error for large neural networks) is always decreasing and converges to some non-zero number ?}
 \end{center}

In this paper, we provide an answer to this question across a range of learning scenarios, including least squares regression, robust regression, and multi-class classification. Our findings indicate that the strength of regularization plays a crucial role in each of these settings. It is well established that the total learning error comprises three components: optimization error, approximation error, and generalization error. As the size of the deep network grows, the approximation error tends to decrease. In contrast, the generalization error does not grow unbounded; rather, when sufficient regularization is applied, it remains bounded above even for large network size. Consequently, as illustrated in Figure \ref{fig:2}, the overall error curve declines but eventually converges to a non-zero constant, which reflects the upper bound imposed by the generalization error.

In summary, the contributions of this paper are twofolds:

\begin{itemize}
    \item  Firstly, we propose establish the statistical consistency of deep or overparameterized neural networks in many different learning tasks including least squares regression, robust regression and classification.

    \item  Secondly, in each above learning task we explain the tail testing error curve converges to a non-zero constant as the size  of deep networks goes  to infinity.

\end{itemize}

\subsection{Related work}

Traditionally, many studies have investigated the statistical risk of least squares estimates under the framework of small size neural networks (e.g., Bauer and Kohler (2019), Schmidt-Hieber (2020), Kohler and Langer (2021), among others). However, these works largely overlook neural networks that are over-parameterized, where the number of parameters significantly exceeds the sample size. This over-parameterization presents unique challenges and properties that are not addressed in their analyses.

\cite{drews2022universalconsistencyoverparametrizeddeep} is  an early paper that studied the statistical consistency of over-parameterized networks. However, the key Lemma 3 they used to bound the generalization error is wrong because they missed the dimension of Taylor polynomials in the exponent of this bound. When this dimension goes to infinity, the upper bound of covering number in their paper will also diverge to infinity. Therefore, the trick in \cite{drews2022universalconsistencyoverparametrizeddeep} fails to work in large neural networks. \cite{wang2021nonasymptotic}  studied overparameterized shallow neural networks with ReLU activation. Specifically, \cite{wang2021nonasymptotic} converts this problem into a group lasso problem. By leveraging techniques from lasso regression, they obtain non-asymptotic results for shallow neural networks with ReLU activation. While this transformation is technically interesting, it limits their study to a specific type of network. This limitation arises because establishing such equivalence becomes challenging when the activation function is not piecewise linear or when networks have more than one hidden layer. \cite{yang2025optimal} established the optimal rates of approximation by shallow ReLu$^k$ neural networks and also gave the consistency rate of large networks  by using this tool. Later on, they improve their proof technique and gave the optimal consistency rate of shallow large neural networks in \cite{yang2024nonparametric}. 

This paper is a following work of  above papers. The main difference is that above papers only considered large neural networks with one hidden layer and only least squares loss was studied. However, it is known that deep learning is powerful largely due to the introduction of the depth. Our goal is to study this problem by using deep learning and consider other commonly used losses such as Huber loss, quantile loss and cross-entropy in classification.  

On the other hand, regularization, whether explicitly through penalty imposition or implicitly through early stopping of training algorithms (e.g. \cite{Yao2007OnES} and \cite{rice2020overfittingadversariallyrobustdeep}), is crucial to controlling the generalization error of neural networks. In this work, we choose the penalty suggested by  \cite{golowich2018size} and  \cite{jiao2023approximation} which can reflect the complexity of deep neural networks. It is interesting to see that this penalty is equivalent to those used in \cite{wang2021nonasymptotic}, \cite{yang2025optimal} and \cite{yang2024nonparametric} when the depth is two. In fact, as argued in \cite{wang2021nonasymptotic}, this penalty is equivalent to $L_2$ penalty for shallow networks. \cite{Goodfellow-et-al-2016} emphasized that $L_2$ regularization (also known as weight decay) is often more effective and widely used in deep learning compared to $L_1$. The smoothness and stability provided by $L_2$ regularization are key reasons for its widespread use.
Interestingly, in Keras training, $L_2$ regularization is set as default. Thus, our work is a generalization of previous works for shallow large networks and is also more relevant to the practice.

\subsection{Notations}
We use $c,c_1,c_2,\ldots$ to denote some positive constants in this paper and  the constant $c>0$ can also vary from line to line. Sometimes, $c(\mathcal{O})$ is also used to denote a positive constant that relies on the object $\mathcal{O}$ only. On the other hand, $a\lesssim b$ denotes there is a universal constant $c>0$ such that $a\le cb$ and $a\gtrsim b$ is defined in  a similar way and $a\asymp b$ means both $a\lesssim b$ and $b\lesssim a$ are satisfied.

\section{Large neural networks for least squares regression}\label{sec:2}
Our  first interest is to estimate the conditional expectation $m(\bm{x}):=\E(Y|\bm{X}=\bm{x}), \bm{x} \in [0,1]^d$ by using an i.i.d. sample $\D_n := \{(\bm{X}_i,Y_i)\}_{i=1}^n$. It is known that there are already many nonparametric  methods, such as kernel smoothing, spline and wavelet.  In this paper, we study large (deep) neural network in nonparametric  regression which is a popular topic and less studied in literature.

In deep learning, we use ReLu activation in our theoretical analysis due to the well known gradient explosion/vanishing phenomenon in the application of backprogation algorithm. For example, see \cite{he2015delving} about the discussion of this problem.   In this case, the neural network with depth $L\in\mathbb{Z}^+$ has the structure
\begin{align}
  g_0(\bm{x}) &:=  \bm{x}, \bm{x}\in [0,1]^d,\nonumber\\
  g_{\ell+1}(\bm{x}) &:= \sigma_{relu}(\bm{A}^\ell g_\ell(\bm{x})+\bm{v}^\ell), \quad \ell=0,1,\ldots, L-1,\nonumber\\
  g(\bm{x})&:=\bm{A}^L g_L(\bm{x}),\label{multinetwork}
\end{align}
where $\bm{A}^\ell\in \mathbb{R}^{N_{\ell+1}\times N_\ell},\bm{v}^\ell\in\mathbb{R}^{N_{\ell+1}}$ with $N_0=d,N_{L+1}=1$ and $ \sigma_{relu}((\bm{x}_1,\ldots,\bm{x}_j)^T):=(\sigma_{relu}(\bm{x}_1),\ldots, \sigma_{relu}(\bm{x}_j))^T$  is defined in element-wise for
any $(\bm{x}_1,\ldots,\bm{x}_j)\in\mathbb{R}^p$ and $j\in\mathbb{Z}^+$.  Meanwhile, $W=\max\{N_2,\ldots,N_{L+1}\}$ is called the network width.  In conclusion, the feed-forward neural network class is given by
\begin{equation}\label{;hkajdk1}
  \N\N_{d, N_L}(W_k,L_k):= \left\{g\ \text{has\ form\ in\ } \eqref{multinetwork} \text{\ with\ width\ } W_k\ \text{and\ depth\ } L_k\right\}
\end{equation}
When $N_L=1$, we also write $ \N\N_{d, N_L}(W_k,L_k)$ as $ \N\N(W_k,L_k)$ in this paper.
 
The consistency of large neural networks relies heavily on the sample error which is equivalent to the analysis of  Gaussian or Rademacher complexity.  The Gaussian/Rademacher complexity for  large neural networks has already been studied in many papers; see e.g.  \cite{neyshabur2015norm}, \cite{gao2016dropout}, \cite{neyshabur2017exploring}, \cite{golowich2020size} and \cite{jiao2023approximation}. An interesting  finding in these literature is that the upper bound of  Gaussian complexity can depends less on both $W_k$ and $L_k$ under certain network norms, which makes it possible to bound the sample error of large neural networks. For any $g\in   \N\N(W_k,L_k)$, we use a popular path norm suggested by both \cite{golowich2018size} and  \cite{jiao2023approximation}
\begin{equation}\label{pingan}
  J(g):= \|(A_{L_k},v^{L_k})\|_1 \|(A_{L_k-1},v^{L_k-1})\|_1\cdots  \|(A_{1},v^{1})\|_1,
\end{equation}
where $\|\cdot\|_1$ denotes the maximum 1-norm of the rows of any matrix. Namely, for any matrix $A=\{a_{i,j},i\in [m],j\in [n]\}$, $\|A\|_1:=\max_{i\in [m]}\sum_{j=1}^{n}|a_{i,j}|$.  Compared with other norms,  an advantage of this network norm is shown below.

\begin{definition}\label{Gaussian_complexity}
   For any fixed points $\{\bm{x}_i\}_{i=1}^n\subseteq\R^d$, define the Gaussian complexity of $\N_k$ by
  $$
  \mathcal{G}(  \N\N(W_k,L_k);\{\bm{x}_i\}_{i=1}^n):=\E_{s_i}\left( \frac{1}{n}\sup_{g\in  \N\N(W_k,L_k)}{\sum_{i=1}^{n}s_i\cdot g(\bm{x}_i)}\right),
  $$
  where $(s_1,\ldots,s_n)$ are independent and each follows standard Gaussian distribution.
\end{definition}

\begin{proposition}[Theorem 3.2 in \cite{golowich2018size}]\label{pro_relu_multilayer}
  The Gaussian complexity of $  \N\N(W_k,L_k,U)$  satisfies
  $$
  \sup_{\substack{\bm{x}_i\in [0,1]^d,i=1,\ldots,n}}\mathcal{G}(   \N\N(W_k,L_k,U);\{\bm{x}_i\}_{i=1}^n)\le c(d)\cdot {M}\sqrt{\frac{L_k}{n}},
  $$
  where $  \N\N(W_k,L_k,U):=\{g\in   \N\N(W_k,L_k): J(g)\le U\}$ for any $U>0$.
\end{proposition}
This proposition tells us  the corresponding Gaussian complexity does not depend on $W_k$ and  relies less on the depth $L_k$. This property matches with  the current applied large language networks that do not have deep depth compared with their training data sizes.  Meanwhile, for shallow network $L=2$, \cite{wang2021nonasymptotic} proved that the path norm in \eqref{pingan} is equivalent to the $L^2$ norm. Importantly, norm \eqref{pingan} is also used in their paper although  the case for shallow networks was studied only.

Then, the regularized large network estimator is given by
\begin{equation}\label{Jpenalty_multi}
\hat{m}_n:=\arg min_{g\in  \N\N(W_k,L_k)}{\frac{1}{n}\sum_{i=1}^n{(Y_i-g(\bm{X}_i))^2}+\lambda_n J(g)},
\end{equation}
where $\lambda_n>0$ is a predefined  penalty strength. 
To analyze the statistical consistency of above estimator,  we introduce  two types of error, namely,  the empirical error
$$
  \|\hat{m}-m\|_n^2:= \frac{1}{n} \sum_{i=1}^{n} (\hat{m}(\bm{X}_i)-m(\bm{X}_i))^2
$$
and the prediction error
$$
  \|\hat{m}-m\|_2^2:= \E_{\bm{X}}(\hat{m}(\bm{X})-m(\bm{X}))^2.
$$
Usually, the MSE (mean  squares error ) of $\hat{m}(\bm{x})$ consists of two parts, namely the approximation error and the generalization error. It is well known that the approximation error decreases monotonically and the generalization error often increases monotonically as the size of network goes to infinity. Therefore, the left problem is to find a way to bound its generalization error (variance term). Traditionally, this error is usually bounded by using the VC dimension of $\N_k$ (this dimension is roughly equal to $k$); see \cite{kohler2021rate} and \cite{bartlett2019nearly}. However, this traditional method does not apply for the case of large neural networks since a large $k$ can only lead to a divergent bound of its generalization error. We will use Proposition \ref{pro_relu_multilayer} to solve this problem.

Similar to \cite{SchmidtHieber2020}, we suppose the true regression function is in the hierarchical composition model below. 

\begin{definition}[H\"{o}lder space]
  For any $\alpha>0$, let $\alpha=r+\beta$ with $\beta\in (0,1]$. Denote by $H^{\alpha}(\R^d)$  the H\"{o}lder space with the norm
   \begin{equation}\label{Holder}
         \|f\|_{H^{\alpha}(\R^d)}:=\max \left\{     \|f\|_{C^{r}(\R^d)}, \max_{\|s\|_1=r}|\partial^s f|_{C^{0,\beta}(\R^d)}\right\},
   \end{equation}
   where $s=(s_1,\ldots,s_d)\in (\mathbb{Z}^+)^{\oplus d}$ is a multi-index and
   $$
   \|f\|_{C^r(\R^d)}:= \max_{\|s\|_1\le r}\|\partial^s f\|_{L^\infty(\R^d)}, \quad |f|_{C^{0,\beta}(\R^d)}:= \sup_{\bm{x}\neq y}\frac{|f(\bm{x})-f(y)|}{\|\bm{x}-y\|_2^\beta}
   $$
   and $\|\cdot\|_{L^\infty}$ is the supremum norm.
\end{definition}

\begin{definition}[Hierarchical composition model]
 Given positive integers $d, l \in \mathbb{N}^{+}$and a subset of $[1, \infty) \times  (0,\infty)\times \mathbb{N}^{+}$, denoted by $\mathcal{P}$, satisfying $\sup _{(\alpha, C, t) \in \mathcal{P}} \max \{\alpha, C, t\}<\infty$, the hierarchical composition model $\mathcal{H}(d, l, \mathcal{P})$ is defined recursively as follows. For $l=1$,
$$
\begin{array}{r}
\mathcal{H}(d, 1, \mathcal{P})=\left\{h: \mathbb{R}^{d} \rightarrow \mathbb{R}: h(\bm{x})=g\left(\bm{x}_{\pi(1)}, \ldots, \bm{x}_{\pi(t)}\right), \text { where } \pi:[t] \rightarrow[d]\right. \text { and } \\
\left.g: \mathbb{R}^{t} \rightarrow \mathbb{R} \text { is in } C\cdot H^\alpha([0,1]^d) \text{ for some } C>0\right\}
\end{array}
$$
and for $l>1$,
$$
\begin{gathered}
\mathcal{H}(d, l, \mathcal{P})=\left\{h: \mathbb{R}^{d} \rightarrow \mathbb{R}: h(\bm{x})=g\left(f_{1}(\bm{x}), \ldots, f_{t}(\bm{x})\right), \text { where } f_{i} \in \mathcal{H}(d, l-1, \mathcal{P})\right. \text { and } \\
\left.g: \mathbb{R}^{t} \rightarrow \mathbb{R} \text { is in } C\cdot H^\alpha(\R^d) \text{ for some } C>0 \right\}
\end{gathered}
$$
\end{definition}

Finally,  assumptions on the distributions of $\bm{X}$ and $Y$ and the corresponding relationship   are also necessary.
\begin{enumerate}
  \item[(C4).] The sample $\{(\bm{X}_i,Y_i)\}_{i=1}^n$ is drawn independently from the population $(\bm{X},Y)$.
   \item[(C5).]  The residual $\varepsilon=Y-\E(Y|\bm{X})\sim N(0,\sigma^2)$ is independent to $\bm{X}$.
\end{enumerate}

The first result is the empirical error bound of the regularized network estimator $\hat{m}(\bm{x})$. By choosing proper $\lambda_n$, we successfully use Proposition \ref{pro_relu_multilayer} to bound its generalization error. The detail of proof is deferred to Section \ref{sec_proof}.

\begin{theorem}[Empirical error of  large neural networks]\label{Theorem1}
  Under conditions (C1-5) and suppose $m\in \mathcal{H}(d, l, \mathcal{P})$, the regularized network estimator $\hat{m}(\bm{x})$ with $\lambda_n=c\sqrt{\frac{L_k\ln^2 n}{n}}$ satisfies
  \begin{equation}\label{ahbjsdhj1}
       \|\hat{m}-m\|_n^2\lesssim  \max\left\{(L_kW_k)^{-2\alpha_1}, (n/L_k)^{-\frac{\beta_1}{2\beta_1+1} }\right\}
  \end{equation}
  with probability at least $1-O(n^{-r})$, where  $r>0$ is a large number and   $\alpha_1= \min_{(\alpha, C, t) \in \mathcal{P}}\left\{ \frac{2\alpha}{t}\right\}$ and $\beta_1:= \min_{(\alpha, C, t) \in \mathcal{P}}\left\{ \frac{\alpha}{t+1}\right\}/l$ and $W_k\gtrsim n^{c(\mathcal{P})}$. Furthermore, the  upper bound in \eqref{ahbjsdhj1} also holds for $  \E\left( \|\hat{m}-m\|_n^2\right)$.
\end{theorem}

Similarly, we  also establish the upper bound about prediction error.

\begin{theorem}[Prediction error of large neural networks]\label{Theorem2}
     Under conditions (C1-5), the regularized network estimator $\hat{m}(\bm{x})$ with $\lambda_n=c\sqrt{\frac{L_k \ln^2 n}{n}}$ satisfies
  \begin{equation}\label{ahbjsdhj1}
      \|\hat{m}-m\|_2^2 \lesssim  \max\left\{(L_kW_k)^{-2\alpha_1}, (n/L_k)^{-\frac{\beta_1}{2\beta_1+1)} }\right\}.
  \end{equation}
  with probability at least $1-O(n^{-r})$, where  $r>0$ is a large number and $W_k\gtrsim n^{c(\mathcal{P})}$.
  
\end{theorem}

Theorem \ref{Theorem1} \& \ref{Theorem2} show that large neural networks are always statistically consistent. For any general regression function, we can guarantee the consistency of large neural networks  even if $k=O(e^n)$. This result is interesting because the size of neural network has no influence on its statistical consistency. Theoretically, we can design any large neural networks in practice without being afraid of  its overfitting problem. This result is different from previous asymptotic results for small  ($k=o(n)$) or sparse neural networks only, such as \cite{SchmidtHieber2020} and \cite{kohler2021rate}.

Secondly, it is no need to increase the size of neural networks if one aims to reduce the  prediction error. According to Theorem \ref{Theorem1}, the error will not reduce anymore if $k$ increases to a large threshold. This result coincides with our simulation result; see Figure \ref{fig:2}. Therefore, our result suggests that large neural networks are useful but we can not make a fetich of them and design very large networks without rational consideration.

\subsection{Connection to random forests}
The random forest (RF) proposed by \cite{breiman2001random} is a popular and powerful nonparametric regression method, which has been widely used in the analysis of tablet data. However, its statistical consistency is still a mystery until today due to its complex structure. Honestly speaking, \cite{scornet2015consistency} is the only one which proved its consistency under  the framework of full trees and the splitting criterion CART. However, they need two technique conditions H(2.1) and H(2.2) that are still hard to be verified  until now. The main finding in this section is that RF is exactly is a large neural network with a special structure, which also satisfies Proposition \ref{pro_relu_multilayer}. Without adding those two additional technical assumptions, we can show that  the generalization error (variance) of RF will not diverge as the number of tree grows.

Let us formulate the structure of random forests. Following the notation in \cite{scornet2015consistency}, $\Theta$ is used to denote a random seed that is designed to
resample $a_n$ data points in the construction of a random  tree and select $q$ variables in its node splitting. Let $\{\Theta_b\}_{b=1}^{B_n}$ be a sequence of independent copies of $\Theta$. For the $b$-th tree, the CART tree is constructed by a re-sampled data $\D_n^b\subseteq \D_n$ whose sample size is $a_n$. This tree partition is denoted by $\{A_b^1,A_b^2,\ldots,A_b^{a_n}\}$ which is data dependent and each contains exactly one data point of $\D_n^b$. To be precise, $A_b^j=[e_{b,j}^1,f_{b,j}^1]\times \cdots \times [e_{b,j}^d,f_{b,j}^d]\subseteq [0,1]^d$ for each index $j$. Thus, the $b$-th tree estimator is
$$
\hat{m}_b(\bm{x}):= \sum_{\bm{X}_i\in\D_n^b}\sum_{j=1}^{a_n} \mathbb{I}(\bm{X}_i\in A_b^j) \mathbb{I}(\bm{x}\in A_b^j) Y_i.
$$
Finally, the forest estimator of conditional mean $m(\bm{x})$ in \cite{breiman2001random} is given by
\begin{equation}\label{8}
  \hat{m}_{B_n,RF}(\bm{x}):= \frac{1}{B_n}\sum_{b=1}^{B_n} \hat{m}_b(\bm{x}).
\end{equation}

\begin{proposition}\label{pro_RF}
  Let $\mathcal{N}\mathcal{N}_{a,b,c}$ be a neural network class with the Heaviside activation $\sigma_0(v):= \I(v\in\R)$, which has three layers with $a$ neurons in the first hidden layer and $b$ neurons in the second hidden layer and $c$ neurons in the final layer. Then, 
  $$
  \hat{m}_{B_n,RF}\in \mathcal{N}\mathcal{N}_{(d+1)a_n^2B_n,a_n(a_n+1)B_n,a_nB_n}
  $$
  such that
   \begin{enumerate}[(a).]
     \item $\hat{m}_{B_n,RF}=\sum_{j=1}^{B_n}g_j$, where $g_j\in \mathcal{N}\mathcal{N}_{(d+1)a_n^2,a_n(a_n+1),a_n}$;
     \item $\|g_j\|_\infty \le \max\{|Y_1|,\ldots,|Y_n|\}/B_n$.
   \end{enumerate}
\end{proposition}

Therefore, we know RF actually is a large neural network because both $a_n$ and $B_n$ diverge to infinity  as $n$ goes to infinity; see consistency conditions in  \cite{scornet2015consistency}. However,  this kind of neural network has its own ability to overcome overfitting instead of using the penalty regression method. This 
 is because that RF has a special structure satisfying two conditions in Proposition \ref{pro_RF} and this special structure plays a similar role with the penalized regression in \eqref{Jpenalty_multi}. Therefore,  the generalization error of RF is controlled by this subtle design and structure. Meanwhile, it is interesting to see that RF, a kind of large neural network, can avoid overfitting adaptively. According to Proposition \ref{pro_RF}, we now define this  kind of neural networks by
\begin{align*}
 NetRF:=\Bigg\{ g\in \mathcal{N}&\mathcal{N}_{(d+1)a_n^2B_n,a_n(a_n+1)B_n,a_nB_n}:  \\
  & g= \sum_{j=1}^{B_n}g_j, g_j\in \mathcal{N}\mathcal{N}_{(d+1)a_n^2,a_n(a_n+1),a_n},\|g_j\|_\infty \le \max\{|Y_1|,\ldots,|Y_n|\}/B_n \Bigg\}.
\end{align*}

By using the classical VC dimension method, it is not difficult to prove the following result.

\begin{proposition}\label{randomforest_rade}
  The Gaussian complexity of $ NetRF $ satisfies
  $$
  \mathcal{G}(NetRF;\{\bm{x}_i\}_{i=1}^n)\le c(d)\cdot \frac{a_n}{\sqrt{n}},
  $$
  where $c(d)$ only depends on the dimension $d$.
\end{proposition}

Therefore, we know our Gaussian complexity condition  is also satisfied in the case of RF. Thus, its generalization error is upper bounded and  independent of the number of trees. Furthermore, we also know from Proposition \ref{randomforest_rade} the parameter $a_n$ plays an important role in its generalization error and has similar effect with the penalty strength $\lambda_n$ in penalized regression. As $a_n=o(\sqrt{n})$, we can ensure the generalization error goes to zero as $n\to\infty$. On the other hand, RF uses a greedy method (CART) to tune parameters in $NetRF$ and thus its approximation error is hard to be analyzed. Until now,  we are only known its consistency for additive models; see \cite{scornet2015consistency} and \cite{klusowski2022large} and this part is out of scope of this paper.

\section{Robust regression for large neural networks  }
\subsection{Huber regression}
When the residual $\varepsilon= Y-\E(Y|\bm{X})$ follows heavy-tailed distribution, it is known that the least squares regression fails to recover the conditional mean function $m(\bm{X})$. To solve this problem, Huber loss, Cauchy loss and Tukey's biweight loss were proposed to estimate $m(\bm{X})$; see \cite{shen2021robust}. Basically, these robust methods were introduced to guard against outliers in the observations. When the input $Y_i$ is too large, these loss functions make a shrinkage and transform the corresponding risk value to a moderate one. In this section, we suppose the residual $\varepsilon$ only has finite moment up to $p$ and $m(\bm{X})$ is upper and lower bounded. Under this setting, previous papers studied Huber regression by using small networks such as \cite{shen2021robust} and \cite{fan2024noise}.  In this section, we aim to study this problem by using large neural networks. 

\begin{assump}\label{assump_heavytail}
  The residual $\varepsilon$ has  zero coditional mean and uniformly bounded conditional $p$-th moments for some $p\ge 1$,
  $$
  \E(\varepsilon|\bm{X}=\bm{x})=0\ \text{and}\ \E(|  \varepsilon|^p|\bm{X}=\bm{x} )\le v_p<\infty\ for \ all\  \bm{x}\in [0,1]^d.
  $$
\end{assump}

Sometimes, the tail error $\varepsilon$ is further known to be symmetric, like $T$ distribution. In this case, we set  the following condition.

\begin{assump}\label{assump_heavytail2}
      For each $\bm{x}\in [0,1]^d$, the conditional distribution of $\varepsilon|\bm{X}=\bm{x}$ is symmetric around $0$. 
\end{assump}

Besides, we assume the regression function is upper and lower bounded.

\begin{assump}\label{asp_bound}
  For some $M>0$, we have $\sup_{\bm{x}}{|m(\bm{x})|}\le M$.
\end{assump}

In this section, we consider the Huber loss to recover the regression function $m(\bm{x}),\bm{x}\in [0,1]^d$, which is defined below.

\begin{definition}\label{Huber_loss}
  Given some parameter $\tau_n\in (0,\infty]$,  Huber loss $\ell_{H,\tau_n}(\cdot)$ is defined as 
  $$
     \ell_{H,\tau_n}(v) =
\begin{cases}
    \frac{1}{2} v^2 & \text{if } |v|\le \tau_n \\
    \tau|v|-\frac{1}{2}\tau_n^2 & \text{if } |v|> \tau_n
\end{cases}
  $$
\end{definition}

From Definition \ref{Huber_loss}, it can be checked that  Huber loss is continuously differentiable with the score function $\ell'_{H,\tau_n}(v)=\min\{ \max{(-\tau_n,v)},\tau_n \}.$ When $\tau_n=\infty$, this loss is equivalent to the squares loss in previous section. According to Assumption \ref{asp_bound}, we now consider the truncated version of network class $\N\N(W_k,L_k)$ below:
\begin{equation}\label{tr_network}
\N\N^M(W_k,L_k):=\{ \min\{ \max{(-M,f)},M \} : f\in \N\N(W_k,L_k) \}.
\end{equation}
When ReLu activation is selected, we know any function in $\N\N^M(W_k,L_k)$ is also a neural network; see also in \eqref{hsfk23}. For any shrinkage parameter $\tau_n>0$, define the empirical Huber loss by
$$
\hat{R}_{\tau}(f)=\frac{1}{n}\sum_{i=1}^{n}{\ell_{H,\tau}(Y_i-f(\bm{X}_i))},\ f\in \N\N^M(W_k,L_k).
$$
Then, the estimator of $m(\bm{x})$ is a regularized large neural network given by
$$
\hat{m}_{H,n}\in \left\{ g : \hat{R}_{\tau}(g)+\lambda_nJ(g)\le \inf_{f\in \N\N^M(W_k,L_k)}{\left(\hat{R}_{\tau}(f)+\lambda_nJ(f)\right)}+\delta_{opt}^2 \right\},
$$
where $\delta_{opt}^2>0 $ is the optimization error and the penalty $J(\cdot)$ is defined in \eqref{pingan}. 

\begin{theorem}[Consistency of $\hat{m}_{H,n}$]\label{Theorem_Huber}
Under Assumption \ref{assump_heavytail} and suppose $m\in \mathcal{H}(d, l, \mathcal{P})$ and $\tau_n\asymp (n/L_k)^{\frac{\beta_1}{(2p-2)(2\beta_1+1)+1}}$, we have
  $$
     \|\hat{m}_{H,n}-m\|_2^2=O_p\left({ \delta_{opt}^2}+ \max\left\{ (L_kW_k)^{-2\alpha_1}, (n/L_k)^{-\frac{1}{4}\cdot\frac{(2p-2)2\beta_1}{(2p-2)(2\beta_1+1)+1}} \right\} \right).
  $$
where $\alpha_1= \min_{(\alpha, C, t) \in \mathcal{P}}\left\{ \frac{2\alpha}{t}\right\}$ and $\beta_1:= \min_{(\alpha, C, t) \in \mathcal{P}}\left\{ \frac{\alpha}{t+1}\right\}/l$ and  $W_k\gtrsim n^{c(\mathcal{P})}$.  When the residual further satisfies Assumption \ref{assump_heavytail2}, we have a faster rate
\begin{equation}\label{fjsfbjk34}
     \|\hat{m}_{H,n}-m\|_2^2=O_p\left({ \delta_{opt}^2}+ \max\left\{ (L_kW_k)^{-2\alpha_1}, (n/L_k)^{-\frac{1}{2}\cdot\frac{(2p-2)2\beta_1}{(2p-2)(2\beta_1+1)+1}} \right\} \right).
\end{equation}
\end{theorem}
When the error has higher moment ($\E|\varepsilon|^p<\infty$ for large $p$), the bound in \eqref{fjsfbjk34} increase to the case in Section \ref{sec:2} where the residual follows Gaussian distribution;  see Theorem \ref{Theorem1}.

\subsection{Quantile regression}
In this section, we consider the quantile regression in which the conditional quantile function $$q_{\tau}(\bm{x}):=\inf\{y:\P(Y\leq y|\bm{X}=\bm{x})>\tau\}, \ \forall \bm{x}\in [0,1]^d$$ is what need to be estimated. Compared with mean regression, quantile regression provides a comprehensive characterization of the conditional distribution of the response variable given the covariates, while also being more robust to outliers and heavy-tailed distributions.  Here, we also use the network in  $\N\N^M(W_k,L_k)$, which is given in \eqref{tr_network}, to estimate $q_\tau(\bm{x}),\bm{x}\in [0,1]^d$. To recover $q_{\tau}(\bm{x})$ from the noised data $\D_n$, the following loss function is considered
$$
\rho_{\tau}(v):= |v|+(2\tau-1)v, \ v\in\R.
$$
Now, consider the empirical risk function
$$
\hat{R}_{\tau}^{qua}(f)=\frac{1}{n}\sum_{i=1}^{n}{\rho_{\tau}(Y_i-f(\bm{X}_i))},\ f\in \N\N^M(W_k,L_k).
$$
Then, our estimator of $q_\tau(\bm{x})$ is a regularized large neural network given by
$$
\hat{q}_{\tau,n}\in \left\{ g : \hat{R}_{\tau}^{qua}(g)+\lambda_n J(g)\le \inf_{f\in \N\N^M(W_k,L_k)}{\left(\hat{R}_{\tau}^{qua}(f)+\lambda_nJ(f)+\delta_{opt}^2 \right)} \right\},
$$
where $\delta_{opt}^2>0 $ is the optimization error  and the penalty $J(\cdot)$ is defined in \eqref{pingan}.

\begin{assump}\label{assump_quantile}
  There are constants $c,\delta,\Delta>0$ such that for any $|v|\le \delta$ and $y\in \{y:|y-q_\tau(\bm{x})|\le \Delta\}$, it holds
  $$
   |F_{Y|\bm{X}=\bm{x}}(y+v)-F_{Y|\bm{X}=\bm{x}}(y)|\ge c|v|, \quad a.s..
  $$
  Moreover, almost surely for $\bm{X}\in [0,1]^d$, $F_{Y|\bm{X}=\bm{x}}(\cdot)$ is a Lipshitz function over $\R$ with the  Lipshitz  constant $L>0$.
\end{assump}

Assumption \ref{assump_quantile} is an adaptive self-calibration governing the conditional distribution of $Y$ given $\bm{X}$, which plays an important role when we establish the relationship between the excess risk and the mean squared error.  This assumption was popularly used in many  papers that studied quantile regression using machine learning tools, such as \cite{feng2024deep}, \cite{padilla2022quantile} and \cite{madrid2022risk}. However, the sizes of networks  in these papers are small and the consistency of their estimators can be  guaranteed if the classical arguments of VC dimension hold. 

\begin{theorem}\label{Theorem_quantile}
   Under Assumption \ref{assump_quantile} and suppose $q_\tau\in \mathcal{H}(d, l, \mathcal{P})$ and $\E|Y|<\infty$, we have
$$
     \|\hat{m}_{H,n}-m\|_2^2=O_p\left( { \delta_{opt}^2}+\max\{(L_kW_k)^{-2\alpha_1}, (n/L_k)^{-\frac{\beta_1}{2\beta_1+1} }\} \right),
$$
  where $\alpha_1= \min_{(\alpha, C, t) \in \mathcal{P}}\left\{ \frac{2\alpha}{t}\right\}$ and $\beta_1:= \min_{(\alpha, C, t) \in \mathcal{P}}\left\{ \frac{\alpha}{t+1}\right\}/l$ and $W_k\gtrsim n^{c(\mathcal{P})}$. 
\end{theorem}

\section{Classification for Large neural network }
Actually, neural networks are mostly used as  powerful tools for classification. For nonparametric regression, people prefer random forests than neural networks. In this section, we show that large neural networks with regularization are also statistically consistent in label classification problems. Let $\CN_k$ be a class of neural networks used for classification. Any classification network in $\CN_k$ usually connects to a feed-forward neural network. Namely,
$$
  \CN_k:=\left\{    \boldsymbol{\Psi}\circ g: g\in \N\N_k(W_k,L_k)\right\},
$$
where $\N\N_k(W_k,L_k)$ is defined in  \eqref{;hkajdk1} and the  output activation is chosen to be the softmax function $ \boldsymbol{\Psi}$. Specifically, if the last hidden layer has $K$ neurons, this softmax function is given by
\begin{equation*}
    \boldsymbol{\Psi} : \mathbb{R}^{K} \to \mathbb{R}^{K},\; (x_1,\ldots,x_K) \to \pr{\frac{e^{x_1}}{\sum_{j=1}^{K}e^{x_j}}, \dots , \frac{e^{x_{K}}}{\sum_{j=1}^{K}e^{x_j}}}.
\end{equation*}

Let us formulate this problem below. Consider a multi-class classification problem with $K$ classes.
Let $\mathcal{X} \;= [0, 1]^d$ be the input space, and $\mathcal{Y} = \{\bm{e}_i\}_{i=1}^K$ be the set of labels where 
$$
\bm{e}_k:=(0,\ldots,0,\underbrace{1}_{k\text{-th\ position}},0,\ldots,0)^T. 
$$
Assume that the data $\pr{\bm{X}, \bm{Y}} \in \mathcal{X} \times \mathcal{Y}$ is generated from the following model:
\begin{equation}
    \bm{Y}_{\cdot,k} \mid \bm{X} = \bm{x} \sim \text{Bernoulli}(\eta_k(\bm{x})), \quad \bm{X} \sim P_{\bm{X}}, \quad k = 1, \dots , K, \label{eq:data-generating model}
\end{equation}
where $\eta_k(\bm{x}) := \mathbb{P}\pr{\bm{Y} = \bm{e}_k \mid \bm{X} = \bm{x}}$ is the true conditional class probabilities, and $P_{\bm{X}}$ is the unknown distribution on the input space $\mathcal{X}$ and $  \bm{Y}_{\cdot,k} $ denotes the $k$-th component of $  \bm{Y} $.
We denote the joint distribution of $\bm{X}$ and $\bm{Y}$ as $P$.
Let $\mathcal{D}_n = \{(\bm{X}_1, \bm{Y}_1),\dots,(\bm{X}_n, \bm{Y}_n)\}$ be an i.i.d. sample with size $n$ from the population distribution $P$.
The goal of the classification problem is to find a function $\bm{f}: \mathcal{X} \to \mathbb{R}^K$ (called the decision function) that predicts $\bm{Y}$ well when $\bm{X}$ are given.
Here, we focus on the nonparametric estimation of conditional class probabilities.

In the estimation of conditional class probabilities, we typically consider the maximum likelihood estimation, i.e., we minimize the negative log-likelihood function.
Let $\bm{p}(\bm{x}) = (p_1(\bm{x}), \dots , p_K(\bm{x}))^\top$ be a model of the conditional class probability to estimate the true one $\bm{\eta}(\bm{x}) = (\eta_1(\bm{x}), \dots , \eta_K(\bm{x}))^\top$.
Given the data $\mathcal{D}_n$, 
the likelihood for the conditional class probability function $\bm{p}(\bm{x})$ is given by
$\prod_{i=1}^n\prod_{k=1}^K p_k(\bm{X}_i)^{Y_{ik}}$.
Here, $Y_{ik}$ is the $k$-th component of $\bm{Y}_i$.
The negative log-likelihood function is
\begin{equation}
    L(\bm{p}) := -\frac{1}{n}\sum_{i=1}^n\sum_{k=1}^K Y_{ik}\log p_k(\bm{X}_i) = -\frac{1}{n}\sum_{i=1}^n \bm{Y}_i^\top \log \bm{p}(\bm{X}_i). \label{eq: negative likelihood function}
\end{equation}
For any  $\boldsymbol{\Psi}\circ g\in\CN_k $, it is natural to define the complexity of classification network by $J^C(\boldsymbol{\Psi}\circ g)=J(g)$ where $J(g)$ is already given in \eqref{pingan}. Then, the regularized maximum likelihood estimator (MLE) is 
\begin{equation}
    \hat{\bm{p}}_{n,k} \in \left\{ \bm{p}_{opt} \in \CN_k :  L({\bm{p}}_{opt})+\lambda_nJ({\bm{p}}_{opt})\le \inf_{\bm{p}\in\CN_k}\{ L(\bm{p})+\lambda_nJ(\bm{p})+\delta_{opt}^2\} \right\},
 \label{eq:mle}
\end{equation}
where $\CN_k$ is a class of candidate functions and $\delta_{opt}^2>0$ denotes the optimization error.
Note that $L(\bm{p})\ge 0$ for each p.d.f. $\bm{p}\in (0,1)$. In this section, all estimators $\hat{\bm{p}}_{n}^k = (\hat{p}_{n,1}, \dots , \hat{p}_{n,K})^\top$ are considered as probability vectors for all $\bm{x} \in \mathcal{X}$, i.e., $p_k(\bm{x}) \geq 0$ for any $\bm{x} \in \mathcal{X}, k \in [K]$ satisfying \(\sum_{k=1}^K p_k(\bm{x}) = 1\) for all $\bm{x} \in \mathcal{X}$.

In density estimation problem, the squared Hellinger distance is always employed to measure the estimation error bound; see \cite{sen2018gentle}.  Actually, for any two probability measures $P, Q$ on the same measurable space, the squared Hellinger distance is defined as
\begin{equation*}
    H^2\pr{P, Q} := \frac{1}{2} \int \pr{\sqrt{dP} - \sqrt{dQ}}^2.
\end{equation*}
and we measure the estimation error by
\begin{equation}
    R(\boldsymbol{\eta}(\bm{X}), \hat{\bm{p}}_{n,k}(\bm{X})) 
    := \mathbb{E}_{\bm{X}}\pr{H^2\pr{\boldsymbol{\eta}(\bm{X}), \hat{\bm{p}}_{n,k}(\bm{X})}}. \label{eq: Hellinger excess risk}
\end{equation}
Since the Hellinger distance is always upper bounded, we can avoid the divergence problem of KL distance which happens in \cite{bos2022convergence} and \cite{bilodeau2023minimax}. (See also discussions in these paper: If the density estimator is piecewise constant, the corresponding KL divergence goes to infinity as $n\to\infty$.) Thus, considering the convergence in terms of the Hellinger distance allows us more convenient to study the convergence rate of  $\hat{\bm{p}}_{n,k}$.

At this step, we makes an assumption on the true conditional density, where we also allow the number of labels $K$ diverges with $n$.  

\begin{assump}\label{class:assump}
  The true conditional density function $\bm{\eta}(\bm{x}), \bm{x}\in [0,1]^d$ is bounded from  below. Namely, there are constants $c\in (0,1)$ and $\gamma\ge 0$ such that
  $$
   \P\pr{ \eta_k(\bm{X})\ge cK^{-\gamma} , \ \forall k\in [K]}=1.
  $$
\end{assump}

For any network $p\in\CN_k$, we can write
$$
\bm{p}(\bm{x})=\pr{\frac{e^{\bm{p}_1^{last}(\bm{x})}}{\sum_{j=1}^{K}e^{\bm{p}_j^{last}(\bm{x})}}, \dots , \frac{e^{\bm{p}_K^{last}(\bm{x})}}{\sum_{j=1}^{K}e^{\bm{p}_j^{last}(\bm{x})}}}.
$$
If Assumption \ref{class:assump} is satisfied,  our Lemma \ref{lemma_neural network representation} shows that the true conditional density $\bm{\eta}(\bm{x})$ also admits a similar decomposition:
\begin{equation}\label{UUUL}
  \bm{\eta}(\bm{x})=\pr{\frac{e^{\bm{\eta}_1^{last}(\bm{x})}}{\sum_{j=1}^{K}e^{\bm{\eta}_j^{last}(\bm{x})}}, \dots , \frac{e^{\bm{\eta}_K^{last}(\bm{x})}}{\sum_{j=1}^{K}e^{\bm{\eta}_j^{last}(\bm{x})}}}^T, \ \bm{x}\in [0,1]^d.
\end{equation}
Meanwhile, $\bm{\eta}_j^{last}(\bm{x})=\ln(c\cdot \bm{\eta}_j(\bm{x}))$ for each $j\in [K]$ and some $c>0$ and this series of functions is unique. If $\bm{\eta}_j^{last}(\bm{x})$ is relatively large, $\eta_j$ is close to $1$; otherwise, the probability  function will decrease to $0$.  Therefore, we call $\bm{\eta}_j^{last}$ the weight function of the $j$-th coordinate of $\eta$, namely $\eta_j$.

\begin{theorem}[Error bound for classification neural networks]\label{theorem_consistency_class}
 Choose $r>0$, $\lambda_n\asymp K^2/\sqrt{n}$ and $L_k\asymp \ln n$. If the true density $\bm{\eta}(x)$ satisfies Assumption \ref{class:assump} and each weight function $\bm{\eta}_j^{last}\in \mathcal{H}(d, l, \mathcal{P})$, we have
  \begin{align*}
   R(\boldsymbol{\eta}(\bm{X}), \hat{\bm{p}}_{n,k}(\bm{X}))\lesssim K^{\frac{3}{2}}\max\left\{(L_kW_k)^{-\alpha_1}, \left(\frac{n}{K} \right)^{-\frac{\beta_1}{\beta_1+2}}\ln n\right\}  + \frac{K^{\frac{3}{2}\vee \gamma}}{\sqrt{n}} +    \delta_{opt}^2
  \end{align*}
  with the probability larger than $1-\ln n \cdot n^{-r}$. In above inequality, $\alpha_1= \min_{(\alpha, C, t) \in \mathcal{P}}\left\{ \frac{2\alpha}{t}\right\}$ and $\beta_1:= \min_{(\alpha, C, t) \in \mathcal{P}}\left\{ \frac{\alpha}{t+1}\right\}/l$ and $W_k\gtrsim n^{c(\mathcal{P})}$ for some $c(\mathcal{P})>0$. 
\end{theorem}

In practice problems, the number of labels $K$ is always fixed. In this case, the error bound in Theorem \ref{theorem_consistency_class} does not depend on the width $W_k$ and we find it is sufficient to guarantee the consistency if  $L_k\asymp\ln n$ only. Similar to previous sections, our result proves the statistical consistency for classification networks when $W_k$ is very large.  On the other hand, from Theorem \ref{theorem_consistency_class} we can guarantee  the consistency property of classification networks for some $K=o(n)$. To our best knowledge, this is the first result in literature that gives consistency result for large classification networks.

\newpage
\section{Proofs}\label{sec_proof}

\subsection{Prerequisite for  Gaussian and Rademacher complexity  }

Similar to the Gaussian complexity in Definition \ref{Gaussian_complexity}, we also need Rademacher complexity in  many proofs and  is given below.

\begin{definition}
 For any fixed points $\{x_i\}_{i=1}^n\subseteq\R^d$, define the Rademacher complexity of $\N_k$ by
  $$
  \mathcal{R}(\N_k;\{x_i\}_{i=1}^n):=\E_{r_i}\left( \frac{1}{n}\sup_{g\in\N_k}{\sum_{i=1}^{n}r_i\cdot g(x_i)}\right),
  $$
  where $(r_1,\ldots,r_n)$ are independent and each follows distribution $\P(r_1=\pm 1)=\frac{1}{2}$.
\end{definition} 

Meanwhile, we also need to introduce two intermediate terms related to Gaussian and Rademacher complexity respectively:
\begin{align*}
   |\mathcal{G}|(\N_k;\{x_i\}_{i=1}^n) &:=\E_{s_i}\left( \frac{1}{n}\sup_{g\in\N_k}{\left| \sum_{i=1}^{n}s_i\cdot g(x_i)\right|}\right)\\
     |\mathcal{R}|(\N_k;\{x_i\}_{i=1}^n) &:=\E_{s_i}\left( \frac{1}{n}\sup_{g\in\N_k}{\left| \sum_{i=1}^{n}r_i\cdot g(x_i)\right|}\right),
\end{align*}
where $s_i\sim N(0,1)$ are independent and $r_i$ are also independent with $\P(r_1=\pm 1)=\frac{1}{2}$.

Without loss of generality, we  assume $0\in\N_k$ in this section. For any $\{x_i\}_{i=1}^n\subseteq\R^d$, we can  bound $ |\mathcal{R}|(\N_k;\{x_i\}_{i=1}^n)$ and $ |\mathcal{G}|(\N_k;\{x_i\}_{i=1}^n)$ by $ \mathcal{G}(\N_k;\{x_i\}_{i=1}^n)$:
\begin{align}
  |\mathcal{R}|(\N_k;\{x_i\}_{i=1}^n)&\le \sqrt{\frac{8}{\pi}}\mathcal{G}(\N_k;\{x_i\}_{i=1}^n) \label{878}\\
  |\mathcal{G}|(\N_k;\{x_i\}_{i=1}^n)&\le 2\mathcal{G}(\N_k;\{x_i\}_{i=1}^n) \nonumber
\end{align}
Therefore, condition (C1) can be also used to bound above two terms. This piece of fact will be frequently used in the following proofs. 

To save space, we only prove \eqref{878} here.  In fact, 
\begin{align}
  |\mathcal{R}|(\N_k;\{x_i\}_{i=1}^n) &=  \frac{1}{n}\E_{s_i} \max\left\{ \sup_{g\in\N_k}{ \sum_{i=1}^{n}s_i\cdot g(x_i)}, \sup_{g\in\N_k}{ -\sum_{i=1}^{n}s_i\cdot g(x_i)} \right\} \nonumber \\
  &\le  \frac{1}{n}\E_{s_i} \left(  \sup_{g\in\N_k}{ \sum_{i=1}^{n}s_i\cdot g(x_i)}+ \sup_{g\in\N_k}{ -\sum_{i=1}^{n}s_i\cdot g(x_i)} \right)\label{09-} \\
  &= 2\mathcal{R}(\N_k;\{x_i\}_{i=1}^n), \label{1-900-}
\end{align}
where \eqref{09-} holds because the two terms in maximum function are all nonnegative. Since $(r_1|s_1|,\cdots,r_n|s_n|)\sim N(\mathbf{0},\mathbf{I}_{n})$, 
\begin{align}
  \mathcal{G}(\N_k;\{x_i\}_{i=1}^n) & = \E_{r_i} \E_{s_i}\left( \sup_{g\in\N_k}{\sum_{i=1}^{n} |s_i|\cdot r_i g(x_i) }\Big|r_1,\cdots,r_n\right) \nonumber\\
  &\ge   \E_{r_i} \sup_{g\in\N_k}\E_{s_i}\left( {\sum_{i=1}^{n} |s_i|\cdot r_i g(x_i) }\Big|r_1,\cdots,r_n\right) \nonumber\\
  &=\sqrt{\frac{2}{\pi}}  \mathcal{R}(\N_k;\{x_i\}_{i=1}^n). \label{1-900-2}
\end{align}
Therefore, the combination of \eqref{1-900-} and \eqref{1-900-2} proves \eqref{878}.

\subsection{Deep neural network approximation with restricted network norm}

In this section, we prove the following result.

\begin{theorem}\label{reg_multi_approx}
  For any $m\in \mathcal{H}(d, l, \mathcal{P})$ with $\sup _{(\alpha, C, t) \in \mathcal{P}} \max \{\alpha, C, t\}<\infty$, we have
  $$
   \inf_{f\in \mathcal{N}_k }\|m-f\|_\infty \lesssim U^{-\frac{\gamma^*}{l}}
  $$
  provided that $W\ge c_1(\mathcal{P})U^{\frac{2t^{**}+\alpha^{**}}{2t^{**}}}$ and $L\ge c_2(\mathcal{P})$ and $W>1$. Here,
  $$
    \gamma^*:= \min_{(\alpha, C, t) \in \mathcal{P}}\left\{ \frac{\alpha}{t+1}\right\} \quad\text{and}\quad (t^{**},\alpha^{**})= \sup_{(\alpha,C,t)\in\mathcal{P}}\frac{\alpha}{t}.
  $$
\end{theorem}

First, we consider a more general  neural network which has $d$ inputs and $o$ outputs and its matrix norm is at most $U$. Namely, 
$$
  \mathcal{N} \mathcal{N}_{d, o}\left(W, L, U\right):=\left\{  g\ \text{has the form in } \eqref{multinetwork}: J(g)\le U \right\},
$$
where the penalty $J(g)$ is defined in \eqref{Jpenalty_multi}. This penalized network class has some properties below which are useful in our network construction later.

\begin{proposition}\label{Pro_largenet}
 Let $\phi_{1} \in \mathcal{N} \mathcal{N}_{d_{1}, o_{1}}\left(W_{1}, L_{1}, U_{1}\right)$ and $\phi_{2} \in \mathcal{N} \mathcal{N}_{d_{2}, o_{2}}\left(W_{2}, L_{2}, U_{2}\right)$.\\
(i) If $d_{1}=d_{2}, o_{1}=o_{2}, W_{1} \leq W_{2}, L_{1} \leq L_{2}$ and $U_{1} \leq U_{2}$, then $$\mathcal{N} \mathcal{N}_{d_{1}, o_{1}}\left(W_{1}, L_{1}, U_{1}\right) \subseteq \mathcal{N} \mathcal{N}_{d_{2}, o_{2}}\left(W_{2}, L_{2}, U_{2}\right).$$
(ii) (Composition) If $o_{1}=d_{2}$, then $\phi_{2} \circ \phi_{1} \in \mathcal{N} \mathcal{N}_{d_{1}, o_{2}}\left(\max \left\{W_{1}, W_{2}\right\}, L_{1}+L_{2}, U_{2} \max \left\{U_{1}, 1\right\}\right)$. Let $A \in \mathbb{R}^{d_{2} \times d_{1}}$ and $\boldsymbol{b} \in \mathbb{R}^{d_{2}}$. Define the function $\phi(\boldsymbol{x}):=\phi_{2}(A \boldsymbol{x}+\boldsymbol{b})$ for $\boldsymbol{x} \in \mathbb{R}^{d_{1}}$, then $\phi \in \mathcal{N} \mathcal{N}_{d_{1}, o_{2}}\left(W_{2}, L_{2}, U_{2} \max \{\|(A, \boldsymbol{b})\|, 1\}\right)$.\\
(iii) (Concatenation) If $d_{1}=d_{2}$, define $\phi(\boldsymbol{x}):=\left(\phi_{1}(\boldsymbol{x}), \phi_{2}(\boldsymbol{x})\right)$, then $$\phi \in \mathcal{N} \mathcal{N}_{d_{1}, o_{1}+o_{2}}\left(W_{1}+\right. \left.W_{2}, \max \left\{L_{1}, L_{2}\right\}, \max \left\{U_{1}, U_{2}\right\}\right).$$ 
(iv) (Linear Combination) If $d_{1}=d_{2}$ and $o_{1}=o_{2}$, then, for any $c_{1}, c_{2} \in \mathbb{R}, c_{1} \phi_{1}+c_{2} \phi_{2} \in \mathcal{N} \mathcal{N}_{d_{1}, o_{1}}\left(W_{1}+W_{2}, \max \left\{L_{1}, L_{2}\right\},\left|c_{1}\right| U_{1}+\left|c_{2}\right| U_{2}\right)$.\\
(v) (Boundness) $\|\phi_1\|_\infty\le J(\phi_1)\le U$, where $\|\cdot\|_\infty$ denotes the supremum norm of any function. 
\end{proposition}

\begin{proof}
  The proof of (i-iv) can be found in \cite{jiao2023approximation}. Now, we prove (v) by induction. Since 
  $$
    \left| \sum_{j=1}^{k}a_j\sigma(\theta_j^Tx+b_j) \right|\le \sqrt{d+1} \sum_{j=1}^{k}|a_j|  \|(\theta_j,b_j)\|_1,
  $$ 
  thus (v) is true for the depth of two. Suppose it holds for all networks with depth less than $L$. Note that $\phi_1\in \mathcal{N} \mathcal{N}_{d_{1}, o_{1}}\left(W_{1}, L_{1}, U_{1}\right)$.  Choose any output of $\phi_1$ which is denoted by $\phi_{1,s}$. Then, we have
  $$
     \phi_{1,s}= {\bf{a}} \cdot \sigma( {\bf{A^{L-1}}} \phi^{L-1}_{1}+{\bf{b}}),
  $$
  where $\phi^{L-1}_{1}\in \mathcal{N} \mathcal{N}_{d_{L-1}, o_{L-1}}\left(W_{1}, L_{1}-1\right)$. Note that ${\bf{a}}$ is a row vector and ${\bf{b}}$ is a column vector. According to the homogeneity property of ReLu activation, we can suppose $\|{\bf{b}}\|_1\ge 1$. Otherwise, we just do the coefficients scaling and the penalty part $\|{\bf{a}}\|_1\| ({\bf{A^{L-1}}}, {\bf{b}})\|_1$ does not change. Therefore, it can be seen 
  \begin{align*}
        \| \phi_{1,s}\|_\infty  &\le \|{\bf{a}}\|_1\| ({\bf{A^{L-1}}}, {\bf{b}})\|_1 \|\phi^{L-1}_{1}\|_\infty\\
        &\le \|{\bf{a}}\|_1\| ({\bf{A^{L-1}}}, {\bf{b}})\|_1  J(\phi^{L-1}_{1}),
  \end{align*}
  which is what we desire. 
\end{proof}

Next, we introduce a approximation result of $\mathcal{N} \mathcal{N}_{d, 1}\left(W, L, U\right)$; see Lemma \ref{lemma_jiaoapp} below. An interesting observation is that this error bound only depends on the network norm. This result was proven by mostly following the network construction in \cite{yarotsky2017error}.

\begin{lemma}[\cite{jiao2023approximation}]\label{lemma_jiaoapp}
  For any $h\in H^\alpha([0,1]^d)$ with $\alpha>0$, we have
  $$
   \inf_{f\in \mathcal{N} \mathcal{N}_{d, 1}\left(W, L, U\right)}\|h-f\|_\infty \lesssim U^{-\frac{\alpha}{d+1}}
  $$
  provided that $W\gtrsim U^{\frac{2d+\alpha}{2d+2}}$ and $L\gtrsim \ln (d+\alpha)$.
\end{lemma}

Now we are ready to make the proof. The key idea is that {neural network approximation is preserved under compositions}. To be specific, if $f$ and $g$ can be approximated by neural networks $\hat{f}$ and $\hat{g}$, each with an $\|\cdot\|_\infty$-error of $\epsilon$, and $g$ is an $L$-Lipschitz function, then $\hat{g} \circ \hat{f}$ approximates $g\circ f$ with an $\|\cdot\|_\infty$-error of $(L + 1)\epsilon$. The former `$\circ$' refers to the network composition, and the latter `$\circ$' refers to function composition. Therefore, suppose the target $f_0$ is a composition of several low-dimensional smooth functions $g_1, \ldots, g_k$, then in order to approximate $f_0$ well, we only need to approximate each $g_i$ sufficiently well.

We define  $C_{\max} = \sup_{(\alpha, C, t) \in \mathcal{P}} C$ and $\alpha_{\max} = \sup_{(\alpha, C, t) \in \mathcal{P}} \alpha$ and $t_{\max} = \sup_{(\alpha, C, t) \in \mathcal{P}} t$. Let $h^{(l)}_1(\bm{x})=f_0$ for arbitrary $f_0$ that belongs to the function class $\mathcal{H}(d, l,\mathcal{P})$ with fixed integer $l>1$. To obtain $h^{(l)}_1(\bm{x}) \in \mathcal{H}(d, l,\mathcal{P})$, one needs to compute various hierarchical composition models at level $i\in \{ 1, \ldots, l-1\}$, the number of which is denoted by $M_i$. At level $i\in \{1, \ldots, l\}$, let $h^{(i)}_j: \R^d \to \R$ be the $j$-th ($j\in \{1, \ldots, M_i\}$) hierarchical composition model. The dependence of $h^{(i)}_j$ on $h^{(i-1)}_{\cdot}$ depends on a smooth function $g^{(i)}_j : \R^{t^{(i)}_j} \to \R$ in $C_j^i\cdot H^{\alpha^{(i)}_j}([0,1]^{t^{(i)}_j})$  for some $(\alpha^{(i)}_j, C_j^i, t^{(i)}_j) \in \mathcal{P}$. Recursively, $h^{(l)}_1(\cdot)$ is defined as 
\begin{align*}
	h^{(i)}_j(\bm{x}) = g^{(i)}_j \Bigg( h^{(i-1)}_{\sum_{\ell=1}^{j-1} t^{(i)}_\ell + 1} (\bm{x}), \ldots , h^{(i-1)}_{\sum_{\ell=1}^{j} t^{(i)}_\ell} (\bm{x})\Bigg) 
\end{align*}
for $j\in \{ 1,\ldots, M_i\}$ and $i \in \{ 2, \ldots, l\}$, and 
\begin{align*}
	h^{(1)}_j(\bm{x}) = g_j^{(1)}\Bigg( \bm{x}_{\pi(\sum_{\ell=1}^{j-1} t^{(1)}_\ell + 1)}, \ldots , \bm{x}_{\pi(\sum_{\ell=1}^{j} t^{(1)}_\ell)}\Bigg)	
\end{align*}
for some $\pi:\{1,\ldots, M_1\}\to \{1,\ldots, d\}$. The quantities $M_1, \ldots, M_l$ can be defined recursively as 
\begin{align*}
	M_i = \begin{cases}
 		1 & \qquad i=l ,  \\
 		\sum_{j=1}^{M_{i+1}} t^{(i+1)}_j & \qquad i\in \{1,\ldots, l-1\},
 \end{cases}
\end{align*}  then it is easy to see that $M_i\le t_{\max}^{l-i}$ for any $i\in \{1, \ldots, l\}$.

Moreover, define 
\begin{align*} 
	C_{f_0} = \max_{i\in \{1,\ldots, l\}, j\in \{1,\ldots, M_i\}} \|g^{(i)}_j\|_\infty \lor 1
\end{align*} and let $\mathcal{D}_{j}^{(i)}$ be the domain of function $g^{(i)}_{j}$ under the hierarchical composition model, i.e., 
\begin{align*}
	\mathcal{D}_{j}^{(i)} = \begin{cases}\Bigg\{ \bigg(h^{(i-1)}_{\sum_{\ell=1}^{j-1} t^{(\ell)}_\ell + 1} (\bm{x}), \ldots , h^{(i-1)}_{\sum_{\ell=1}^{j} t^{(\ell)}_\ell} (\bm{x}) \bigg): \bm{x}\in [0,1]^d\Bigg\}& i \in \{2,\ldots, l\} \\
		[0,1]^{t^{(1)}_j} & i=1.
 \end{cases}
\end{align*} 
It is easy to see that $T_{f_0}$ can be upper bounded by the universal constant $C_{\max}$.  We thus have $\mathcal{D}_j^{(i)} \subseteq [-C_{\max}, C_{\max}]^{t^{(i)}_j}$. Without loss of generality we may assume $\mathcal{D}_j^{(i)}=[-C_{\max}, C_{\max}]^{t^{(i)}_j}$; otherwise we can simply extend $g_j^{(i)}$ to the cube $[-C_{\max},C_{\max}]^{t^{(i)}_j}$ and the following analysis remains valid.

\medskip
\noindent {\sc Step 1. Construction of neural network.} In the rest of the proof, for notational convenience we use $\mathcal{F}(N,L)$ to denote a deep ReLU neural network with width $N$ and depth $L$. 

Fix $i \in \{1,\ldots, l\}$ and $j\in \{1, \ldots, M_i\}$. Note that each $g^{(i)}_j$ is a smooth function in $H^{\alpha^{(i)}_j}([-C_{\max}, C_{\max}]^{t^{(i)}_j})$.
\begin{align*}
	\bar{g}^{(i)}_j(\bm{z}) = g^{(i)}_j(2C_{\max}\bm{z}-C_{\max}) ~~\text{for}~~ \bm{z} \in [0,1]^{t^{(i)}_j},
\end{align*} 
so that  $\bar{g}^{(i)}_j$  is a smooth function in $H^{\alpha^{(i)}_j}([0,1]^{t^{(i)}_j})$, and satisfies
\begin{align*}
	g^{(i)}_j(\bm{z}) = \bar{g}^{(i)}_j\bigg(\frac{\bm{z}+C_{\max}}{2C_{\max}}\bigg) ~~\text{for}~~ \bm{z}\in \mathcal{D}_j^{(i)}.
\end{align*}

For any given $W, L\in \mathbb{N}$, Lemma \ref{lemma_jiaoapp} ensures that there exists a function $\tilde{g}_j^{(i)}$ from some deep ReLU neural network $\tilde{g}_j^{(i)}$ with width $W'\ge C_1U^{\frac{2t^{(i)}_j+\alpha^{(i)}_j}{2t^{(i)}_j+2}}$ and depth $L'\ge 2\log_2( t^{(i)}_j+\alpha^{(i)}_j )+2$ such that 
\begin{align*}
	\Bigg\|\tilde{g}_j^{(i)}\bigg(\frac{\bm{z}+C_{max}}{2C_{max}}\bigg) - \bar{g}^{(i)}_j\bigg(\frac{\bm{z}+C_{max}}{2C_{max}}\bigg)\Bigg\|_\infty \le C_2 (U)^{-\frac{\alpha_{j}^{(i)}}{1+t_{j}^{(i)}}} \le C_2 (U)^{-\gamma^*}\text{ for all } \bm{z} \in \mathcal{D}_j^{(i)}.
\end{align*}
It should be noted that the constants $C_1$ and $C_2$ depend on the parameters $(\alpha^{(i)}_j, t_j^{(i)})$. Since there are only finitely many $g^{(i)}_j$, we can simply choose $(C_1 , C_2)$ to be the largest among all $(C_1 , C_2)$ depending on $(\alpha^{(i)}_j, t_j^{(i)})$. Here  both $C_1$ and $C_2$ are also universal constants that only depend on $\alpha_{\max}$ and $t_{\max}$.

Next, consider a `truncated' version of $\tilde{g}_j^{(i)}$, defined as
\begin{align*}
	\hat{g}_{j}^{(i)}(\bm{z}) = \max\{\min\{\tilde{g}_{j}^{(i)}(\bm{z}),C_{max}\}, -C_{max}\}
\end{align*} 
where $\sigma(v) = \max\{v, 0\}$ is the ReLU activation function. For any $v_1,v_2\in\R$, $v_1= \sigma(v_1)-\sigma(-v_1)$, $|v|=\sigma(v_1)+\sigma(-v_1)$. Meanwhile, $\min(v_1,v_2)=\frac{1}{2}(v_1+v_2-|v_1-v_2|)$ and $\max(v_1,v_2)=\frac{1}{2}(v_1+v_2+|v_1-v_2|)$. Thus, we can rewrite $\hat{g}_{j}^{(i)}(\bm{z})$ in the neural network form:
\begin{equation}\label{hsfk23}
\begin{pmatrix}
  \frac{1}{2}&  -\frac{1}{2}& \frac{1}{2}& \frac{1}{2}
\end{pmatrix}
\circ \sigma\circ 
\left[
\begin{pmatrix}
  1 \\
  -1 \\
  1 \\
  -1 
\end{pmatrix}
\begin{pmatrix}
  \frac{1}{2}&  -\frac{1}{2}& \frac{1}{2}& \frac{1}{2}
\end{pmatrix}
\bm{x}+
{\bf{v}}
\right]
\circ \sigma\circ
\left[
\begin{pmatrix}
  1 \\
  -1 \\
  1 \\
  -1 
\end{pmatrix}
\tilde{g}_{j}^{(i)}(\bm{z})-
{\bf{v}}
\right],
\end{equation}
where ${\bf{v}}=( -C_{max},  C_{max},  C_{max},  -C_{max})^T$ and above $\bm{x}$ denotes the input of such linear transformation. Thus, $\hat{g}_{j}^{(i)}(\bm{z})\in \mathcal{N} \mathcal{N}_{1,1}\left(4, 3, 8C_{max}^2\right)$ provided that $C_{max}\ge 2$.

Note that $\|T_{C_{max}} f-g\|_\infty \le \epsilon$ if $\|g\|_\infty \le C_{max}$ and $\|f-g\|_\infty \le \epsilon$. Therefore, we have $\hat{g}_{j}^{(i)} \in \mathcal{N} \mathcal{N}_{t_j^{(i)},1}\left(W', L'+2, 8C_{max}^2 \max\{U,1\}\right)$ and 
\begin{align}
\label{eq:nn-approx-error-single}
	\Bigg\|\hat{g}_j^{(i)}\bigg(\frac{\bm{z}+C_{max}}{2C_{max}}\bigg) - \bar{g}^{(i)}_j\bigg(\frac{\bm{z}+C_{max}}{2C_{max}}\bigg)\Bigg\|_\infty \le C_2 (U)^{-\frac{\alpha_{j}^{(i)}}{1+t_{j}^{(i)}}} \le C_2 (U)^{-\gamma^*}\text{ for all } \bm{z} \in \mathcal{D}_j^{(i)}.
\end{align} 

Now we are ready to construct a neural network $f^\dagger$ to approximate $f_0=h^{(l)}_1$. To be specific, our construction proceeds recursively as 
\begin{align*}
 	\hat{h}^{(1)}_j(\bm{x}) = \hat{g}^{(1)}_j\Bigg(\frac{\bm{x}_{\pi(\sum_{\ell=1}^{j-1} t^{(1)}_\ell + 1)}+C_{max}}{2C_{max}}, \ldots , \frac{\bm{x}_{\pi(\sum_{\ell=1}^{j} t^{(1)}_\ell)}+C_{max}}{2C_{max}}\Bigg)
 \end{align*} and
 \begin{align*}
  	\hat{h}^{(i)}_j(\bm{x}) = \hat{g}^{(i)}_j\left( \frac{\hat{h}^{(i-1)}_{\sum_{\ell=1}^{j-1} t^{(i)}_\ell + 1} (\bm{x})+C_{max}}{2C_{max}}, \ldots , \frac{\hat{h}^{(i-1)}_{\sum_{\ell=1}^{j} t^{(i)}_\ell} (\bm{x})+C_{max}}{2C_{max}}\right).
 \end{align*} 
 The corresponding composited network, denoted by $\hat{f} = \hat{g}(\alpha_1 \hat{h}_1(\bm{x}) + \beta_1, \ldots, \alpha_k \hat{h}_k(\bm{x}) + \beta_k)$, is realized by first applying network composition $L_i \circ \hat{h}_i$ for each $i \in \{1,\ldots, k\}$, where $L_i(\bm{x}) = \alpha_i \bm{x} + \beta_i$, followed by network parallelization $(L_1 \circ \hat{h}_1(\bm{x}), \ldots, L_k \circ \hat{h}_k(\bm{x}))$, and then followed by network composition $\hat{g} \circ (L_1 \circ \hat{h}_1(\bm{x}), \ldots, L_k \circ \hat{h}_k(\bm{x}))$. For $i \in \{ 1, \ldots, k\}$, assume the deep ReLU neural network $\hat{h}_i: \mathbb{R}^d\to\mathbb{R}$ has depth $L_{h_i}$ and width $W_{h_i}$, and the deep ReLU neural network $\hat{g}$ has depth $L_g$ and width $W_g$. We conclude that the  network composition $\hat{f}$ has depth $(\max L_{h_i}) + L_g$ and width $(\sum_{i=1}^k W_{h_i}) \lor W_g$.

Based on the recursive construction of neural networks, {we set $f^\dagger$ to be $\hat{h}^{(l)}_1$. Now it suffices to calculate the width, depth and approximation error of $\hat{h}^{(l)}_1$.} These quantities will also be calculated recursively. 
 
\medskip
\noindent {\sc Step 2. Specifying lower bounds of  width and depth and $J(f^\dagger)$.} The goal is to calculate the lower bounds of width and depth of each $\hat{h}^{(i)}_j$ from $i=1$ to $i=l$ and the penalty $J(f^\dagger)$. Let ${W}^{(i)}_j$ and $L^{(i)}_j$ be the lower bounds of   width and depth of the network $\hat{h}^{(i)}_j$. First, by Lemma \ref{lemma_jiaoapp} and the discussion before, for each $j \in \{1, \ldots , M_i\}$, the two lower bounds satisfy
\begin{align*}
	W^{(1)}_{j} =  C_1U^{\frac{2t^{**}+\alpha^{**}}{2t^{**}}}, \quad L^{(1)}_{j} =  2\log_2( t_{max}+\alpha_{max} )+4,\quad J(\hat{h}^{(i)}_j)= 16C_{max}^2 \max\{U,1\}
\end{align*} 
where $(t^{**},\alpha^{**})= \sup_{(\alpha,C,t)\in\mathcal{P}}\frac{\alpha}{t}$.

Now suppose we have already calculated the depth and width for all $\hat{h}^{(i-1)}_j$. Then, based on our discussion of the composited network before, for any given $j\in\{1,\ldots, M_i\}$, the depth and width of $\hat{h}^{i}$ satisfy
\begin{align*}
	& L^{(i)}_j = \max_{j\in P(i,j)}L^{(i-1)}_j +  2\log_2( t_{max}+\alpha_{max} )+2, \qquad W^{(i)}_j = \sum_{j\in P(i,j)} W^{(i-1)}_{j},\\
&  J(\hat{h}^{(i)}_j)=  J(\hat{h}^{(i-1)}_j) 16C_{max}^2 \max\{U,1\}
\end{align*} 
where $P(i,j)=\{\sum_{\ell=1}^{j-1} t_{\ell}^{(i)}+1, \ldots , \sum_{\ell=1}^j t_{\ell}^{(i)}\}$. Using the above recursive calculation, the lower bound of depth of $f^\dagger=\hat{h}^{(l)}_1$ can be written as
\begin{align*}
	\bar{L} = 2l(\log_2( t_{max}+\alpha_{max} )+1),
\end{align*} 
while the lower bound of depth of $f^\dagger=\hat{h}^{(l)}_1$ can be written as \begin{align*}
	\bar{N} = N^{(l)}_1 \le  \underbrace{C_1t_{max}^{l-1}}_{C_3} U^{\frac{2t^{**}+\alpha^{**}}{2t^{**}}}.
\end{align*} 
Meanwhile, the penalty of $f^\dagger$ is $J(f^\dagger)= ( 16C_{max}^2 \max\{U,1\})^l$.

\medskip
\noindent {\sc Step 3. Approximation error.} We claim that \begin{align}
\label{eq:nn-approx-error-est}
	\|\hat{h}^{(i)}_j - h^{(i)}_j\|_\infty \le C_3 ( C\sqrt{{t_{\max}}} + 1)^{i-1} (NL)^{-2\gamma^*}.
\end{align} 
We prove inequality \eqref{eq:nn-approx-error-est} by mathematical induction, starting with the case of $i=1$. By our discussion in Step~1, let $\bm{z} = \big( \bm{x}_{\pi(\sum_{\ell=1}^{j-1} t^{(1)}_\ell + 1)}, \ldots, \bm{x}_{\pi(\sum_{\ell=1}^{j} t^{(1)}_\ell)}\big)$, we have for all $\bm{x}\in [0,1]^d$ that
\begin{align*}
	|\hat{h}^{(1)}_j(\bm{x}) - h^{(1)}_j(\bm{x})| &= \Bigg|\hat{g}^{(1)}_j\bigg(\frac{\bm{z}+C_{max}}{2C_{max}}\bigg) - g_j^{(1)}(\bm{z})\Bigg| \\
	&= \Bigg|\hat{g}^{(1)}_j\bigg(\frac{\bm{z}+C_{max}}{2C_{max}}\bigg) - \bar{g}_j^{(1)} \bigg(\frac{\bm{z}+C_{max}}{2C_{max}} \bigg)\Bigg| \\
	& \leq  C_2 (U)^{-\gamma^*} ,
\end{align*}
where the last step follows from \eqref{eq:nn-approx-error-single}.

Suppose \eqref{eq:nn-approx-error-est} holds for $i-1$ and $j\in \{1,\ldots, M_{i-1}\}$. Write $\bm{z}=\big( h^{(i-1)}_{\sum_{\ell=1}^{j-1} t^{(i)}_\ell + 1} (\bm{x}), \ldots, h^{(i-1)}_{\sum_{\ell=1}^{j} t^{(i)}_\ell} (\bm{x})\big)$ and $\hat{\bm{z}} = \big( \hat{h}^{(i-1)}_{\sum_{\ell=1}^{j-1} t^{(i)}_\ell + 1} (\bm{x}), \ldots, \hat{h}^{(i-1)}_{\sum_{\ell=1}^{j} t^{(i)}_\ell} (\bm{x})\big)$ for $\bm{x}\in[0,1]^d$. We have
\begin{align*}
	|\hat{h}^{(i)}_j(\bm{x}) - h^{(i)}_j(\bm{x})| &= \Bigg|\hat{g}^{(i)}_j\bigg(\frac{\hat{\bm{z}}+C_{max}}{2C_{max}}\bigg)-g_j^{(i)}(\bm{z})\Bigg| \\
	&\leq \Bigg|\hat{g}^{(i)}_j\bigg(\frac{\hat{\bm{z}}+C_{max}}{2C_{max}}\bigg)-g_j^{(i)}(\hat{\bm{z}})\Bigg| + |g_j^{(i)}(\hat{\bm{z}})-g_j^{(i)}(\bm{z})|.
\end{align*}
Together, \eqref{eq:nn-approx-error-single} and the fact that $\hat{\bm{z}} \in [-U, U]^{t^{(i)}_j}$ imply
\begin{align}
\label{eq:nn-approx-est1}
\Bigg|\hat{g}^{(i)}_j\bigg(\frac{\hat{\bm{z}}+C_{max}}{2C_{max}}\bigg)-g_j^{(i)}(\hat{\bm{z}})\Bigg| = \Bigg|\hat{g}^{(i)}_j\bigg(\frac{\hat{\bm{z}}+C_{max}}{2C_{max}}\bigg)-\bar{g}_j^{(i)}\bigg(\frac{\hat{\bm{z}}+C_{max}}{2C_{max}}\bigg)\Bigg|	\le C_2 (U)^{-\gamma^*}.
\end{align}
Since $g^{(i)}_j$ is at least $C_{max}$-Lipschitz (see its definition in \eqref{Holder}), we further have
\begin{align*}
|g_j^{(i)}(\hat{\bm{z}})-g_j^{(i)}(\bm{z})| &\le C_{max} \|\hat{\bm{z}} - \bm{z}\|_2 \\
	&\le C_{max}\sqrt{{t_{\max}}} \|\hat{\bm{z}} - \bm{z}\|_\infty \\
	& \leq  C_{max}\sqrt{{t_{\max}}} (1 + C_{max}\sqrt{{t_{\max}}})^{i-2} C_3 (U)^{-\gamma^*} ,
\end{align*} 
where the last inequality follows from the induction. Putting together the pieces, we obtain
\begin{align*}
	|\hat{h}^{(i)}_j(\bm{x}) - h^{(i)}_j(\bm{x})|
	&\leq \Bigg|\hat{g}^{(i)}_j\bigg(\frac{\hat{\bm{z}}+C_{max}}{2C_{max}}\bigg)-g_j^{(i)}(\hat{\bm{z}})\Bigg| + |g_j^{(i)}(\hat{\bm{z}})-g_j^{(i)}(\bm{z})| \\
	&\le C_3 (U)^{-\gamma^*} + C_3 C\sqrt{{t_{\max}}} (1 + C_{max}\sqrt{{t_{\max}}})^{i-2} (U)^{-\gamma^*} \\
	&\le C_3 (1 + C\sqrt{{t_{\max}}})^{i-1} (U)^{-\gamma^*}.
\end{align*}
Finally, we conclude that
\begin{align*}
	\|f^\dagger - f_0\|_\infty = \|\hat{h}^{(l)}_1 - h^{(l)}_1\|_\infty \le \underbrace{C_3 ( C_{max}\sqrt{{t_{\max}}} + 1)^{l-1}}_{c_5} (U)^{-\gamma^*} ,
\end{align*}
as claimed. \hfill\(\Box\)\\

\subsection{Proofs of Theorem \ref{Theorem1}-\ref{Theorem2}}

\textit{Proof of Theorem \ref{Theorem1}.}
First, we fix $\bm{X}_1,\ldots, \bm{X}_n$. Define a constrained neural network space indexed by $k$:
\begin{equation}\label{ii}
 \N\N_{d,1}(W_k,L_k,U_n):=\{g\in   \N\N(W_k,L_k): J(g)\le U_n\}.
\end{equation}
with some $U_n>0$ related to $n$. Let $m_k^*\in \mathcal{N} \mathcal{N}_{d, 1}\left(W_k, L_k, U_n\right)$ be the network given in Theorem \ref{reg_multi_approx} satisfying 
 $$
    \|m-m_k^*\|_\infty \lesssim U_n^{-\frac{\gamma^*}{l}}=U_n^{-\beta_1}.
 $$
 If we use unconstrained coefficients of network class $\mathcal{N} \mathcal{N}_{d, 1}\left(W_k, L_k\right)$, which is larger than $\mathcal{N} \mathcal{N}_{d, 1}\left(W_k, L_k,U_n\right)$, to approximate $m$, Proposition 3.4 in \cite{fan2024noise} tells us
 $$
     \|m-m_k^*\|_\infty \lesssim (L_kW_k)^{-\alpha_1}.
 $$
 In conclusion, 
 \begin{equation}\label{n_approx}
   \|m-m_k^*\|_\infty\le \max\{c(L_kW_k)^{-\alpha_1}, c'U_n^{-\beta_1}\}.
 \end{equation}

 Since $\hat{m}$ is the minimizer of the empirical risk function, we know
\begin{equation}\label{ULll8}
\frac{1}{n}\sum_{i=1}^{n}(Y_i-\hat{m}(\bm{X}_i))^2+\lambda_n J(\hat{m})\le \frac{1}{n}\sum_{i=1}^{n}(Y_i-g_1(\bm{X}_i))^2+\lambda_n J(m_k^*).
\end{equation}
 In other words,
$$
\frac{1}{n}\sum_{i=1}^{n}(Y_i-m(\bm{X}_i)+m(\bm{X}_i)+\hat{m}(\bm{X}_i))^2+\lambda_n J(\hat{m})\le \frac{1}{n}\sum_{i=1}^{n}(Y_i-m(\bm{X}_i)+m(\bm{X}_i)-m_k^*(\bm{X}_i))^2+\lambda_n J(m_k^*).
$$
with probability equal to $1$. Simplify above inequality. Then, we get
\begin{equation}\label{dsajhbJK}
 \|\hat{m}-m\|_n^2+\lambda_nJ(\hat{m})\le \frac{2}{n}\sum_{i=1}^{n}{\varepsilon_i(\hat{m}(\bm{X}_i)-m_k^*(\bm{X}_i))}+\|m-m_k^*\|_n^2+\lambda_nJ(m_k^*).
\end{equation}

Now, we suppose the event $A_n:=\{\max_{1\le i\le n}{|Y_i|}\le \ln n\}$ happens. Set $m_k^*=0$ in \eqref{ULll8} in temporary, it can be known that
\begin{equation}\label{jnasfbdhk}
  \frac{1}{n}\sum_{i=1}^{n}(Y_i-\hat{m}(\bm{X}_i))^2+\lambda_n J(\hat{m}) \le \ln^2 n.
\end{equation}
 According  to \eqref{jnasfbdhk}, $\hat{m}\in \mathcal{N} \mathcal{N}_{d, 1}\left(W_k, L_k, B_n\right)$ with $B_n=O(\frac{\ln n^2}{\lambda_n})$. For any network $f\in\mathcal{N} \mathcal{N}_{d, 1}\left(W_k, L_k, B_n\right)$, it is known $f-m_k^*\in \mathcal{N} \mathcal{N}_{d, 1}\left(2W_k, L_k, B_n+U_n\right)$ by (iv) in Proposition \ref{Pro_largenet}. Now, construct another network space
$$
\mathcal{G}_{\delta}:=\left\{ f-m_k^*: J(g-m_k^*)\le \delta , f-m_k^*\in \mathcal{N} \mathcal{N}_{d, 1}\left(2W_k, L_k, B_n+U_n\right)\right\}.
$$
with $\delta\in (0, B_n+U_n)$ and consider the corresponding Gaussian process below
$$
\mathcal{G}_{\delta}\to\R:\ \ \ \ g\in\mathcal{G}_{\delta}\mapsto\frac{1}{\sqrt{n}}\sum_{i=1}^{n}\frac{\varepsilon_i}{\sigma}g(\bm{X}_i).
$$
Note that $\mathcal{G}_{\delta}$ is indexed by finite parameters and each neural network in $\mathcal{G}_{\delta}$ is continuous w.r.t. these parameters.  Thus it is a separable space w.r.t. the supremum norm. Namely, for any $\eta>0$, there is a series of functions $\{g_j\}_{j=1}^\infty\subseteq \mathcal{G}_{\delta}$ such that for any $g\in \mathcal{G}_{\delta} $, we can find $j^*\in\Z$:
$$
  \sup_{\bm{x}\in [0,1]^d}{\left|g(\bm{x})-g_{j^*}(\bm{x})\right|}\le \eta.
$$
The above inequality leads that the defined Gaussian process is also separable. Since (v) in Proposition \ref{Pro_largenet} holds, the application of Borell-Sudakov-Tsirelson concentration inequality (see Theorem 2.5.8 in \cite{gin2015mathematical}) implies
\begin{equation}\label{sfADGHJFGBH}
  \P\left( \sup_{g\in \mathcal{G}_{\delta}}{\left|\frac{1}{n}\sum_{i=1}^{n}\varepsilon_i g(\bm{X}_i)\right|}\ge \E\sup_{g\in \mathcal{G}_{\delta}}{\left|\frac{1}{n}\sum_{i=1}^{n}\varepsilon_i g(\bm{X}_i)\right|}+2\delta r\Big| \bm{X}_1,\ldots,\bm{X}_n\right)\le e^{-\frac{nr^2}{2\sigma^2}}.
\end{equation}
Let $\delta_j=2^{j-1}\sigma/\sqrt{n}$, $j=1,2,\ldots,\lfloor \log_2((B_n+U_n)\sqrt{n}/\sigma)\rfloor+1$. From \eqref{jnasfbdhk}, we know $\hat{m}-m_k^*\in \mathcal{G}_{\delta_{j^*}}$ a.s. for some $j^*$, where $j^*$ is a random index. Thus, we have the following probability bound
\begin{align}\label{svdfhb}
 & \P\left( \bigcup_{j=1}^{\lfloor \log_2(B_n\sqrt{n}/\sigma)\rfloor+1}\left\{\sup_{g\in \mathcal{G}_{\delta_j}}{\left|\frac{1}{n}\sum_{i=1}^{n}\varepsilon_i g(\bm{X}_i)\right|}\ge \E\sup_{g\in \mathcal{G}_{\delta_j}}{\left|\frac{1}{n}\sum_{i=1}^{n}\varepsilon_i g(\bm{X}_i)\right|}+2\delta_j r\right\}\Big| \bm{X}_1,\ldots,\bm{X}_n\right) \nonumber\\
  &\le \lfloor \log_2((B_n+U_n)\sqrt{n}/\sigma)+1\rfloor\cdot e^{-\frac{nr^2}{2\sigma^2}},
\end{align}
whose RHS  does not depend on any $\delta_j,j=1,2,\ldots$.

For any $J(\hat{m}-m_k^*)$, we can find $j^*$ satisfying $\delta_{j^*}\le J(\hat{m}-m_k^*)< \delta_{j^*+1}$. Replace $r$ in \eqref{svdfhb} by $\sigma r \sqrt{\ln n/n}$.
Then, with probability larger than $1-\lfloor \log_2((B_n+U_n)\sqrt{n}/\sigma)+1\rfloor\cdot n^{-r}-\P(A_n)$,
\begin{equation}
     \frac{1}{n}\sum_{i=1}^{n}\varepsilon_i (\hat{m}(\bm{X}_i)-m_k^*(\bm{X}_i))  \le H(2J(\hat{m}-m_k^*))+4J(\hat{m}-m_k^*)\cdot \sigma r \sqrt{\frac{\ln n}{n}},     \label{hbds}
\end{equation}
where  for any $\delta>0$ we  define  the function
 $$H(\delta):=\E\sup_{g\in \mathcal{N} \mathcal{N}_{d, 1}\left(2W_k, L_k, \delta\right)}{\left| \frac{1}{n} \sum_{i=1}^{n} \varepsilon_i g(\bm{X}_i)\right|}.$$
From Proposition \ref{pro_relu_multilayer}, we know
\begin{equation}\label{afsdghbjhjgasf}
H(\delta)\lesssim \delta  \sqrt{\frac{L_k}{n}}.
\end{equation}
Therefore, the combination of \eqref{hbds} and \eqref{afsdghbjhjgasf} implies
\begin{equation}\label{fsdjyg}
  \frac{1}{n}\sum_{i=1}^{n}\varepsilon_i (\hat{m}(\bm{X}_i)-m_k^*(\bm{X}_i)) \le c\cdot J(\hat{m}-m_k^*) \sqrt{\frac{L_k}{n}},
\end{equation}
where $c>0$  is a universal constant.

Then, the combination of \eqref{n_approx}, \eqref{dsajhbJK} and \eqref{fsdjyg} and Proposition \ref{Gaussian_complexity} implies 
\begin{align}\label{BGVJH}
   \|\hat{m}-m\|_n^2+\lambda_nJ(\hat{m})&\le \frac{2}{n}\sum_{i=1}^{n}{\varepsilon_i(\hat{m}(\bm{X}_i)-m_k^*(\bm{X}_i))}+\|m-m_k^*\|_n^2+\lambda_n J(m_k^*)\nonumber\\
   &\le c\cdot J(\hat{m}-m_k^*) \sqrt{\frac{L_k \ln n}{n}}+\max\{c(L_kW_k)^{-2\alpha_1}, c'U_n^{-2\beta_1}\} + \lambda_n U_n
\end{align}
holds with probability larger than $1-\lfloor \log_2((B_n+U_n)\sqrt{n}/\sigma)+1\rfloor\cdot n^{-r} -\P(A_n)$, where $\beta_1:= \min_{(\alpha, C, t) \in \mathcal{P}}\left\{ \frac{\alpha}{t+1}\right\}/l$ and $\alpha_1= \min_{(\alpha, C, t) \in \mathcal{P}}\left\{ \frac{2\alpha}{t}\right\}$.  From (iv) in Proposition \ref{Pro_largenet}, it is known that $J(\hat{m}-m_k^*)\le J(\hat{m})+J(m_k^*)$.  At this point, we take $\lambda_n= 2c \sqrt{\frac{L_k\ln n}{n}}$ and $U_n=n^{\frac{1}{2(2\beta_1+1)}}$. Then, \eqref{BGVJH} implies
\begin{equation*}
    \|\hat{m}-m\|_n^2+\frac{1}{2}\lambda_nJ(\hat{m})\le  c\max\{(L_kW_k)^{-2\alpha_1}, (n/L_k)^{-\frac{\beta_1}{2\beta_1+1} }\}.
\end{equation*}

On the other hand,
\begin{align}
		\P\left(\max_{1\leq i\leq n}|\varepsilon_i|> c\cdot\ln{n}\right)=&1- \P\left(\max_{1\leq i\leq n}|\varepsilon_i|\leq c\cdot\ln{n}\right)\nonumber\\
		=&1-\left[\P(|\varepsilon_1|\leq c\cdot\ln{n})\right]^n
		\leq 1-(1-c\cdot e^{-c\cdot \ln^2{n}})^n\nonumber \\
		=& 1-e^{n\cdot \ln(1-c\cdot e^{-c\cdot \ln^2{n}})} \nonumber\\
		\leq & -n\cdot \ln(1-c\cdot e^{-c\cdot \ln^2{n}})\label{opipo}\\
		\leq & c\cdot n\cdot e^{-c\cdot \ln^2{n}}\leq c\cdot n^{-r},\label{duhsihrdksh}
	\end{align}
where $\eqref{opipo}$ is obtained from the basic inequality $1+v\le e^{v}, v\in\mathbb{R}$; and \eqref{duhsihrdksh} is due to the fact $\lim_{v\to 0}{\frac{\ln(1+v)}{v}}=1$. Therefore, the combination of  \eqref{BGVJH} and \eqref{duhsihrdksh} shows that
\begin{equation*}
    \|\hat{m}-m\|_n^2\le  c\max\{(L_kW_k)^{-2\alpha_1}, (n/L_k)^{-\frac{\beta_1}{2\beta_1+1} }\}.
\end{equation*}
holds with probability larger than $1-c\cdot n^{-r}$ and $r>0$ is a large number. Since the above inequality holds for any fixed $(\bm{X}_1,\ldots, \bm{X}_n)$,  inequality \eqref{ahbjsdhj1} holds with the same probability by the law of total probability.

Next, we prove the upper bound in \eqref{ahbjsdhj1} is also true for $\E( \|\hat{m}-m\|_n^2)$. By  calculations, we have
\begin{align*}
   \left|\frac{1}{n}\sum_{i=1}^{n}\varepsilon_i (\hat{m}(\bm{X}_i)-m_k^*(\bm{X}_i))\right| & \leq \left( \frac{1}{n}\sum_{i=1}^{n}\varepsilon_i^2\right)^{\frac{1}{2}}\left( \frac{1}{n}\sum_{i=1}^{n}(\hat{m}(\bm{X}_i)-m_k^*(\bm{X}_i))^2\right)^{\frac{1}{2}} \\
   &= \left( \frac{1}{n}\sum_{i=1}^{n}\varepsilon_i^2\right)^{\frac{1}{2}} \cdot \|\hat{m}-m_k^*\|_n \\
   &= \left( \frac{1}{n}\sum_{i=1}^{n}\varepsilon_i^2\right)^{\frac{1}{2}} \cdot \|\hat{m}-m+m-m_k^*\|_n \\
   &\le  \left( \frac{1}{n}\sum_{i=1}^{n}\varepsilon_i^2\right)^{\frac{1}{2}} \cdot (\|\hat{m}-m\|+\|m-m_k^*\|_n) \\
   &\le \left(\frac{1}{n}\sum_{i=1}^{n}\varepsilon_i^2+ \frac{1}{4}\|\hat{m}-m\|_n^2 \right)+ \left(\frac{1}{n}\sum_{i=1}^{n}\varepsilon_i^2+ \frac{1}{4}\|m_k^*-m\|_n^2 \right)\\
   &= \frac{2}{n}\sum_{i=1}^{n}\varepsilon_i^2+\frac{1}{4}\|\hat{m}-m\|_n^2+\frac{1}{4}\|m_k^*-m\|_n^2,
\end{align*}
where in the last two line we use the basic inequality $ab\le a^2+\frac{1}{4}b^2$. Substitute the above inequality to \eqref{dsajhbJK}. Then, we have
$$
\|\hat{m}-m\|_n^2+\lambda_n J(\hat{m}) \le \frac{4}{n}\sum_{i=1}^{n}\varepsilon_i^2+ \frac{1}{2}\|\hat{m}_n-m\|_n^2+\frac{1}{2}\|m_k^*-m\|_n^2\ \ a.s..
$$
Namely,
\begin{equation}\label{hbgjsdfjhbk}
   \frac{1}{2}\|\hat{m}-m\|_n^2+\lambda_n J(\hat{m}) \le \frac{4}{n}\sum_{i=1}^{n}\varepsilon_i^2+\frac{1}{2}\|m_k^*-m\|_n^2\ \ a.s..
\end{equation}
Define the event
$$
B_n:=\left\{    \|\hat{m}-m\|_n^2+\frac{1}{2}\lambda_n J(\hat{m})\le  c\max\{k^{-2\alpha_1}, (n/L_k)^{-\frac{\beta_1}{2\beta_1+1} }\} \right\}.
$$
Let $\mathbb{I}(B_n)$ be the indicator function of the event $B_n$. Then, we can bound the above expectation  by using the following decomposition.
\begin{align}
  \E\left(  \|\hat{m}-m\|_n^2+\frac{1}{2}\lambda_n J(\hat{m}) \right) & \le  \E\left(  (\|\hat{m}-m\|_n^2+\frac{1}{2} J(\hat{m}))\mathbb{I}(B_n) \right)\nonumber \\
   &  + \E\left(  (\|\hat{m}-m\|_n^2+\frac{1}{2}\lambda_nJ(\hat{m}))\mathbb{I}(B_n^c) \right) \nonumber\\
   & := I+II. \label{;asdhjfghbasjk}
\end{align}
The first part $I$ can be bounded by using result in \eqref{ahbjsdhj1}. Namely,
\begin{equation}\label{sdfbjkbfk}
  I\le c\max\{(L_kW_k)^{-2\alpha_1}, (n/L_k)^{-\frac{\beta_1}{2\beta_1+1} }\}.
\end{equation}
On the other hand, we know from the last paragraph that $\P(B_n)\ge 1-c\cdot n^{-r}$. By using this probability bound, we  use \eqref{hbgjsdfjhbk} to bound Part $II$ below.
\begin{align}
  II &\le \E\left( \left( \frac{8}{n}\sum_{i=1}^{n}\varepsilon_i^2 + \|m-m_k^*\|_n^2 \right)\mathbb{I}(B_n^c) \right) \nonumber\\
  &\le \E\left( \left( \frac{8}{n}\sum_{i=1}^{n}\varepsilon_i^2 + ck^{-\alpha} \right)\mathbb{I}(B_n^c) \right) \nonumber\\
  &\le 8\E\left( \left(\frac{1}{n}\sum_{i=1}^{n}{\varepsilon_i^2} \right)\mathbb{I}(B_n^c)\right)+\P(B_n^c)\nonumber\\
  &\le 8 \sqrt{\E \left( \left(\frac{1}{n}\sum_{i=1}^{n}{\varepsilon_i^2} \right)\right)^2}\cdot \sqrt{\P(B_n^c)}+\P(B_n^c)\nonumber\\
  &\le 24c\cdot n^{-\frac{r}{2}}+c\cdot n^{-r},\label{sdfgjhb}
\end{align}
where $r$ is a large number and $r\ge 2$. Finally, the combination of \eqref{;asdhjfghbasjk}, \eqref{sdfbjkbfk} and  \eqref{sdfgjhb} gives us
$$
  \E\left( \|\hat{m}-m\|_n^2+\frac{1}{2}\lambda_n J(\hat{m})\right)\le  c\max\{(L_kW_k)^{-2\alpha_1}, (n/L_k)^{-\frac{\beta_1}{2\beta_1+1} }\} .
$$
This completes the proof.  \hfill\(\Box\)\\

\noindent \textit{Proof of Theorem \ref{Theorem2}.} The proof is similar to Theorem \ref{Theorem_Huber}. \hfill\(\Box\)\\

\subsection{Proof of Proposition  \ref{pro_RF}.}
\begin{figure}[ht]
    \includegraphics[height=0.25\textheight]{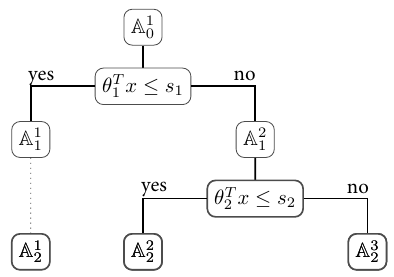}
    \centering
    \caption{This ODT  has two layers and three leaves denoted by $\mathbb{A}_2^1, \mathbb{A}_2^2, \mathbb{A}_2^3$. Note that $\mathbb{A}_1^1$ is not partitioned anymore and thus $\mathbb{A}_1^1=\mathbb{A}_2^1$. Meanwhile, it can be seen that $\mathbb{A}_2^1=\{x:\theta_1^Tx\le s_1\}$, $\mathbb{A}_2^2=\{x:\theta_1^Tx> s_1\}\cap\{x:\theta_2^Tx\le s_2\}$ and $\mathbb{A}_2^3=\{x:\theta_1^Tx> s_1\}\cap\{x:\theta_2^Tx> s_2\}$.}
    \label{ODT11fig2}
\end{figure}

At the beginning, we analyze the first tree $T_{\D_n^1}$. Let $\mathbb{A}_1,\mathbb{A}_2,\ldots,\mathbb{A}_{a_n}$ be $a_n$ leaves of $T_{\D_n^1}$. Then, we know each $\mathbb{A}_j$ is generated after performing $\mathcal{C}_j\in\Z^+$ cuts in $[0,1]^d$ with $\mathcal{C}_j\le a_n-1$. Since each tree partition corresponds with a direction $\theta\in\R^p$ and a threshold $s\in\R$, we can denote each $\mathbb{A}_j$ by 
        $$
          \mathbb{A}_j=\tilde  A_{j.1}\cap\cdots\cap  \tilde A_{j.\mathcal{C}_j},
        $$        
        where $\tilde A_{j.\ell} = \{x\in [0,1]^p: \theta_{j,\ell}^Tx > s_\ell\} $  or $\tilde A_{j.\ell} = \{x\in [0,1]^p: \theta_{j,\ell}^Tx \le s_\ell\} $ for each $\ell=1,2,\ldots,\mathcal{C}_j$
and $\theta_{j,\ell}\in\R^p, s_\ell\in\R$. Note that $\theta_{j,\ell}$ only consists of $d-1$ numbers of $0$ and a number of  $1$.  In Figure \ref{ODT11fig2}, we give an example of such representation of tree leaves.
    
Meanwhile, note that the following equation holds 
        \begin{equation}\label{asbdBerr}
         \mathbb{I}(x\in  \mathbb{A}_j)= \sigma_0\left(\sum_{\ell=1}^{\mathcal{C}_j}\sigma_0(s_\ell-\theta_{j,\ell}^Tx)-\mathcal{C}_j\right)
       \end{equation}
        if 
        \begin{equation}\label{asbdBerr2}
            \mathbb{A}_j=\{x\in [0,1]^p: \theta_{j,1}^Tx\le s_1\}\cap\cdots\cap  \{x\in [0,1]^p:\theta_{j,\mathcal{C}_j}^Tx\le s_{\mathcal{C}_j}\}.
        \end{equation}

Since $\mathbb{I}(\{x\in[0,1]^p: \theta^\top x>s\})=\sigma_0(0)-\sigma_0(s-\theta^\top x)$, we can assume \eqref{asbdBerr} holds without loss of generality. This is because that if $\theta_{j,\ell}^\top x>s$ we only need to replace $\sigma_0(s_\ell-\theta_{j,\ell}^Tx)$ by $\sigma_0(0)-\sigma_0(s_\ell-\theta_{j,\ell}^\top x)$ in \eqref{asbdBerr}. Recall that $\bar{Y}_{\mathbb{A}_j}$ is the constant estimator in the region
$\mathbb{A}_j$. Therefore, the first tree in the boosting process is equal to
        $$
            \sum_{j=1}^{{a_n}}{\bar{Y}_{\mathbb{A}_j} \sigma_0\left(\sum_{\ell=1}^{\mathcal{C}_j}\sigma_0(s_\ell-\theta_{j,\ell}^Tx)-\mathcal{C}_j\right)},
        $$
        which is a neural network with three layers. Therefore, $T_{\D_n^1}$ can be regarded as a neural network with  $\sum_{j=1}^{a_n}{\mathcal{C}_j}$ neurons in the first hidden layer and  $a_n$ neurons in the second hidden layer. 
        
        Since feed-forward neural networks have additive structures, we know 
       RF defined in \eqref{8} is  in the following  neural network class
        $$
         \left\{ \sum_{i=1}^{B_n}\sum_{j=1}^{a_n} a_{i,j}\sigma_0\left(  \sum_{\ell=1}^{a_n}\sigma_0(\theta_{i,j,\ell}^Tx+s_{i,j,\ell})b_{i,j,\ell}+v_{i,j}\right): a_{i,j},b_{i,j,\ell},s_{i,j,\ell},v_{i,j}\in\R,\theta_{i,j,\ell}\in\R^d\right\},
        $$
        which has $B_n{a_n}^2(d+1)$ parameters ($\theta_{i,j,\ell},s_{i,j,\ell}$) in the first  hidden layer and  $B_n{a_n}({a_n}+1)$ parameters ($b_{i,j,\ell},v_{i,j}$) in the second  hidden layer  and  $B_n{a_n}$ parameters ($a_{i,j}$) in the final  hidden layer. This completes the proof.
\hfill\(\Box\)\\

\subsection{Proofs of Theorem \ref{Theorem_Huber}-\ref{Theorem_quantile}}

First, we need a lemma below. 

\begin{lemma}\label{lemma_sbdj}
 Let  $\ell(\cdot)$ be a Lipshitz loss function satisfying
 $
   |\ell(\bm{x}_1)-\ell(\bm{x}_2)|\le F_n\|\bm{x}_1-\bm{x}_2\|_2,\ \forall \bm{x}_1,\bm{x}_2\in\R^d.
 $
For any function $f$, give its empirical risk and population risk by
$$
\hat{R}_{\ell}(f)=\frac{1}{n}\sum_{i=1}^{n}{\ell(Y_i-f(\bm{X}_i))},\ \ R_{\ell}(f):=\E( {\ell(Y-f(\bm{X}))}).
$$
Then, we define the regularized network estimator by
$$
\hat{m}_{\ell,n}\in \left\{ g\in \hat{R}_{\ell}(f) : \hat{R}_{\tau}(g)+\lambda_nJ(g)\le \inf_{f\in \N\N^M(W_k,L_k)}{\left(\hat{R}_{\ell}(f)+\lambda_nJ(f)\right)}+\delta_{opt}^2 \right\},
$$
where $\delta_{opt}^2>0 $ and the penalty $J(\cdot)$ is defined in \eqref{pingan}. Suppose $\E|Y|^p<\infty$ for some $p\ge 1$ and $\bm{X}\in [0,1]^d$. For any $f_k^{*}\in\mathcal{N}\N^M(L_k,W_k)$, the excess risk satisfies
\begin{equation}\label{*ghh}
   R_\ell(\hat{m}_{\ell,n})-R_\ell(m)+\lambda_n  J(\hat{m}_{\ell,n})=O_p\left(\underbrace{ \delta_{opt}^2}_{optimization\ error}+ \underbrace{R(f_k^{*})-R(m)}_{approximation\ error}+ \underbrace{\frac{J(f_k^{*})+F_n}{\sqrt{n/L_k}}}_{sample\ error}\right)
\end{equation}
with $\lambda_n\asymp F_n \sqrt{\frac{L_k}{n}}$.
\end{lemma}

\begin{remark}
 We call the last term the sample error because this error always decreases to zero as the sample size $n\to\infty$.
\end{remark}

\begin{proof}

Our analysis is based on the following risk decomposition.
\begin{equation}\label{ews}
  \begin{aligned}
  R(\hat{m}_{\ell,n})-R(m)+\lambda_n J(\hat{m}_{\ell,n}) &:= \underbrace{R(\hat{m}_{\ell,n})-\hat{R}(\hat{m}_{\ell,n})}_{\text{I: stochastic error}} \\
  & + \underbrace{\hat{R}(\hat{m}_{\ell,n})+\lambda_n J(\hat{m}_{\ell,n})-\hat{R}(m_{k}^*)-\lambda_nJ(f_k^{*})}_{\text{II: optimization error}} \\
   & + \underbrace{\hat{R}(f_k^{*})-R(f_k^{*})}_{\text{III}}\\
   &+ \underbrace{R(f_{k,M}^*)-R(m)+\lambda_nJ(f_k^{*})}_{\text{IV: approximation error}},
\end{aligned}
\end{equation}
where $f_{k}^*\in \mathcal{N}\N^M(L_k,W_k)$ is a function used to approximate $m(\bm{x})$. In fact, $R(f_{k}^*)-R(m)$ in Part IV is the commonly defined approximation error. With a slight abuse of term, we also call Part IV the approximation error in this proof.

\medskip
\noindent {\sc Analysis of Part I:} For large neural network estimators, the analysis of generalization error is the key part. Define $\N\N^M(W_k,L_k,\delta):=\{g\in \N\N^M(W_k,L_k): J(g)\le \delta\}$. Since $0\in \N\N^M(W_k,L_k)$, from the definition of $\hat{m}_{\ell,n}$ it is known that
\begin{equation}\label{kusk}
  J(\hat{m}_{\ell,n})\le \frac{1}{\lambda_n}\left( \frac{1}{n}\sum_{i=1}^{n}{\ell(Y_i)}+\delta_{opt}^2\right).
\end{equation}
In order to bound the magnitude of $ J(\hat{m}_{\ell,n})$, we need to establish the concentration inequality of $\frac{1}{n}\sum_{i=1}^{n}{\ell(Y_i)}$. Here, we consider the Markov inequality since $Y$ has the $p$-th moment only. For any $\varepsilon>0$,

\begin{equation}\label{k;ko}
  \P\left( \left| \frac{1}{n}\sum_{i=1}^{n}\left(  \ell(Y_i)-\E(\ell(Y))\right)  \right|\ge \varepsilon\right)\le \frac{\E\left| \frac{1}{n}\sum_{i=1}^{n}\left(  \ell(Y_i)-\E(\ell(Y))\right)\right|^p}{\varepsilon^p}.
\end{equation}
Let $Z_i:= \ell(Y_i)-\E(\ell(Y))$. When $p\ge 2$, from Zygmund inequality 
\begin{equation}\label{lkl}
  \E\left|\sum_{i=1}^{n}Z_i\right|^p\le c_p\left( \left(\sum_{i=1}^{n}\E(Z_k^2)\right)^{\frac{p}{2}}+\sum_{i=1}^{n}\E|Z_i|^p \right).
\end{equation}
By the Lipshitz property of Huber loss,  
\begin{align*}
  \E(H(Y_i)^2) & \le \E(F_n Y_i)^2 \le F_n^2\E(Y_i^2) \\
   \E(H(Y_i)^p) & \le \E|F_n Y_i|^p \le F_n^p\E|Y_i|^p.
\end{align*}
When $p\in [1,2)$, from Chatterji inequality, we have
\begin{equation}\label{iosjfd}
  \E|\sum_{i=1}^{n}{Z_i}|^p\le 2^{2-p} \sum_{i=1}^{n} \E|Z_i|^p.
\end{equation}
The combination of \eqref{k;ko}, \eqref{lkl} and \eqref{iosjfd} implies that for any $t_n>0$,
\begin{equation}\label{8811}
    \P\left( \left| \frac{1}{n}\sum_{i=1}^{n}\left(  \ell(Y_i)-\E(\ell(Y))\right)  \right|\ge t_n\right)\le \frac{2\E|Y|^p F_n^p}{n^{\frac{p}{2}}t_n^p}.
\end{equation}
Therefore, with the probability larger than $1-p_{1}$, 
\begin{equation*}
  \frac{1}{n}\sum_{i=1}^{n}{\ell(Y_i)}\le 1+\E(\ell(Y))\le 1+\E(F_n|Y|\I(|Y|>F_n))\le 1+ \tau_nv_p^{\frac{1}{p}},
\end{equation*}
where $p_{v}:= {2\E|Y|^p F_n^p n^{-\frac{p}{2}}v^{-p}}, v>0$. When $\lambda_n\asymp n^{-\frac{1}{2}}$ and $F_n=o(n)$ and $\delta_{opt}=o(1)$, the above inequality and \eqref{kusk} implies
\begin{equation}\label{oik}
   J(\hat{m}_{\ell,n}) \lesssim n^2
\end{equation}
with the probability larger than $1-p_{1}$. Let $b_i, i=1,\ldots,n$ be i.i.d. Rademacher variables with $\P(b_i=\pm 1)=\frac{1}{2}$. 
At this step, we decompose Part I as follows.
\begin{equation}\label{khjbkj}
\begin{aligned}
  I= &\E(\ell(Y-\hat{m}_{\ell,n}(\bm{X})))-\frac{1}{n} \sum_{i=1}^{n} \hat{m}_{\ell,n}(Y_i-\hat{m}_{\ell,n}(\bm{X}_i))\\
  &= \E(\ell(Y-\hat{m}_{\ell,n}(\bm{X}))-\ell(Y))-\frac{1}{n}\sum_{i=1}^{n}\left( \ell(Y_i-\hat{m}_{\ell,n}(\bm{X}_i))-\ell(Y_i)\right)\\
  &+ \frac{1}{n}\sum_{i=1}^{n}(\ell(Y_i)-\E(\ell(Y))).
\end{aligned}
\end{equation}
This decomposition implies we need to analyze the empirical process $|\frac{1}{n}\sum_{i=1}^{n}(U(f)-\E(U(f))|$, where $U(f):=\ell(Y-f(\bm{X}))-\ell(Y)$ and $f\in \N\N^M(W_k,L_k)$. According to the Lipshitz property of Huber loss and (iv) in Proposition \ref{Pro_largenet},  $|U(f)|\le F_n\|f\|_\infty\le F_n J(f)$ for any $f\in \N\N^M(W_k,L_k)$.  Thus, for any $\delta>0$ and $r>0$, the Micdonald inequality tells us that
\begin{equation}\label{bnf}
  \P\left(\sup_{f\in \N\N^M(W_k,L_k,\delta)}{|(\P_n-\P)U(f)|}\ge \E\sup_{f\in \N\N^M(W_k,L_k,\delta)}{|(\P_n-\P)U(f)|}+2\delta F_n r \right)\le e^{-\frac{nr^2}{2}}.
\end{equation}
From \eqref{oik}, the upper bound of $J(\hat{m}_{\ell,n})$ is $n^2$ with probability larger than $1-p_{t_n}$. Set $B_n=n^2$ and $\delta_j=2^{j-1}/\sqrt{n}, j=1,2,\ldots, \lfloor \log_2(B_n\sqrt{n})\rfloor+1$. Thus, from \eqref{bnf}, the  probability of  below  union sets  holds.
\begin{equation}\label{s12}
\begin{aligned}
   &\P\left( \bigcup_{j=1}^{\lfloor \log_2(B_n\sqrt{n})\rfloor+1} \left\{ \sup_{f\in \N_k^M(\delta)}{|(\P_n-\P)U(f)|}\ge \E\sup_{f\in \N_k^M(\delta)}{|(\P_n-\P)U(f)|}+2\delta F_n r\right\} \right)\\
   &\le \lfloor \log_2(B_n\sqrt{n})+1\rfloor e^{-\frac{nr^2}{2}}.
\end{aligned}
\end{equation}
For any $\hat{m}_{\ell,n}$, there is $j^*$ such that $J(\hat{m}_{\ell,n})\in[\delta_{j^*}, \delta_{j^*+1}]$. Replace $r$ in \eqref{s12} by $\sqrt{\frac{\ln n}{n}}\cdot r$. With the probability larger than $1-\lfloor \log_2(B_n\sqrt{n})+1\rfloor e^{-\frac{nr^2}{2}}-p_{t_n}$,
\begin{equation}\label{sghjdhj}
   {|(\P_n-\P) U(\hat{m}_{\ell,n})|}\le \sup_{0<\delta\le 2J(\hat{m}_{\ell,n})}\E\sup_{f\in \N\N^M(W_k,L_k,\delta)}{|(\P_n-\P)U(f)|}+4F_n J(\hat{m}_{\ell,n}) \sqrt{\frac{\ln n}{n}} r.
\end{equation}

Next, we consider the upper bound of $\E\sup_{f\in \N_k^M(\delta)}{|(\P_n-\P)U(f)|}$ in \eqref{sghjdhj}. By symmetrical inequality, we have 
\begin{align*}
 & \E\sup_{f\in \N\N^M(W_k,L_k,\delta)}{|(\P_n-\P)U(f)|} 
   \le \E\sup_{f\in \N\N^M(W_k,L_k,\delta)}{|\frac{2}{n}\sum_{i=1}^{n}U(f(\bm{X}_i,Y_i))b_i|} \\
   & =  \E\sup_{f\in \N\N^M(W_k,L_k,\delta)}{|\frac{2}{n}\sum_{i=1}^{n} (\ell(Y_i-f(\bm{X}_i))-\ell(Y_i))b_i|} \\
   & =  \E \E \left(\sup_{f\in \N\N^M(W_k,L_k,\delta)}{|\frac{2}{n}\sum_{i=1}^{n} (\ell(Y_i-f(\bm{X}_i))-\ell(Y_i))b_i|}| Y_1,\ldots,Y_n \right).
\end{align*}
Let $h_i(u):= \ell(y_i-u)- \ell(y_i)$ be a real function where $y_i\in\R$. Then, $h_i(u)$ is a Lipschitz function satisfying
$$
|h_i(u)-h_i(v)|\le |\ell(y_i-u)-\ell(y_i-v)|\le F_n|u-v|.
$$
Thus, the application of contraction inequality shows that
\begin{equation}\label{jhbdx}
  \E\sup_{f\in \N\N^M(W_k,L_k,\delta)}{|(\P_n-\P)U(f)|}\le \frac{2F_n}{n}\E\sup_{f\in\N\N^M(W_k,L_k,\delta)}|\sum_{i=1}^{n}f(\bm{X}_i)b_i|\le F_n\delta \sqrt{ \frac{L_k}{n}}.
\end{equation}

Finally, the combination of \eqref{jhbdx}, \eqref{khjbkj} and \eqref{sghjdhj} gives that with the probability larger than $1-\lfloor \log_2(B_n\sqrt{n})+1\rfloor e^{-\frac{nr^2}{2}}-p_{t_n}-p_1$,
$$
 I\le cF_nJ(\hat{m}_{\ell,n}) \sqrt{\frac{L_k}{n}}+t_n
$$
and we take $\lambda_n= 2cF_n \sqrt{\frac{L_k}{n}}$.

\medskip
\noindent {\sc Analysis of Part II:} This part is obtained by the definition of $\hat{m}_{H,F_n}$. Since $f_k^{*}\in\N_k^M$, 
$$
II\le \delta_{opt}^2.
$$

\medskip
\noindent {\sc Analysis of Part III:} Since $f_k^{*}\in [-M, M]$ is bounded, similar analysis that is used to obtain \eqref{8811} shows that 
$$
 \P\left( \left| \frac{1}{n}\sum_{i=1}^{n}\left(  \ell(Y_i-f_k^{*}(\bm{X}_i))-\E(\ell(Y-f_k^{*}(\bm{X})))\right)  \right|\ge t_n\right)\le \frac{2\max\{\E|Y|^p, c(p)\} F_n^p}{n^{\frac{p}{2}}t_n^p},
$$
where $c(p)>0$ is a constant that depends on $p$ only and $t_n>0$.
\end{proof}

\noindent\textit{Proof of Theorem \ref{Theorem_Huber}.} Firstly, we bound the approximation error $R(f_k^{*})-R(m)$ in \eqref{*ghh}. Recall the score function $\ell'_{H,\tau}(v)=\min\{ \max{(-\tau_n,v)},\tau_n \}.$  Take the Taylor expansion of $\ell'_{H,\tau}(v)$ at $v\in\R$. Then, for any $w\in\R$,
$$
  \ell_{H,\tau_n}(v+w)-\ell_{H,\tau_n}(v)= \ell'_{H,\tau_n}(v)w+\int_0^w{ \ell'_{H,\tau}(v+t)(w-t)dt}.
$$
Let $\Delta f(\bm{X}):= f_k^{*}(\bm{X})-m(\bm{X})$.  Using above equality, the following relationship hold:
\begin{equation}\label{fdghjgfh}
  \begin{aligned}
       R(f_k^{*})-R(m)&= \E(\ell_{H,\tau_n}(\varepsilon+\Delta f(\bm{X})))-\E( \ell_{H,\tau_n}(\varepsilon))\\
       &= \E( \ell_{H,\tau_n}(\varepsilon) (f_k^{*}(\bm{X})+m(\bm{X})))\\
       &+ \E\left( \int_{0}^{m(\bm{X})-f_k^{*}(\bm{X})} \I(|\varepsilon+t|\le \tau_n)(m(\bm{X})-f_k^{*}(\bm{X})-t)dt \right)\\
       &\le \frac{1}{2}\sup_{\bm{x}} |\E( \ell'_{H,\tau}(\varepsilon)|\bm{X}=\bm{x})|^2+ \frac{1}{2}\|f_k^{*}(\bm{X})-m(\bm{X})\|_2^2\\
       &+ \frac{1}{2}\|f_k^{*}(\bm{X})-m(\bm{X})\|_2^2.
  \end{aligned}
\end{equation}
Since $\E(\varepsilon|\bm{X}=\bm{x})=0$, thus $\E(\varepsilon\I(\varepsilon>0)|\bm{X}=\bm{x})=-\E(\varepsilon \I(\varepsilon<0)|\bm{X}=\bm{x})$. By using this equality, it can be checked that
\begin{align*}
  |\E(\ell_{H,\tau_n}'(\varepsilon)|\bm{X}=\bm{x})| & = |\E(-\I(|\varepsilon|>\tau_n)\varepsilon+ \I(\varepsilon>\tau_n)\tau_n-\I(\varepsilon<-\tau_n)\tau_n|\bm{X}=\bm{x} )| \\
   &\le \E((|\varepsilon-\tau_n)\I(|\varepsilon|>\tau_n))|\bm{X}=\bm{x})\\
   & \le \E(|\varepsilon|(|\varepsilon|/2)^{p-1})\\
   &= v_p  \tau_n^{1-p}.
\end{align*}
According to \eqref{fdghjgfh}, we have 
$$
 R(f_k^{*})-R(m)\le \frac{1}{2}\left( \frac{v_p}{\tau_n^{p-1}}\right)^2+ \|f_k^{*}(\bm{X})-m(\bm{X})\|_2^2.
$$
Based on \eqref{*ghh} and analysis in Lemma \ref{lemma_sbdj}, with the probability larger than $1-\lfloor \log_2(B_n\sqrt{n})+1\rfloor e^{-\frac{nr^2}{2}}-3p_{t_n}-p_1$,

\begin{align}
 R(\hat{m}_{H,n})-R(m)+\lambda_n   J(\hat{m}_{\ell,n})&\le \frac{c\tau_nJ(\hat{m}_{H,n})}{\sqrt{n}}+ 2t_n+ \delta_{opt}^2\nonumber\\
 &+\frac{1}{2}\left( \frac{v_p}{\tau_n^{p-1}}\right)^2+ \|f_k^{*}(\bm{X})-m(\bm{X})\|_2^2+ \lambda_nJ(f_k^{*}),\label{jiljiI}
 \end{align}
where $p_{t_n}:= {2\E|Y|^p \tau_n^p}{n^{-\frac{p}{2}}t_n^{-p}}$. Now, we take $\lambda_n:= 2\tau_n\sqrt{L_k}/\sqrt{n}$ and $t_n:= \tau_n\sqrt{L_k}/\sqrt{n}.$ Since $J(\hat{m}_{H,n})\ge 0$, we can delete this term on the RHS of \eqref{jiljiI}.  Let $f_k^*\in \mathcal{N} \mathcal{N}^M_{d, 1}\left(W_k, L_k, U_n\right)$ be the network given in Theorem \ref{reg_multi_approx} satisfying 
 $$
    \|m-f_k^*\|_\infty \lesssim U_n^{-\frac{\gamma^*}{l}}.
 $$
To minimize \eqref{jiljiI}, set $\tau_n^{2-2p}=U_n^{-2\beta_1}=\tau_n U_n n^{-\frac{1}{2}}L_k^{\frac{1}{2}}$. Namely, we get $\tau_n\asymp (n/L_k)^{\cdot\frac{\beta_1}{(2p-2)(2\beta_1+1)+1}}$ and \eqref{jiljiI} implies
$$
 R(\hat{m}_{H,n})-R(m)\le \max\left\{ (L_kW_k)^{-\alpha_1}, (n/L_k)^{-\frac{1}{2}\cdot\frac{2(2p-2)\beta_1}{(2p-2)(2\beta_1+1)+1}} \right\}.
$$

Next, we need to find the relationship between the excess risk $ R(\hat{m}_{H,n})-R(m)$ and the error $\|\hat{m}_{H,n}-m\|_2$. This part can be done by using  previous results of Huber loss, for example Proposition  3.1 in \cite{fan2024noise}. Namely, if Assumption \ref{assump_heavytail} is satisfied, 
$$
  \|\hat{m}_{H,n}-m\|_2^2\le 8\max\left\{  v_p\tau_n^{1-p},   R(\hat{m}_{H,n})-R(m) \right\}.
$$
If both Assumption \ref{assump_heavytail} and Assumption \ref{assump_heavytail2} are satisfied, then
$$
  \|\hat{m}_{H,n}-m\|_2^2\le 4( R(\hat{m}_{H,n})-R(m) ).
$$
Finally, the combination of \eqref{*ghh} and above two inequalities completes the proof. \hfill\(\Box\)\\

\

\noindent\textit{Proof of Theorem \ref{Theorem_quantile}.} Firstly, we bound the approximation error $R(f_k^{*})-R(q_{\tau})$ in \eqref{*ghh}. Since $\rho_\tau(\cdot)$ is a convex function, the generalization of Newton-Leibniz formula tells us
$$
\rho_\tau(w-v)-\rho_\tau(w)=-v(\tau-\I(w\le 0))+\int_{0}^{v}(\I(w\le z)-\I(w\le 0))dz, \ \ \forall w,v\in\R.
$$ 
Thus, for any functions $f_1(\bm{x}),f_2(\bm{x})$, we have
\begin{align*}
\rho_\tau(Y-f_1(\bm{X}))-\rho_\tau(Y-f_2(\bm{X})) &= -(f_1(\bm{X})-f_2(\bm{X}))(\tau-1\{Y\leq f_2(\bm{X})\}) \\
& +\int_0^{f_1(\bm{X})-f_2(\bm{X})}[1\{Y\leq f_2(\bm{X})+z\}-1\{Y\leq f_2(\bm{X})\}]\,dz \\
&= -(f_1(\bm{X})-f_2(\bm{X}))(\tau-1\{Y\leq q_\tau(\bm{X})\}) \\
& -(f_1(\bm{X})-f_2(\bm{X}))(1\{Y\leq q_\tau(\bm{X})\}-1\{Y\leq f_2(\bm{X})\}) \\
& +\int_0^{f_1(\bm{X})-f_2(\bm{X})}[1\{Y\leq f_2(\bm{X})+z\}-1\{Y\leq f_2(\bm{X})\}]\,dz.
\end{align*}
Taking expectations on above equality. By Fubini's theorem, it is known that
\begin{align}
&\mathbb{E} \left( \rho_\tau(Y - f_1(\bm{X})) - \rho_\tau(Y - f_2(\bm{X})) \right)\nonumber\\
&=  - \mathbb{E} \left( (f_1(\bm{X}) - f_2(\bm{X})) \mathbb{E} \left( (1\{Y \leq q_\tau(\bm{X})\} - 1\{Y \leq f_2(\bm{X})\}) \bigg| \bm{X} \right) \right) \nonumber\\
&\quad + \mathbb{E} \Bigg( \int_0^{f_1(\bm{X}) - f_2(\bm{X})} \Big[ \mathbb{E} \left( 1\{Y \leq f_2(\bm{X}) + z \} \bigg| \bm{X} \right)\nonumber \\
&\quad  - \mathbb{E} \left( 1\{Y \leq f_2(\bm{X})\} \bigg| \bm{X} \right) \Big] dz \Bigg). \label{j,sdf,}
\end{align}

Firstly, take $f_1=f_k^*$ and $f_2=q_\tau$ in \eqref{j,sdf,}. According to the Lipshitz property of conditional distribution $F_{Y|\bm{X}}(\cdot)$ in Assumption \ref{assump_quantile}, thus 
\begin{equation}\label{Ukm2}
    \mathbb{E} \left( \rho_\tau(Y - f_k^*(\bm{X})) - \rho_\tau(Y - q_\tau(\bm{X})) \right) \lesssim E(f_k^*(\bm{X})-q_\tau(\bm{X}))^2.
\end{equation}

Secondly, in \eqref{j,sdf,} take any $f_1\in \mathcal{N}^M \mathcal{N}_{d, 1}\left(W_k, L_k\right)$ and $f_2=f_k^*\in\mathcal{N}_{d, 1}\left(W_k, L_k,U_n\right)$ satisfying $\|f_2-q_\tau\|_\infty\le \Delta$. Here, $f_k^*$ is chosen to be the function in the proof of Theorem \ref{Theorem_Huber}. Define the function
$$
   \kappa(v)=\int_{0}^{v} (F_{Y|\bm{X}=\bm{x}}(q_\tau(\bm{x})+z)-F_{Y|\bm{X}=\bm{x}}(q_\tau(\bm{x}))dz,\ v\in\R.
$$
If $v>2\delta^*$ where $\delta^*$ is given in Assumption \ref{assump_quantile}, $\kappa(v)\ge \int_{\delta^*}^{v}\delta^* dz =(v- \delta^*)\delta^*>\frac{\delta^*}{2}v $. If $0<v\le 2\delta^*$, $\kappa(v)\ge\int_{0}^{v/2}zdz\ge \frac{v^2}{8}$ by Assumption \ref{assump_quantile}. With a similar argument, we can show  $\kappa(v)\gtrsim D^2(v)$ for all $v\in\R$ where $D^2(v):=\min\{|v|,v^2\}$. On the other hand, $D^2(v)\ge \frac{1}{2M}v^2$ when $|v|\le 2M$. Therefore, by using \eqref{j,sdf,} and Cauchy-Schwarz inequality 
\begin{align}
  &\mathbb{E} \left( \rho_\tau(Y - f_1(\bm{X})) - \rho_\tau(Y - f_k^*(\bm{X})) \right) \nonumber\\
  & \ge - \mathbb{E} \left( (f_1(\bm{X}) - f_k^*(\bm{X})) \mathbb{E} \left( (1\{Y \leq q_\tau(\bm{X})\} - 1\{Y \leq f_k^*(\bm{X})\}) \bigg| \bm{X} \right) \right) 
  + \frac{1}{2M}\E(f_k^*(\bm{X})-f_1(\bm{X}))^2 \nonumber\\
  &\ge  -\E(|f_1(\bm{X}) - f_k^*(\bm{X})||f_k^*(\bm{X})-q_\tau(\bm{X})|)+\frac{1}{2M}\E(f_k^*(\bm{X})-f_1(\bm{X}))^2 \nonumber\\
  &\ge -[\E(f_1(\bm{X}) - f_k^*(\bm{X}))^2]^{\frac{1}{2}}[\E(q_\tau(\bm{X}) - f_k^*(\bm{X}))^2]^{\frac{1}{2}}+\frac{1}{2M}\E(f_k^*(\bm{X})-f_1(\bm{X}))^2 \nonumber\\
  &\ge \frac{1}{4M}\E(f_k^*(\bm{X})-f_1(\bm{X}))^2 - M\E(q_\tau(\bm{X}) - f_k^*(\bm{X}))^2, \label{Hk}
\end{align}
where in the last line the inequality $ab\le \frac{1}{4M}a^2+b^2M$ is used. 

Now, from \eqref{*ghh}, \eqref{Ukm2} and \eqref{Hk}, we have
\begin{equation}\label{fbk}
  \E(\hat{p}_{\tau,n}(\bm{X})-q_\tau(\bm{X}))^2\lesssim { \delta_{opt}^2}+ \E(f_k^*(\bm{X})-q_\tau(\bm{X}))^2+ {\frac{J(f_k^{*})}{\sqrt{n/L_k}}},
\end{equation}
with  probability approaching to $1$. Note that $\E(f_k^*(\bm{X})-q_\tau(\bm{X}))^2\le \max\{(L_kW_k)^{-2\alpha_1}, U_n^{-2\beta_1}\}$ with $f_k^*\in \mathcal{N}_{d, 1}\left(W_k, L_k,U_n\right)$. Taking the optimal $U_n=(n/L_k)^{\frac{1}{2(2\beta_1+1)}}$, from \eqref{fbk} we have
$$
  \E(\hat{p}_{\tau,n}(\bm{X})-q_\tau(\bm{X}))^2\lesssim { \delta_{opt}^2}+\max\{(L_kW_k)^{-2\alpha_1}, (n/L_k)^{-\frac{\beta_1}{2\beta_1+1} }\}.
$$
This completes the proof.  \hfill\(\Box\)\\

\subsection{Proof of Theorem \ref{theorem_consistency_class}}

The proof begins with the representation of the true (conditional) density function $\bm{\eta}(\bm{x})$ in the  neural network form. 

\begin{lemma}\label{lemma_neural network representation}
Under Assumption \ref{class:assump}, there is a  series of functions  ${\eta}_j^{last}(\bm{x}),j\in [K]$ such that
\begin{equation}\label{ii1}
  \eta(\bm{x})=\pr{\frac{e^{\eta_1^{last}(\bm{x})}}{\sum_{j=1}^{K}e^{\eta_j^{last}(\bm{x})}}, \dots , \frac{e^{\eta_K^{last}(\bm{x})}}{\sum_{j=1}^{K}e^{\eta_j^{last}(\bm{x})}}}^T, \ \bm{x}\in [0,1]^d.
\end{equation}
Meanwhile, $\eta_j^{last}(\bm{x})=\ln(c\cdot \eta_j(\bm{x}))$ for each $j=1,\ldots,K$ and some $c>0$. Each $\eta_j^{last}(\bm{x})$ is also bounded from up and below. 

\end{lemma}

\begin{proof}
  Let $z_j=e^{\eta_j^{last}(\bm{x})}$ for each $j\in [K]$. Suppose \eqref{ii1} is true. Then, we get the equation
  $$
    z_j=  \eta_k(\bm{x})\cdot \sum_{\ell=1}^{K} z_\ell, \ \forall j\in [K].
  $$
  Write above equations in the following matrix form:
  
  \[
  \underbrace{\begin{pmatrix}
     \eta_1(\bm{x}) & & \\
    & \ddots & \\
    & &  \eta_K(\bm{x})
  \end{pmatrix}
  \begin{pmatrix}
    1 \\
    \vdots \\
    1
  \end{pmatrix}
\begin{pmatrix}
  1 & \cdots & 1
\end{pmatrix}}_{\bm{A}}
 \begin{pmatrix}
    z_1 \\
    \vdots \\
    z_K
  \end{pmatrix}
=
 \begin{pmatrix}
    z_1 \\
    \vdots \\
    z_K
  \end{pmatrix}.
\]
Therefore, we know $(z_1,\ldots,z_K)^T$ must be the eigenvector of $\bm{A}$ and the corresponding eigenvalue is $1$. Let $\bm{z}^*:=(z_1^*,\ldots,z_K^*)^T=( \eta_1(\bm{x}),\ldots,  \eta_K(\bm{x}))^T$. By using the fact that $\sum_{j=1}^{K} \eta_j(\bm{x})=1$, $\bm{z}^*$ is indeed the eigenvector of $\bm{A}$ with the corresponding eigenvalue  $1$. Thus, above linear programming has at least a  solution. Note that other $K-1$ eigenvalues of $\bm{A}$ are all $0$. Thus, any such solution $(z_1,\ldots,z_K)^T$ must be  parallel to $\bm{z}^*$.
\end{proof}

Recall that the neural network density estimator is given by
$$
\hat{\bm{p}}_{n,k}(\bm{x})=\pr{\frac{e^{\hat{p}^{last}_{n,k,1}(\bm{x})}}{\sum_{j=1}^{K}e^{\hat{p}_{n,k,j}^{last}(\bm{x})}}, \ldots , \frac{e^{\hat{p}_{n,k,K}^{last}(\bm{x})}}{\sum_{j=1}^{K}e^{\hat{p}_{n,k,j}^{last}(\bm{x})}}}^T,
$$
where $\hat{p}_{n,k,j}^{last}$ is the $j$-th neuron's output in the last hidden layer. Lemma \ref{lemma_neural network representation} sheds light that the consistency of $\hat{\bm{p}}_{n,k}$ can be guaranteed if each $\hat{p}_{n,k,j}^{last}$ can approximate ${\eta}_j^{last}$ well. Later, we will prove Theorem \ref{theorem_consistency_class} along this route.  For any random function $g(\bm{X}, \bm{Y})$,  define its empirical expectation by
$$
  \E_n f(\bm{X},\bm{Y}):= \frac{1}{n}\sum_{i=1}^{n}f(\bm{X_i},\bm{Y_i}).
$$
First, we establish an Oracle inequality related to $\hat{\bm{p}}_{n,k}$.

\begin{lemma}[Oracle inequality of $\hat{\bm{p}}_{n,k}$] \label{Lemma_Oracle inequality_class}
For any neural network $\tilde{\bm{p}}_k\in \CN_k$,
$$
\begin{aligned}
  &(\E_n-\E)\pr{ \frac{1}{2}\bm{Y}^T\ln\pr{\frac{\hat{\bm{p}}_{n,k}+\tilde{\bm{p}}_k}{2\tilde{\bm{p}}_k}}  }+  \frac{\lambda_n}{4}J(\tilde{\bm{p}}_k)+ \delta_{opt}^2\\
  &\ge R\pr{\frac{\hat{\bm{p}}_{n,k}+\tilde{\bm{p}}_k}{2}, \tilde{\bm{p}}_k} +\lambda_n J(\hat{\bm{p}}_{n,k})-2(1+c_0)\sqrt{R\pr{\frac{\hat{\bm{p}}_{n,k}+\tilde{\bm{p}}_k}{2}, \tilde{\bm{p}}_k}R(\tilde{\bm{p}}_k,\bm{\eta})}\ a.s..
\end{aligned}
$$
\end{lemma}

\begin{proof}
  Since both $\hat{\bm{p}}_{n,k}$ and $\tilde{\bm{p}}_k$ are in the neural network class $\CN_k$, thus they are positive and the inequality we need to prove is well-defined. By Jensen's inequality, we have
  \begin{equation*}
    \frac{1}{2} \ln\pr{ \frac{  \hat{\bm{p}}_{n,k}(\bm{x})+\tilde{\bm{p}}_k(\bm{x})  }{2\tilde{\bm{p}}_k(\bm{x})} }\ge  \frac{1}{4} \ln\pr{\frac{ \hat{\bm{p}}_{n,k}(\bm{x})}{\tilde{\bm{p}}_k(\bm{x})}}.
  \end{equation*}
  By the definition of $\hat{\bm{p}}_{n,k}$,
  $$
  \E_n(-\bm{Y}^T\ln(\hat{\bm{p}}_{n,k}))+\lambda_n J(\hat{\bm{p}}_{n,k})\le \E_n(-\bm{Y}^T\ln{\tilde{\bm{p}}_k})+ \lambda_nJ(\tilde{\bm{p}}_k)+ \delta_{opt}^2.
  $$
  The combination of above two equations give that
  \begin{align}\label{sjdhfdgb}
    \frac{\lambda_n}{4}(J(\hat{\bm{p}}_{n,k})-J(\tilde{\bm{p}}_k)) &\le \E_n\pr{\frac{1}{4}\bm{Y}^T\ln\pr{\frac{\hat{\bm{p}}_{n,k}}{\tilde{\bm{p}}_k}} }+ \delta_{opt}^2  \nonumber\\
     & \le (\E_n-\E)\pr{ \frac{1}{2}\bm{Y}^T\ln\pr{\frac{\hat{\bm{p}}_{n,k}+\tilde{\bm{p}}_k}{2\tilde{\bm{p}}_k}}  }+\E\pr{ \frac{1}{2}\bm{Y}^T\ln\pr{\frac{\hat{\bm{p}}_{n,k}+\tilde{\bm{p}}_k}{2\tilde{\bm{p}}_k}}  }+  \delta_{opt}^2.
  \end{align}
  Since $\ln v\le v-1, \ \forall v>0$, \eqref{sjdhfdgb} implies 
  \begin{equation}\label{bj}
     \frac{\lambda_n}{4}(J(\hat{\bm{p}}_{n,k})-J(\tilde{\bm{p}}_k)) \le (\E_n-\E)\pr{ \frac{1}{2}\bm{Y}^T\ln\pr{\frac{\hat{\bm{p}}_{n,k}+\tilde{\bm{p}}_k}{2\tilde{\bm{p}}_k}}  }+\E\pr{ \frac{1}{2}\bm{Y}^T\ln\pr{\frac{\hat{\bm{p}}_{n,k}+\tilde{\bm{p}}_k}{2\tilde{\bm{p}}_k}}  }+ \delta_{opt}^2.
  \end{equation}
  
  On the other hand, we have
  \begin{align}
    &\E\pr{ \bm{Y}^T\pr{1-\sqrt{\frac{\hat{\bm{p}}_{n,k}+\tilde{\bm{p}}_k}{2\tilde{\bm{p}}_k}}}} \nonumber\\
    &= \int\int \bm{y}^T\pr{1-\sqrt{\frac{\hat{\bm{p}}_{n,k}(\bm{x})+\tilde{\bm{p}}_k(\bm{x})}{2\tilde{\bm{p}}_k(\bm{x})}}}dP(y|\bm{x})dP_X(\bm{x})   \nonumber\\
     &= \int \sum_{k=1}^{K}\pr{1- \sqrt{\frac{\hat{\bm{p}}_{n,k}+\tilde{\bm{p}}_k}{2\tilde{\bm{p}}_k}}} \tilde{\bm{p}}_k(\bm{x}) dP_X(\bm{x}) \nonumber\\
    & + \int \sum_{k=1}^{K}\pr{1- \sqrt{\frac{\hat{\bm{p}}_{n,k}+\tilde{\bm{p}}_k}{2\tilde{\bm{p}}_k}}} (\bm{\eta}_k(\bm{x})-\tilde{\bm{p}}_k(\bm{x})) dP_X(\bm{x})  \nonumber\\
    &= R\pr{\frac{\hat{\bm{p}}_{n,k}+\tilde{\bm{p}}_k(\bm{x})}{2},\tilde{\bm{p}}_k} \nonumber\\
    &+ \int \sum_{k=1}^{K}\pr{1- \sqrt{\frac{\hat{\bm{p}}_{n,k}+\tilde{\bm{p}}_k}{2\tilde{\bm{p}}_k}}}    (\sqrt{\bm{\eta}_k(\bm{x})}-\sqrt{\tilde{\bm{p}}_k(\bm{x})})(\sqrt{\bm{\eta}_k(\bm{x})}+\sqrt{\tilde{\bm{p}}_k(\bm{x})})dP_X(\bm{x})  \nonumber\\
    &= R\pr{\frac{\hat{\bm{p}}_{n,k}+\tilde{\bm{p}}_k}{2}, \tilde{\bm{p}}_k} \nonumber\\
    &+ \int \sum_{k=1}^{K}\pr{1- \sqrt{\frac{\hat{\bm{p}}_{n,k}+\tilde{\bm{p}}_k}{2\tilde{\bm{p}}_k}}}\sqrt{\tilde{\bm{p}}_k(\bm{x})}    (\sqrt{\bm{\eta}_k(\bm{x})}-\sqrt{\tilde{\bm{p}}_k(\bm{x})})\left(1+ \sqrt{\frac{\bm{\eta}_k(\bm{x})}{\tilde{\bm{p}}_k(\bm{x})}}\right)dP_X(\bm{x})  \nonumber\\
    &\ge R\pr{\frac{\hat{\bm{p}}_{n,k}+\tilde{\bm{p}}_k(\bm{x})}{2},\tilde{\bm{p}}_k}-2(1+c_0)\int H\pr{ \frac{\hat{\bm{p}}_{n,k}+\tilde{\bm{p}}_k}{2},\tilde{\bm{p}}_k }H\pr{\tilde{\bm{p}}_k, \bm{\eta}_k}dP_X(\bm{x})   \nonumber\\
    &(\text{by Assumption ???}) \nonumber\\
    &\ge R\pr{\frac{\hat{\bm{p}}_{n,k}+\tilde{\bm{p}}_k}{2}, \tilde{\bm{p}}_k} - 2(1+c_0)\cdot\sqrt{  R\pr{\frac{\hat{\bm{p}}_{n,k}+\tilde{\bm{p}}_k}{2}, \tilde{\bm{p}}_k} R(\tilde{\bm{p}}_k,\bm{\eta}_k)}. \nonumber\\
    &(\text{by Cauchy-Schwarz inequality}) \label{j2hsdj}
  \end{align}
  
Therefore, the combination of \eqref{j2hsdj} and \eqref{bj} completes the proof.
\end{proof}

Lemma \ref{Lemma_Oracle inequality_class} tells us $(\E_n-\E)\pr{ \frac{1}{2}\bm{Y}^T\ln\pr{\frac{\hat{\bm{p}}_{n,k}+\tilde{\bm{p}}_k}{2\tilde{\bm{p}}_k}}  }$ is the most  important term we need to analyze. This term relates to the empirical process
\begin{equation}\label{class_empricalprocess}
  (\E_n-\E)\pr{ \frac{1}{2}\bm{Y}^T\ln\pr{ \frac{\bm{p}+\tilde{\bm{p}}_k}{2\tilde{\bm{p}}_k}} } ,\ \ \bm{p}\in \mathcal{P}_k,
\end{equation}
where $\mathcal{P}_k$ is a probability density function class related to $\hat{\bm{p}}_{n,k}$. We will specify the class $\mathcal{P}_k$ later. First, we bound the expectation of the supremum of this empirical process. 

\begin{lemma}\label{sdhv_lemma}
 Let $\mathcal{P}_k=\left\{  \bm{p}(\bm{x})=\pr{\frac{e^{\bm{p}_1^{last}(\bm{x})}}{\sum_{j=1}^{K}e^{\bm{p}_j^{last}(\bm{x})}}, \dots , \frac{e^{\bm{p}_K^{last}(\bm{x})}}{\sum_{j=1}^{K}e^{\bm{p}_j^{last}(\bm{x})}}} \right\}$ be a subset of classification neural network class $\CN_k$. For any $\tilde{\bm{p}}_k\in\CN_k$, we have
$$
  \E\left( \sup_{\bm{p}\in \mathcal{P}_k}(\E_n-\E)\pr{ \frac{1}{2}\bm{Y}^T\ln\pr{ \frac{\bm{p}(\bm{X})+\tilde{p}_k(\bm{X})}{2\tilde{p}_k}} } \right)\le \E\pr{\frac{2\sqrt{2}K}{n}\sup_{\bm{p}\in \mathcal{P}_k } \sum_{i=1}^n\sum_{j=1}^{K} \bm{p}_j^{last}(X_i)r_{i,j}      },
$$
where $r_{i,j}, i=1,\ldots,n, j=1,\ldots,K$ be i.i.d. Rademecher variables with $\P(r_{i,j}=\pm 1)=\frac{1}{2}$.
\end{lemma} 

\begin{proof}

  Let $\bm{Y}_i=(Y_{1,i},\ldots, Y_{K,i})^T$, $\bm{p}=({p}_1,\ldots,{p}_K)^T$ and $\tilde{p}_k=(\tilde{p}_{k,1},\ldots,\tilde{p}_{k,K})^T$. Note that 
\begin{align*}
  &\sup_{\bm{p}\in \mathcal{P}_k}(\E_n-\E)\pr{ \frac{1}{2}\bm{Y}^T\ln\pr{ \frac{\bm{p}(\bm{X})+\tilde{\bm{p}}_k(\bm{X})}{2\tilde{\bm{p}}_k(\bm{X})}} } \\
  & =\sup_{\bm{p}\in \mathcal{P}_k} \sum_{j=1}^{K}\frac{1}{n}\sum_{i=1}^{n}\left( \frac{1}{2}Y_{j,i}\ln\pr{ \frac{p_j(\bm{X})+\tilde{p}_{k,j}(\bm{X})}{2\tilde{p}_{k,j}(\bm{X})}} -  \E\left[ \frac{1}{2}Y_{j,i}\ln\pr{ \frac{p_j(\bm{X})+\tilde{p}_{k,j}(\bm{X})}{2\tilde{p}_{k,j}(\bm{X})}} \right] \right)  \\
  & \le \sum_{j=1}^{K} \sup_{p_j\in \mathcal{G}_j} \frac{1}{n}\sum_{i=1}^{n}\left( \frac{1}{2}Y_{j,i}\ln\pr{ \frac{p_j(\bm{X})+\tilde{p}_{k,j}(\bm{X})}{2\tilde{p}_{k,j}(\bm{X})}} -  \E\left[ \frac{1}{2}Y_{j,i}\ln\pr{ \frac{p_j(\bm{X})+\tilde{p}_{k,j}(\bm{X})}{2\tilde{p}_{k,j}(\bm{X})}} \right] \right).
\end{align*}

Taking expectation on the above inequality. By the symmetrical inequality, we have
\begin{align}
  &\E \pr{ \sup_{\bm{p}\in \mathcal{P}_k}(\E_n-\E)\pr{ \frac{1}{2}\bm{Y}^T\ln\pr{ \frac{\bm{p}(\bm{X})+\tilde{p}_k(\bm{X})}{2\tilde{p}_k(\bm{X})}} }    }  \nonumber  \\
   & \le  \sum_{j=1}^{K} \E\left[ \sup_{\bm{p}\in \mathcal{P}_k} \frac{1}{n}\sum_{i=1}^{n}\left( \frac{1}{2}Y_{j,i}\ln\pr{ \frac{p_j(\bm{X})+\tilde{p}_{k,j}(\bm{X})}{2\tilde{p}_{k,j}(\bm{X})}} -  \E\left[ \frac{1}{2}Y_{j,i}\ln\pr{ \frac{p_j(\bm{X})+\tilde{p}_{k,j}(\bm{X})}{2\tilde{p}_{k,j}(\bm{X})}} \right] \right) \right] \nonumber\\
   &\le \sum_{j=1}^{K} \E\left[ \sup_{\bm{p}\in \mathcal{P}_k} \frac{1}{n}\sum_{i=1}^{n}\left( Y_{j,i}\ln\pr{ \frac{p_j(\bm{X})+\tilde{p}_{k,j}(\bm{X})}{2\tilde{p}_{k,j}(\bm{X})}} r_{i} \right) \right], \label{vlal12}
\end{align}
where $r_{i}, i=1,\ldots,n$ are i.i.d. Rademecher variables that are independent to $r_{i,j}, i=1,\ldots,n, j=1,\ldots,K$.

  Since $\tilde{\bm{{p}}}_k\in\CN_k$, we have
  $$
    \tilde{\bm{{p}}}_k(x)= \pr{ {\tilde{p}}_{k,1}(x),\dots,{\tilde{p}}_{k,K}(x) }= \pr{\frac{e^{{\tilde{p}}_{k,1}^{last}(\bm{x})}}{\sum_{j=1}^{K}e^{{p}_{k,j}^{last}(\bm{x})}}  \dots , \frac{e^{{\tilde{p}}_{k,K}^{last}(\bm{x})}}{\sum_{j=1}^{K}e^{{\tilde{p}}_{k,j}^{last}(\bm{x})}} }.
  $$

  For each $\ell\in [K]$, we construct a function:

\[
G_\ell(v_1, \ldots, v_K; \tilde{{p}}_{k,1}^{last}, \ldots, \tilde{{p}}_{k,K}^{last};\bm{x})  := \ln \left( \frac{e^{v_\ell}}{e^{v_1}+ \cdots+ e^{v_K}} \left(1 + \sum_{m \neq \ell} e^{\tilde{{p}}_{k,m}^{last}(\bm{x})-\tilde{{p}}_{k,\ell}^{last}(\bm{x})}\right) + \frac{1}{2} \right).
\]

Fix \( \tilde{{p}}_{k,1}^{last}, \ldots, \tilde{{p}}_{k,K}^{last} \), then $G_\ell$ is a function w.r.t. $v_1, \ldots, v_K$ only. Meanwhile, we can bound its partial derivatives as follows.

\begin{align*}
\frac{\partial G_\ell}{\partial v_\ell} &= \frac{1}{e^{v_\ell} \cdot e^{v_k}} \cdot C \cdot \frac{e^{v_1} (e^{v_1} +\cdots + e^{v_k}) - e^{2v_\ell}}{(e^{v_1} +\cdots + e^{v_k})^2} \\
 &= \frac{C e^{v_\ell}}{e^{v_1} +\cdots + e^{v_k}} \cdot \pr{\frac{C e^{v_\ell}}{e^{v_1} +\cdots + e^{v_k}}+\frac{1}{2}}^{-1}\cdot\frac{e^{v_1} +\cdots + e^{v_k} - e^{v_\ell}}{e^{v_1} +\cdots + e^{v_k}}\leq 1
\end{align*}
and when $j\ne \ell$,

\begin{align*}
\frac{\partial G_\ell}{\partial v_j} &= \frac{C}{e^{v_\ell} \cdot e^{v_k}} \cdot \frac{e^{v_\ell} e^{v_j}}{(e^{v_1} +\cdots + e^{v_k})^2}\\
 & = \frac{C e^{v_\ell}}{e^{v_1} +\cdots + e^{v_k}} \cdot \pr{\frac{C e^{v_\ell}}{e^{v_1} +\cdots + e^{v_k}}+\frac{1}{2}}^{-1}\cdot\frac{e^{v_j}}{e^{v_1} +\cdots + e^{v_k}}\\
 & \leq \frac{e^{v_j}}{e^{v_1} +\cdots + e^{v_k}},
\end{align*}
where \( C = \left(1 + \sum_{m \neq \ell} e^{\tilde{\bm{{p}}}_{k,m}^{last}(\bm{x})-\tilde{\bm{{p}}}_{k,\ell}^{last}(\bm{x})}\right)  \). Since all above partial derivatives are positive, some calculations give that

\begin{equation}\label{njl;k2}
\|\nabla G_\ell\|_2^2 \leq 1 + \sum_{j \neq l} \frac{e^{2v_j}}{(e^{v_1} \cdots + e^{v_k})^2} = 2.
\end{equation}
An important observation is that $G_\ell$ is a Lipshitz function whose Lipshitz constant is independent to the value of $C$.

 Since $\bm{p}\in\CN_k$, we write
  $$
    \bm{p}(x)= \pr{ {p}_1(x),\dots, {p}_K(x) }= \pr{\frac{e^{{p}_{1}^{last}(\bm{x})}}{\sum_{j=1}^{K}e^{{p}_j^{last}(\bm{x})}}  \dots , \frac{e^{{p}_K^{last}(\bm{x})}}{\sum_{j=1}^{K}e^{{p}_j^{last}(\bm{x})}} }
  $$
  and the right hand side of \eqref{vlal12} can be written as
  \begin{align*}
  &\E\left[ \sup_{\bm{p}\in \mathcal{P}_k} \frac{1}{n}\sum_{i=1}^{n}\left( Y_{j,i}\ln\pr{ \frac{p_j(\bm{X})+\tilde{p}_{k,j}(\bm{X})}{2\tilde{p}_{k,j}(\bm{X})}} r_{i} \right) \right]\\
  &=\E\left[ \frac{1}{n}\sup_{\bm{p}\in \mathcal{P}_k} \sum_{i=1}^{n}r_{i,j}Y_{j,i}G_j(p_{1}^{last}, \ldots, p_{K}^{last}; \tilde{p}_{k,1}^{last}, \ldots, \tilde{p}_{k,K}^{last};X_i) \right].
  \end{align*}
  
  At this step, we prove in induction that for each $m\in [n]\cup \{0\}$,
  \begin{align}
    &\E\left[ \frac{1}{n}\sup_{\bm{p}\in \mathcal{P}_k} \sum_{i=1}^{n}r_{i}Y_{j,i}G_j(p_{1}^{last}, \ldots, p_{K}^{last}; \tilde{p}_{k,1}^{last}, \ldots, \tilde{p}_{k,K}^{last};\bm{X}_i) \right]  \nonumber\\
    & \le \E\left[ \frac{1}{n}    \sup_{\bm{p}\in \mathcal{P}_k} \Bigg\{2\sqrt{2}\sum_{i=1}^{m}\sum_{j=1}^{K} p_j^{last}(\bm{X}_i)r_{i,j}  +
        \sum_{i=m+1}^{n}r_{i}Y_{j,i}G_j(p_{1}^{last}, \ldots, p_{K}^{last}; \tilde{p}_{k,1}^{last}, \ldots, \tilde{p}_{k,K}^{last};\bm{X}_i) \Bigg\}\right]. \label{oooo12}
  \end{align}
  
  When $m=n$, \eqref{oooo12} is what we need to prove. 
  
  When $m=0$, \eqref{oooo12} holds and is just an equation. Suppose \eqref{oooo12} holds for $m-1$, namely
  \begin{align}
    &\E\left[ \frac{1}{n}\sup_{\bm{p}\in \mathcal{P}_k} \sum_{i=1}^{n}r_{i}Y_{j,i}G_j(p_{1}^{last}, \ldots, p_{K}^{last}; \tilde{p}_{k,1}^{last}, \ldots, \tilde{p}_{k,K}^{last};\bm{X}_i) \right]  \nonumber\\
    & \le \E\left[ \frac{1}{n}    \sup_{\bm{p}\in \mathcal{P}_k} \Bigg\{2\sqrt{2}\sum_{i=1}^{m-1}\sum_{j=1}^{K} p_j^{last}(\bm{X}_i)r_{i,j}  +
        \sum_{i=m}^{n}r_{i}Y_{j,i}G_j(p_{1}^{last}, \ldots, p_{K}^{last}; \tilde{p}_{k,1}^{last}, \ldots, \tilde{p}_{k,K}^{last};\bm{X}_i) \Bigg\}\right]. \label{oooo123}
  \end{align}
  Now, we consider the case for $m$. According to the assumption \eqref{oooo123},
    \begin{align}
    &\E\left[ \frac{1}{n}\sup_{\bm{p}\in \mathcal{P}_k} \sum_{i=1}^{n}r_{i}Y_{j,i}G_j(p_{1}^{last}, \ldots, p_{K}^{last}; \tilde{p}_{k,1}^{last}, \ldots, \tilde{p}_{k,K}^{last};\bm{X}_i) \right]  \nonumber\\
    & \le \E\Bigg[ \frac{1}{n}    \sup_{\bm{p}\in \mathcal{P}_k}\Bigg\{2\sqrt{2} \sum_{i=1}^{m-1}\sum_{j=1}^{K} p_j^{last}(\bm{X}_i)r_{i,j}  +
        \sum_{i=m+1}^{n}r_{i}Y_{j,i}G_j(p_{1}^{last}, \ldots, p_{K}^{last}; \tilde{p}_{k,1}^{last}, \ldots, \tilde{p}_{k,K}^{last};\bm{X}_i) \nonumber \\
    &+  r_{m}Y_{j,m}G_j(p_{1}^{last}, \ldots, p_{K}^{last}; \tilde{p}_{k,1}^{last}, \ldots, \tilde{p}_{k,K}^{last};\bm{X}_m)   \Bigg\}\Bigg] \nonumber\\
    &:= \E \left( \frac{1}{n}\sup_{\bm{p}\in \mathcal{P}_k} \left\{ h(p)+  r_{m}Y_{j,m}G_j(p_{1}^{last}, \ldots, p_{K}^{last}; \tilde{p}_{k,1}^{last}, \ldots, \tilde{p}_{k,K}^{last};\bm{X}_m) \right\} \right) \nonumber\\
    &= \E\pr{ \frac{1}{n} \cdot\E_m\pr{ \sup_{\bm{p}\in \mathcal{P}_k} \left\{ h(p)+  r_{m}Y_{j,m}G_j(p_{1}^{last}, \ldots, p_{K}^{last}; \tilde{p}_{k,1}^{last}, \ldots, \tilde{p}_{k,K}^{last};\bm{X}_m) \right\}   } }, \label{bdhnc23}
  \end{align}
where the notation $\E_m$ means we take expectation w.r.t. $r_m$ only while fixing other random variables. Now,  define two p.d.f.s
\begin{align*}
 \bm{p}^+ &\in \arg\sup_{\bm{p}\in \mathcal{P}_k} \left\{ h(\bm{p})+  Y_{j,m}G_j({p}_{1}^{last}, \ldots, {p}_{K}^{last}; \tilde{{p}}_{k,1}^{last}, \ldots, \tilde{{p}}_{k,K}^{last};\bm{X}_m) \right\} \\
 \bm{p}^{-} &\in \arg\sup_{\bm{p}\in \mathcal{P}_k} \left\{ h(\bm{p})-  Y_{j,m}G_j({p}_{1}^{last}, \ldots, {p}_{K}^{last}; \tilde{{p}}_{k,1}^{last}, \ldots, \tilde{{p}}_{k,K}^{last};\bm{X}_m) \right\}.
\end{align*}
Since $\bm{p}^+, \bm{p}^- \in\CN_k$, we have
  \begin{align*}
    \bm{p}^+(\bm{x}) &= \pr{ {p}^+_{1}(\bm{x}),\dots,{p}^+_{k}(\bm{x}) }= \pr{\frac{e^{{p}_{1}^{+,last}(\bm{x})}}{\sum_{j=1}^{K}e^{{p}_{j}^{+,last}(\bm{x})}},  \dots , \frac{e^{{p}_{k}^{+,last}(\bm{x})}}{\sum_{j=1}^{K}e^{{p}_{j}^{+,last}(\bm{x})}} }, \bm{x}\in [0,1]^d\\
       \bm{p}^-(\bm{x}) &= \pr{ {p}^-_{1}(\bm{x}),\dots,{p}^-_{k}(\bm{x}) }= \pr{\frac{e^{{p}_{1}^{-,last}(\bm{x})}}{\sum_{j=1}^{K}e^{{p}_{j}^{-,last}(\bm{x})}},  \dots , \frac{e^{{p}_{k}^{-,last}(\bm{x})}}{\sum_{j=1}^{K}e^{{p}_{j}^{-,last}(\bm{x})}} }, \bm{x}\in [0,1]^d
  \end{align*}

From \eqref{bdhnc23}, the following relationships hold
\begin{align}
 &\E_m\pr{ \sup_{\bm{p}\in \mathcal{P}_k} \left\{ h(\bm{p})+  r_{m}Y_{j,m}G_j({p}_{1}^{last}, \ldots, {p}_{K}^{last}; \tilde{{p}}_{k,1}^{last}, \ldots, \tilde{{p}}_{k,K}^{last};X_m) \right\}   }   \nonumber\\
   = \frac{1}{2}\Bigg( & h(\bm{p}^+)+  Y_{j,m}G_j({{p}}_{1}^{+,last}, \ldots, {{p}}_{K}^{+,last}; \tilde{p}_{k,1}^{last}, \ldots, \tilde{p}_{k,K}^{last};\bm{X}_m) \nonumber\\
    + &h(\bm{p}^-)-  Y_{j,m}G_j({{p}}_{1}^{-,last}, \ldots, {{p}}_{K}^{-,last}; \tilde{{p}}_{k,1}^{last}, \ldots, \tilde{{p}}_{k,K}^{last};\bm{X}_m)          \Bigg)\nonumber\\
   \le   \frac{1}{2}\Bigg(  & h(\bm{p}^+)+   h(\bm{p}^-)+\sqrt{2}\|\bm{p}^{+,last}(\bm{X}_m) -\bm{p}^{-,last}(\bm{X}_m)\|_2\Bigg), \label{bbaksndal2}
\end{align}
where in the last line we  set $\bm{p}^{+,last}(\bm{X}_m):=(p^{+,last}_1(\bm{X}_m),\ldots,p^{+,last}_K(\bm{X}_m))^T$, $\bm{p}^{-,last}(\bm{X}_m):=(p^{-,last}_1(\bm{X}_m),\ldots,p^{-,last}_K(\bm{X}_m))^T$  and use the Lipshitz property of $G_j$ (see its Lipshitz constant in \eqref{njl;k2}). According to Khintchine's inequality,
\begin{equation}\label{Khintchine}
   \|\bm{p}^{+,last}(\bm{X}_m) -\bm{p}^{-,last}(\bm{X}_m)\|_2\le 2\E_{r_{m,j}}\left| \sum_{j=1}^{K} (\bm{p}_j^{+,last}(\bm{X}_m) -\bm{p}_j^{-,last}(\bm{X}_m))r_{m,j}\right|,
\end{equation}
where the expectation is only taken w.r.t. $r_{m,j}, j=1,\ldots,K$. Therefore, \eqref{bbaksndal2} can be upper bounded as follows
\begin{align}
  & \frac{1}{2}\Bigg(  h(\bm{p}^+)+   h(\bm{p}^-)+\sqrt{2}\|\bm{p}^{+,last}(\bm{X}_m) -\bm{p}^{-,last}(\bm{X}_m)\|_2 \Bigg) \nonumber\\
  & \le  \sup_{\bm{p}_1,\bm{p}_2\in \mathcal{P}_k} \left\{ \frac{h(\bm{p}_1)}{2}+   \frac{h(\bm{p}_2)}{2}+\frac{\sqrt{2}}{2}\|\bm{p}_1^{last}(\bm{X}_m) -\bm{p}_2^{last}(\bm{X}_m)\|_2 \right\}, \label{yhgdskh}
\end{align}
where recall that
  \begin{align*}
    \bm{p}_1(x) &= \pr{\frac{e^{p_{1,1}^{last}(\bm{x})}}{\sum_{j=1}^{K}e^{p_{1,j}^{last}(\bm{x})}},  \dots , \frac{e^{p_{1,k}^{last}(\bm{x})}}{\sum_{j=1}^{K}e^{p_{1,j}^{last}(\bm{x})}} }, \bm{x}\in [0,1]^d\\
      \bm{p}_2(x) &= \pr{\frac{e^{p_{2,2}^{last}(\bm{x})}}{\sum_{j=2}^{K}e^{p_{2,j}^{last}(\bm{x})}},  \dots , \frac{e^{p_{2,k}^{last}(\bm{x})}}{\sum_{j=2}^{K}e^{p_{2,j}^{last}(\bm{x})}} }, \bm{x}\in [0,1]^d
  \end{align*}
and $\bm{p}_1^{last}:= (p_{1,1}^{last},\ldots,p_{1,K}^{last})^T$ and  $\bm{p}_2^{last}:= (p_{2,1}^{last},\ldots,p_{2,K}^{last})^T$. Therefore, the  combination of \eqref{Khintchine} and \eqref{yhgdskh} leads that
\begin{align*}
 & \frac{1}{2}\Bigg(  h(\bm{p}^+)+   h(\bm{p}^-)+\sqrt{2}\|p^{+,last}(\bm{X}_m) -p^{-,last}(\bm{X}_m)\|_2 \Bigg) \nonumber\\
   &\leq \E_{r_{m,j}} \pr{\frac{1}{2}h(\bm{p}^+)+  \frac{1}{2} h(\bm{p}^-) +\sqrt{2}\left| \sum_{j=1}^{K} (p_j^{+,last}(\bm{X}_m) -p_j^{-,last}(\bm{X}_m))r_{m,j}\right| }\\
  & \le \E_{r_{m,j}}\pr{ \sup_{\bm{p}_1,\bm{p}_2\in \mathcal{P}_k} \left\{ \frac{h(\bm{p}_1)}{2}+   \frac{h(\bm{p}_2)}{2}+ \sqrt{2}\left| \sum_{j=1}^{K} (p_{1,j}^{last}(\bm{X}_m) -p_{2,j}^{last}(\bm{X}_m))r_{m,j}\right| \right\}  }\\
  &= \E_{r_{m,j}}\pr{ \sup_{\bm{p}_1,\bm{p}_2\in \mathcal{P}_k} \left\{ \frac{h(\bm{p}_1)}{2}+   \frac{h(\bm{p}_2)}{2}+ \sqrt{2} \sum_{j=1}^{K} (p_{1,j}^{last}(\bm{X}_m) -p_{2,j}^{last}(\bm{X}_m))r_{m,j} \right\}  } \\
  &( \text{We can exchange\ } \bm{p}_1\ \text{and } \bm{p}_2\ \text{to achieve this point.})\\
  &=  \E_{r_{m,j}}\pr{ \sup_{\bm{p}_1\in \mathcal{P}_k} \left\{ \frac{h(\bm{p}_1)}{2}+ \sqrt{2} \sum_{j=1}^{K} p_{1,j}^{last}(\bm{X}_m)r_{m,j} \right\}  }+\E_{r_{m,j}}\pr{ \sup_{\bm{p}_2\in \mathcal{P}_k} \left\{ \frac{h(\bm{p}_2)}{2}- \sqrt{2} \sum_{j=1}^{K} p_{2,j}^{last}(\bm{X}_m)r_{m,j} \right\}  }\\
  &= 2\E_{r_{m,j}}\pr{ \sup_{\bm{p}_1\in \mathcal{P}_k} \left\{ \frac{h(\bm{p}_1)}{2}+ \sqrt{2} \sum_{j=1}^{K} p_{1,j}^{last}(\bm{X}_m)r_{m,j} \right\}  }\\
  &= \E_{r_{m,j}}\pr{ \sup_{\bm{p}_1\in \mathcal{P}_k} \left\{ {h(\bm{p}_1)}+ 2\sqrt{2} \sum_{j=1}^{K} p_{1,j}^{last}(\bm{X}_m)r_{m,j} \right\}  }.
\end{align*}
According to above inequality and \eqref{bdhnc23}, \eqref{oooo12}  holds indeed for the case  $m$. Finally, our result holds due to \eqref{vlal12}.
\end{proof}

According to Lemma \ref{Lemma_Oracle inequality_class}, the second step is to establish the concentration  inequality of the empirical process \eqref{class_empricalprocess},
$$
(\E_n-\E)\pr{ \frac{1}{2}\bm{Y}^T\ln\pr{ \frac{\bm{p}+\tilde{\bm{p}}_k}{2\tilde{\bm{p}}_k}} },\ \ \bm{p}\in \mathcal{P}_k,
$$
where $\mathcal{P}_k$ is a density function class we will specify later. Our analysis begins with a slight generalization of McDiarmid's inequality.

\begin{lemma}[McDiarmid's inequality for random vectors]\label{McDiarmid's inequality vec}
  Let $\bm{Z}_i\in\mathcal{\bm{Z}}\subseteq\R^{d+K}, i=1,\ldots,n$ be i.i.d. random vectors. Let $g: \mathcal{\bm{Z}}^n\to\R$ satisfy 
  \begin{equation}\label{class_bound_assump}
    \sup_{\bm{z}_1,\ldots,\bm{z}_n,\bm{z}_n'\in\mathcal{\bm{Z}}}|g(\bm{z}_1,\ldots,\bm{z}_n)-g(\bm{z}_1,\ldots,\bm{z}_{i-1},\bm{z}_i',\bm{z}_{i+1},\ldots,\bm{z}_n)|\le c_i, \quad 1\le i\le n,
  \end{equation}
  where $c_1,\ldots,c_n$ are positive constants. For any $t>0$, we have 
  $$
   \P\pr{  g(\bm{Z}_1,\ldots,\bm{Z}_n) -\E(g(\bm{Z}_1,\ldots,\bm{Z}_n))\ge t}\le e^{-\frac{2t^2}{\sum_{i=1}^{n}c_i^2}}.
  $$
\end{lemma}

\begin{remark}
  This result reveals an important observation that the tail probability does not depend on the dimension $d+K$ as long as $g$ satisfies bounded difference property \eqref{class_bound_assump}. This point is not pointed out and observed in literature.
\end{remark}

\begin{proof}
  Write $W:= g(\bm{Z}_1,\ldots,\bm{Z}_n)$ and $\E_i(\cdot):=\E(\cdot|\bm{Z}_1,\ldots,\bm{Z}_i)$. Define the martingale difference  $\Delta_i(\bm{Z}_1,\ldots,\bm{Z}_i)= \E_i( W)-\E_{i-1}(W)$. For each $\Delta_i$, we fix $\bm{Z}_1=\bm{z}_1,\ldots,\bm{Z}_{i-1}=\bm{z}_{i-1}$ with $\bm{z}_1,\ldots,\bm{z}_{i-1}\in\mathcal{\bm{Z}}$. Since $\bm{Z}_1,\ldots,\bm{Z}_n$ are independent, 
  $$
\begin{aligned}
& \left|\Delta_{i}\left(\bm{z}_{1}, \ldots, \bm{z}_{i}, \bm{Z}\right)\right| \\
= & \left|\mathbb{E}\left[g\left(\bm{z}_{1}, \ldots, \bm{z}_{i-1}, \bm{Z}, \bm{Z}_{i+1}, \ldots, \bm{Z}_{n}\right)\right]-\mathbb{E}\left[g\left(\bm{z}_{1}, \ldots, \bm{z}_{i-1}, \bm{Z}_{i}, \bm{Z}_{i+1}, \ldots, \bm{Z}_{n}\right)\right]\right| \\
= & \left|\mathbb{E}\left[g\left(\bm{z}_{1}, \ldots, \bm{z}_{i-1}, \bm{Z}, \bm{Z}_{i+1}, \ldots, \bm{Z}_{n}\right)-g\left(\bm{z}_{1}, \ldots, \bm{z}_{i-1}, \bm{Z}_{i}, \bm{Z}_{i+1}, \ldots, \bm{Z}_{n}\right)\right]\right| \\
\leq & \mathbb{E}\left[\left|g\left(\bm{z}_{1}, \ldots, \bm{z}_{i-1}, \bm{Z}, \bm{Z}_{i+1}, \ldots, \bm{Z}_{n}\right)-g\left(\bm{z}_{1}, \ldots, \bm{z}_{i}, \bm{Z}_{i-1}, \bm{Z}_{i+1}, \ldots, \bm{Z}_{n}\right)\right|\right] \\
\leq & c_{i}.
\end{aligned}
$$
Therefore, for any $\lambda>0$, the moment generation function of $W-\E(W)$ can be bounded below:
$$
\begin{aligned}
\mathbb{E} e^{\lambda(W-\mathbb{E}(W))} & =\mathbb{E} e^{\lambda \sum_{i=1}^{n} \Delta_{i}}=\mathbb{E}\left[\mathbb{E}_{n-1}\left(e^{\lambda\left(\sum_{i=1}^{n-1} \Delta_{i}\right)+\lambda \Delta_{n}}\right)\right] \\
& =\mathbb{E}\left[e^{\lambda\left(\sum_{i=1}^{n-1} \Delta_{i}\right)}\right] \mathbb{E}_{n-1}\left[e^{\lambda \Delta_{n}}\right] \\
& \leq \mathbb{E}\left[e^{\lambda\left(\sum_{i=1}^{n-1} \Delta_{i}\right)}\right] e^{\lambda^{2} c_{n}^{2} / 2} \\
&(\text{by Hoeffding's Lemma; see Lemma 2.2 in \cite{boucheron2013concentration}})\\
& \cdots \\
& \leq e^{\lambda^{2}\left(\sum_{i=1}^{n} c_{i}^{2}\right) / 2}.
\end{aligned}
$$
Then, we can get the probability tail bound according to the standard Chernoff's argument.
\end{proof}

\begin{theorem}[Oracle inequality for classification neural networks]\label{theorem_consistency_class22}
 Choose $r>0$, $\lambda_n\asymp K^2/\sqrt{n}$ and $\tilde{\bm{p}}_k\in\CN_k$ with $\tilde{\bm{p}}_k(X)>cK^{-\gamma}/2\ a.s.$. Under Assumption \ref{class:assump}, we have
  \begin{align*}
    &R(\boldsymbol{\eta}(\bm{X}), \hat{\bm{p}}_{n,k}(\bm{X}))\\
     &  \lesssim  \underbrace{\inf_{c\in\R} \sum_{j=1}^{K} \left(\sum_{j=1}^{K}\|\tilde{p}_{k,j}^{last}-\ln \eta_j-c\|_\infty^2\right)^{\frac{1}{2}} }_{approximation\ error}+ \underbrace{{\frac{K^{\frac{3}{2}\vee \gamma}}{\sqrt{n}}} +  \lambda_n J(\tilde{p}_k)}_{sample\ error}+ \underbrace{ \delta_{opt}^2}_{optimization\ error}
  \end{align*}
  with the probability larger than $1-\ln n \cdot n^{-r}$.
\end{theorem}

\begin{proof}

Now, we define the density class $\mathcal{P}_k$ as follows.  For any $\delta>0$, define the density class
$$
\mathcal{Q}_{k,\delta}:=\left\{ \bm{p}\in\CN_k:   \frac{1}{K}+\sqrt{K}J^C(\bm{p}) \le \delta \right\}.
$$
According to Assumption \ref{class:assump} and the definition of $\tilde{\bm{{p}}}_{k}$, we at least have
$
\sup_{j\in [K],x\in [0,1]^d}\|{\tilde{p}}^{last}_{k,j}(x)- {\eta}_j(x)\|_\infty \le 1.
$
Since $\sup_{x\in [0,1]^d, j\in [K]}\| {\eta}_j(x)\|_\infty \le 1$, 
$$
\sup_{j\in [K],x\in [0,1]^d}\|{\tilde{p}}^{last}_{k,j}(x)\|_\infty \le 2.
$$

Let $\bm{Z}_i=(\bm{Y}_i,\bm{X}_i)^T\in \R^{d+K}$. Next, we show that the supremum of empirical process  \eqref{class_empricalprocess}: 
$$
 F_{\mathcal{Q}_{k,\delta}}(\bm{Z}_1,\ldots,\bm{Z}_n):=\sup_{\bm{p}\in \mathcal{Q}_{k,\delta}}\pr{  \frac{1}{n}\sum_{i=1}^{n}\bm{Y}_i^T\ln\pr{\frac{\bm{p}(\bm{X}_i)}{2\tilde{\bm{{p}}}_k(\bm{X}_i)} + \frac{1}{2}}  - \E\left[ \bm{Y}_i^T\ln\pr{\frac{\bm{p}(\bm{X}_i)}{2\tilde{\bm{{p}}}_k(\bm{X}_i)} + \frac{1}{2}}\right] }
$$
satisfies the bounded difference property \eqref{class_bound_assump}. 

Note that $|\sup_{n\ge 1} \{a_n\}- \sup_{n\ge 1} \{b_n\}|\le \sup_{n\ge 1} |a_n-b_n|$ for any two selected sequences. Choose another vector $\bm{z}_1'=(\bm{x}_1',\bm{y}_1')^T\in \mathcal{Z}$. Then,
\begin{align*}
  &|F_{\mathcal{Q}_{k,\delta}}(\bm{z}_1,\ldots,\bm{z}_n)- F_{\mathcal{Q}_{k,\delta}}(\bm{z}_1',\ldots,\bm{z}_n)| \\
  &\le \frac{1}{n}\sup_{\bm{p}\in \mathcal{Q}_{k,\delta}} \left| \sum_{i=1}^{n}\bm{y}_i^T\ln\pr{\frac{\bm{p}(\bm{x}_i)}{2\bm{\tilde{p}}_k(\bm{x}_i)}+\frac{1}{2}}-  \bm{y}_1'^T\ln\pr{\frac{\bm{p}(\bm{x}_1')}{2\tilde{\bm{{p}}}_k(\bm{x}_1')}+\frac{1}{2}}- \sum_{i=2}^{n}\bm{y}_i^T\ln\pr{\frac{\bm{p}(\bm{x}_i)}{2\tilde{\bm{{p}}}_k(\bm{x}_i)}+\frac{1}{2}}  \right|\\
  &=  \frac{1}{n}\sup_{\bm{p}\in \mathcal{Q}_{k,\delta}} \left| \bm{y}_1^T\ln\pr{\frac{\bm{p}(\bm{x}_1)}{2\tilde{\bm{{p}}}_k(\bm{x}_1)}+\frac{1}{2}}-  \bm{y}_1'^T\ln\pr{\frac{\bm{p}(\bm{x}_1')}{2\tilde{\bm{{p}}}_k(\bm{x}_1')}+\frac{1}{2}} \right|\\
  &= \frac{1}{n}  \left|  \ln\pr{\frac{p_{j_1}(\bm{x}_1)}{2{\tilde{p}}_{k,j_1}(\bm{x}_1)}+\frac{1}{2}}-  \ln\pr{\frac{p_{j_2}(\bm{x}_1')}{2{\tilde{p}}_{k,j_2}(\bm{x}_1')}+\frac{1}{2}} \right|\\
  &(\text{Suppose } j_1, j_2\ \text{are positions where } y_1, y_1'\ \text{take } 1 )\\
  &=\frac{1}{n}  \left|  \ln\pr{\frac{p_{j_1}(\bm{x}_1)}{2{\tilde{p}}_{k,j_1}(\bm{x}_1)}+\frac{1}{2}}+\ln 2-  \ln\pr{\frac{p_{j_2}(\bm{x}_1')}{2{\tilde{p}}_{k,j_2}(\bm{x}_1')}+\frac{1}{2}}-\ln 2 \right|\\
  &= \frac{1}{n}  \left|  \ln\pr{\frac{p_{j_1}(\bm{x}_1)}{{\tilde{p}}_{k,j_1}(\bm{x}_1)}+1}-  \ln\pr{\frac{p_{j_2}(\bm{x}_1')}{{\tilde{p}}_{k,j_2}(\bm{x}_1')}+1} \right|\\
  &\le \frac{1}{n}  \left[ \ln\pr{\frac{p_{j_1}(\bm{x}_1)}{{\tilde{p}}_{k,j_1}(\bm{x}_1)}+1}+  \ln\pr{\frac{p_{j_2}(\bm{x}_1')}{{\tilde{p}}_{k,j_2}(\bm{x}_1')}+1} \right]\\
  &\le \frac{1}{n}  \left[ \ln\pr{\frac{2p_{j_1}(\bm{x}_1)}{K^{-\gamma}}+1}+ \ln\pr{\frac{2p_{j_2}(\bm{x}_1')}{K^{-\gamma}}+1} \right]\\
  &(\text{by Assumption \ref{class:assump} and the definition of } {\tilde{p}}_{k})\\
  &\le \frac{1}{n}\cdot {2K^{\gamma}} \left[ p_{j_1}(\bm{x}_1)+p_{j_2}(\bm{x}_1') \right]\\
  &(\text{by}\ \ln(1+v)\le v, v>0)\\
  &\le \frac{1}{n}\cdot {4K^\gamma} \|\bm{p}\|_\infty
\end{align*}

Now, we construct a multivariate function
\begin{equation}\label{MJks'}
  G(v_1,\ldots,v_K):= \frac{e^{v_1}}{\sum_{i=1}^{K} e^{v_i}}, v_i\in\R.
\end{equation}
Some basic calculations show that
$$
\left|G(v_1,\ldots,v_K)- \frac{1}{K}\right|\le \|\nabla G\|_2\pr{\sum_{i=1}^{K}v_i^2}^{\frac{1}{2}}\le \sqrt{K}\max_{i}|v_i|.
$$
Recall $\bm{p}\in\CN_k$ has the structure:
  $$
    \bm{p}(x)= \pr{\frac{e^{{p}_{1}^{last}(\bm{x})}}{\sum_{j=1}^{K}e^{p_j^{last}(\bm{x})}}  \dots , \frac{e^{{p}_{K}^{last}(\bm{x})}}{\sum_{j=1}^{K}e^{{p}_{j}^{last}(\bm{x})}} }.
  $$
Therefore,
$$
\|\bm{p}\|_\infty\le \frac{1}{K}+\sqrt{K}\max_{j}|p_j^{last}|\le \frac{1}{K}+\sqrt{K}J^C(\bm{p})\le \delta
$$
and
$$
  |F_{\mathcal{Q}_{k,\delta}}(\bm{z}_1,\ldots,\bm{z}_n)- F_{\mathcal{Q}_{k,\delta}}(\bm{z}_1',\ldots,\bm{z}_n)| \le \frac{\delta}{n}\cdot {4K^\gamma}.
$$
With the similar argument, it can be shown the above difference inequality  also holds for other coordinates. 

Thus, according to Lemma \ref{McDiarmid's inequality vec}, for any $r,\delta>0$,
\begin{equation}\label{class_micsgle}
  \P\pr{F_{\mathcal{Q}_{k,\delta}}(\bm{Z}_1,\ldots,\bm{Z}_n)\ge \E(F_{\mathcal{Q}_{k,\delta}}(\bm{Z}_1,\ldots,\bm{Z}_n))+2\delta r }\le e^{-\frac{nr^2}{32K^{2\gamma}}}.
\end{equation}
Set $\delta_j:= 2^j/\sqrt{n},j=1,2,\ldots,B_n$ with $B_n=\lfloor \log_2(c n^{\tau+1/2})\rfloor+1$.  According to \eqref{class_micsgle}, 
\begin{equation}\label{ldjlsjd3}
     \P\pr{ \bigcup_{j=1}^{B_n} \left\{ F_{\mathcal{Q}_{k,\delta_j}}(\bm{Z}_1,\ldots,\bm{Z}_n)\ge \E(F_{\mathcal{Q}_{k,\delta_j}}(\bm{Z}_1,\ldots,\bm{Z}_n))+2\delta_j r \right\} }\le B_n e^{-\frac{nr^2}{32K^{2\gamma}}}.
\end{equation}

On the other hand, the constant density predictor $\bm{p}^{cons}=\pr{\frac{1}{K},\ldots, \frac{1}{K}}^T\in\CN_k$ and its last hidden layer always outputs $0$. Thus, $J\pr{\bm{p}^{cons}}=0$. According to the definition of $\hat{\bm{p}}_n$, it is known that
$$
  \lambda_n J(\hat{\bm{p}}_n) \le -\frac{1}{n}\sum_{i=1}^n \bm{Y}_i^\top \log \hat{\bm{p}}_n(\bm{\bm{X}}_i)+\lambda_n J(\hat{\bm{p}}_n)\le - \frac{1}{n}\sum_{i=1}^n \bm{Y}_i^\top \log\bm{p}^{cons}_n+\delta_{opt}^2.
$$
Namely, for some $\tau>0$,
$$
  J(\hat{\bm{p}}_n) \le \frac{1}{\lambda_n} \pr{\ln K+ \delta_{opt}^2}\lesssim n^{\tau},
$$
as long as $\lambda_n\asymp n^{-\tau_1}$, $K\asymp n^{\tau_2}$ and $\delta_{opt}^2 \asymp n^{\tau_3}$ with $\tau_1,\tau_2,\tau_3>0$. Therefore, an important observation is that for some constant $c>0$,
\begin{equation}\label{giakh34}
  \hat{\bm{p}}_n\in \mathcal{Q}_{k,c n^{\tau}}\ \ a.s..
\end{equation}

By \eqref{giakh34}, it can be seen that $\hat{\bm{p}}_n\in \mathcal{Q}_{k, \delta_{B_n}}\ \ a.s..$  Thus, there is $j^*\in\\bm{Z}^+$ such that 
$$
\delta_{j^*}<\pr{\frac{1}{K}+\sqrt{K}J^C(\hat{\bm{p}}_{n,k})}\le\delta_{j^*+1} \ \ \text{or}\ \ \pr{\frac{1}{K}+\sqrt{K}J(\hat{\bm{p}}_{n,k})}\le \delta_{1}.
$$

\textbf{Case 1}: The event $\{\delta_{j^*}<\pr{\frac{1}{K}+\sqrt{K}J(\hat{\bm{p}}_{n,k})}\le\delta_{j^*+1} \}$ happens for some $j^*$.  Replace  $r$ in \eqref{ldjlsjd3} by $rK^{\gamma}\sqrt{\frac{\ln n}{n}}$. Therefore, \eqref{ldjlsjd3} shows that with the probability larger than $1-B_n\cdot n^{-r}$,
\begin{align}
  & \frac{1}{n}\sum_{i=1}^{n}\bm{Y}_i^T\ln\pr{\frac{\hat{\bm{p}}_{n,k}(\bm{X}_i)}{2\bm{\tilde{p}}_k(\bm{X}_i)} + \frac{1}{2}}  - \E\left[ \bm{Y}_i^T\ln\pr{\frac{\hat{\bm{p}}_{n,k}(\bm{X}_i)}{2\bm{\tilde{p}}_k(\bm{X}_i)} + \frac{1}{2}}\right]\nonumber\\
 & \le  T(K^{-\frac{3}{2}}+J(\hat{\bm{p}}_{n,k})) +K^\gamma\sqrt{\frac{\ln n}{n}}, \nonumber
\end{align}
where for any $\delta>0$, define
$
\mathcal{P}_{k,\delta}:=\left\{ \bm{p}\in\CN_k:  J^C(\bm{p}) \le \delta \right\}
$
and
$$
T(\delta):= \E\pr{ F_{  \mathcal{P}_{k, \delta }  }(\bm{Z}_1,\ldots,\bm{Z}_n)}.
$$

On the other hand, by Lemma \ref{sdhv_lemma} it is known that for any $\delta>0$,
\begin{align}
 T(\delta) & \le  \E\pr{\frac{2\sqrt{2}K}{n}\sup_{\bm{p}\in \mathcal{P}_{k,\delta} } \sum_{i=1}^n\sum_{j=1}^{K} \bm{p}_j^{last}(\bm{X}_i)r_{i,j}      } \nonumber\\
  &\le \frac{2\sqrt{2}K}{n}\sum_{j=1}^{K} \E\pr{\sup_{\bm{p}_j^{last}\in \mathcal{G}_j } \sum_{i=1}^n \bm{p}_j^{last}(\bm{X}_i)r_{i,j} }\nonumber\\
  &\le \frac{2\sqrt{2}K^2\sqrt{L_k}}{\sqrt{n}}\delta, \label{yykdhkh2}
\end{align}
where in the last line we use Proposition \ref{pro_relu_multilayer}. In conclusion, 
\begin{align}
  & \frac{1}{n}\sum_{i=1}^{n}\bm{Y}_i^T\ln\pr{\frac{\hat{\bm{p}}_{n,k}(\bm{X}_i)}{2\bm{\tilde{p}}_k(\bm{X}_i)} + \frac{1}{2}}  - \E\left[ \bm{Y}_i^T\ln\pr{\frac{\hat{\bm{p}}_{n,k}(\bm{X}_i)}{2\bm{\tilde{p}}_k(\bm{X}_i)} + \frac{1}{2}}\right]\nonumber\\
 & \le  2\sqrt{2}\frac{K^{\frac{3}{2}}}{\sqrt{n/L_k}}+2\sqrt{2}\frac{K^2}{\sqrt{n/L_k}}J(\hat{\bm{p}}_{n,k}) + K^\gamma\sqrt{\frac{\ln n}{n}} \label{NHhno}
\end{align}
holds with the probability larger than $1-B_n\cdot n^{-r}$.

\textbf{Case 2}: The event $\{{\frac{1}{K}+\sqrt{K}J(\hat{\bm{p}}_{n,k})}\le \delta_{1} \}$ happens.  Replace  $r$ in \eqref{class_micsgle} by $rK^\gamma\sqrt{\frac{\ln n}{n}}$. Equation \eqref{class_micsgle} shows that with the probability larger than  $1- n^{-r}$,
\begin{align}
  & \frac{1}{n}\sum_{i=1}^{n}\bm{Y}_i^T\ln\pr{\frac{\hat{\bm{p}}_{n,k}(\bm{X}_i)}{2\bm{\tilde{p}}_k(\bm{X}_i)} + \frac{1}{2}}  - \E\left[ \bm{Y}_i^T\ln\pr{\frac{\hat{\bm{p}}_{n,k}(\bm{X}_i)}{2\bm{\tilde{p}}_k(\bm{X}_i)} + \frac{1}{2}}\right]\nonumber \\
  & \le  T(\delta_1) + 4r\sqrt{\frac{\ln n}{n}}\nonumber\\
  & \le  \frac{4\sqrt{2}K^2\sqrt{L_k}}{{n}} + 4rK^\gamma\sqrt{\frac{\ln n}{n}}, \label{l[l[]}
\end{align}
where in the last line \eqref{yykdhkh2} is used. 

In conclusion, the combination of \eqref{NHhno} and \eqref{l[l[]} shows that with the probability larger than $1-(B_n+1)n^{-r}$,
\begin{align}
  & \frac{1}{n}\sum_{i=1}^{n}\bm{Y}_i^T\ln\pr{\frac{\hat{\bm{p}}_{n,k}(\bm{X}_i)}{2\bm{\tilde{p}}_k(\bm{X}_i)} + \frac{1}{2}}  - \E\left[ \bm{Y}_i^T\ln\pr{\frac{\hat{\bm{p}}_{n,k}(\bm{X}_i)}{2\bm{\tilde{p}}_k(\bm{X}_i)} + \frac{1}{2}}\right]\nonumber\\
 & \le   4\sqrt{2}\frac{K^{\frac{3}{2}}}{\sqrt{n/L_k}}+2\sqrt{2}\frac{K^2}{\sqrt{n/L_k}}J(\hat{\bm{p}}_{n,k})  + 4rK^\gamma\sqrt{\frac{\ln n}{n}}. \label{NHh234no}
\end{align}

Substitute \eqref{l[l[]} into \eqref{NHh234no} and set $\lambda_n=\frac{4\sqrt{2}K^2}{\sqrt{n/L_k}}$. Then, it holds

\begin{align}
  & 4\sqrt{2}\frac{K^{\frac{3}{2}}}{\sqrt{n/L_k}}+ 4rK^\gamma\sqrt{\frac{\ln n}{n}}+  \frac{\lambda_n}{4}J(\tilde{\bm{p}}_k)+ \delta_{opt}^2\nonumber\\
  &\ge R\pr{\frac{\hat{\bm{p}}_{n,k}+\tilde{\bm{p}}_k}{2}, \tilde{\bm{p}}_k} +\frac{\lambda_n}{2} J(\hat{\bm{p}}_{n,k})-2(1+c_0)\sqrt{R\pr{\frac{\hat{\bm{p}}_{n,k}+\tilde{\bm{p}}_k}{2}, \tilde{\bm{p}}_k}R(\tilde{\bm{p}}_k,\bm{\eta})}\nonumber\\
  &\ge  R\pr{\frac{\hat{\bm{p}}_{n,k}+\tilde{\bm{p}}_k}{2}, \tilde{\bm{p}}_k} -2(1+c_0)\sqrt{R\pr{\frac{\hat{\bm{p}}_{n,k}+\tilde{\bm{p}}_k}{2}, \tilde{\bm{p}}_k}R(\tilde{\bm{p}}_k,\bm{\eta})}\label{YYX/}
\end{align}
with the probability larger than $1-(B_n+1)n^{-r}$. For any $v^2-va\le b$ with $a,b>0$, we have $v^2\le 2a^2+ 8b$. With this result and \eqref{YYX/}, 
\begin{equation}\label{JJhfdksh!}
  R\pr{\frac{\hat{\bm{p}}_{n,k}+\tilde{\bm{p}}_k}{2}, \tilde{\bm{p}}_k}\lesssim R(\tilde{\bm{p}}_k,\bm{\eta})+ \frac{K^{\frac{3}{2}}}{\sqrt{n/L_k}} + K^\gamma\sqrt{\frac{\ln n}{n}}+ {\lambda_n}J(\tilde{\bm{p}}_k)+ \delta_{opt}^2.
\end{equation}

At this step, we need introduce a lemma to deal with terms $ R\pr{\frac{\hat{\bm{p}}_{n,k}+\tilde{\bm{p}}_k}{2}, \tilde{\bm{p}}_k}$ and $ R(\tilde{\bm{p}}_k,\bm{\eta})$. 

\begin{lemma}\label{lem:excess risk of convex combination}
  For all conditional class probabilities \(\bm{p}\in \CN_k\) and \( \bm{q}\), we have
  \begin{equation}\label{KBNKN}
        R(\bm{p}, \bm{q}) \leq 16R\left(\frac{\bm{p} + \bm{q}}{2}, \bm{q}\right)\quad \text{and}\quad R(p,\eta)\le \inf_{c\in\R}\frac{1}{2} \sum_{j=1}^{K} \left(\sum_{j=1}^{K}\|p^{last}_j-\ln \eta_j-c\|_\infty^2\right)^{\frac{1}{2}}
   \end{equation}
   
\end{lemma}
\begin{proof}
    Consider the first part. Recall $p=(p_1,\ldots,p_K)$ and $q= (p_1,\ldots,p_K)$ are p.d.f.s. Note that
    \begin{align*}
        \left\vert\sqrt{p_k(\bm{x})} - \sqrt{q_k(\bm{x})}\right\vert 
        &= \frac{\left\vert p_k(\bm{x}) - q_k(\bm{x})\right\vert}{\sqrt{p_k(\bm{x})} + \sqrt{q_k(\bm{x})}} \\
        &= 2\frac{\sqrt{\frac{p_k(\bm{x}) + q_k(\bm{x})}{2}} 
        + \sqrt{q_k(\bm{x})}}{\sqrt{p_k(\bm{x})} + \sqrt{q_k(\bm{x})}} \left\vert\sqrt{\frac{p_k(\bm{x}) + q_k(\bm{x})}{2}} - \sqrt{q_k(\bm{x})}\right\vert \\
        &= 2\frac{\sqrt{\frac{p_k(\bm{x}) }{2}} +\sqrt{\frac{p_k(\bm{x}) }{2}} 
        + \sqrt{q_k(\bm{x})}}{\sqrt{p_k(\bm{x})} + \sqrt{q_k(\bm{x})}} \left\vert\sqrt{\frac{p_k(\bm{x}) + q_k(\bm{x})}{2}} - \sqrt{q_k(\bm{x})}\right\vert\\
        &\leq 4\left\vert\sqrt{\frac{p_k(\bm{x}) + q_k(\bm{x})}{2}} - \sqrt{q_k(\bm{x})}\right\vert.
    \end{align*}
    This implies
    \begin{align*}
        H^2(\bm{p}, \bm{q}) &= \frac{1}{2}\sum_{k=1}^K \left\vert\sqrt{p_k(\bm{x})} - \sqrt{q_k(\bm{x})}\right\vert^2 
        \leq 16 \cdot \frac{1}{2}\sum_{k=1}^K \left\vert\sqrt{\frac{p_k(\bm{x}) + q_k(\bm{x})}{2}} - \sqrt{q_k(\bm{x})}\right\vert^2 
        = 16H^2\left(\frac{\bm{p} + \bm{q}}{2}, \bm{q}\right).
    \end{align*}
    Thus, \(R\) satisfies the first part of \eqref{KBNKN} by definition.
    
     Consider the second part. Write $\eta= (\eta_1,\ldots,\eta_K)$. For any $j\in [K]$, it is known 
     $$
       \left\vert\sqrt{p_k(\bm{x})} - \sqrt{\eta_k(\bm{x})}\right\vert^2 \le \left| p_k(\bm{x})-\eta_k(\bm{x}) \right|. 
     $$
     Since Lemma \ref{lemma_neural network representation} holds and $\|\nabla G\|_2\le 1$ where $G$ is defined in \eqref{MJks'}, 
     \begin{equation}\label{Uildasl}
        \left| p_k(\bm{x})-\eta_k(\bm{x}) \right|\le \left(\sum_{j=1}^{K}\|p^{last}_j-\ln \eta_j-c\|_\infty^2\right)^{\frac{1}{2}}.
     \end{equation}
     The combination of above two inequalities completes the proof.
\end{proof}

Finally, the combination of \eqref{KBNKN} and \eqref{JJhfdksh!} completes the proof of Theorem \ref{theorem_consistency_class}.  
\end{proof}

\noindent\textit{Proof of Theorem \ref{theorem_consistency_class}.} The proof is established on Theorem \ref{theorem_consistency_class22}, from which
  \begin{align*}
    &R(\boldsymbol{\eta}(\bm{X}), \hat{\bm{p}}_{n,k}(\bm{X}))\\
     &  \lesssim  \underbrace{\inf_{c\in\R} \sum_{j=1}^{K} \left(\sum_{j=1}^{K}\|\tilde{p}_{k,j}^{last}-\ln \eta_j-c\|_\infty^2\right)^{\frac{1}{2}} }_{approximation\ error}+ \underbrace{{\frac{K^{\frac{3}{2}\vee \gamma}}{\sqrt{n/\ln n}}} +  \lambda_n J(\tilde{p}_k)}_{sample\ error}+ \underbrace{ \delta_{opt}^2}_{optimization\ error}.
  \end{align*}
  For each $j\in [K]$, let $\tilde{p}_{k,j}^{last}\in \mathcal{N} \mathcal{N}_{d, 1}\left(W_k, L_k, U_n\right)$ be the network given in Theorem \ref{reg_multi_approx} satisfying 
 $$
    \|m-m_k^*\|_\infty \lesssim U_n^{-\frac{\gamma^*}{l}}=U_n^{-\beta_1}.
 $$
 According to Proposition 3.4 in \cite{fan2024noise}, we further have
 $$
     \|m-m_k^*\|_\infty \lesssim (L_kW_k)^{-\alpha_1}.
 $$
 Since $J^C(\tilde{p}_k)\le \max_{j\in [k]} J(\tilde{p}_{k,j}^{last})\le U_n$, it holds
    \begin{align*}
    &R(\boldsymbol{\eta}(\bm{X}), \hat{\bm{p}}_{n,k}(\bm{X}))\\
     &  \lesssim  K^{\frac{3}{2}}\max\{(L_kW_k)^{-\alpha_1}, U_n^{-\beta_1}\}  + \frac{K^{\frac{3}{2}\vee \gamma}}{\sqrt{n}} +  \frac{K^2}{\sqrt{n}} U_n\ln n+  \delta_{opt}^2.
  \end{align*}
  Taking the optimal $U_n=\left(\frac{n}{K} \right)^{\frac{1}{\beta_1+2}}$, then
$$
R(\boldsymbol{\eta}(\bm{X}), \hat{\bm{p}}_{n,k}(\bm{X}))\lesssim K^{\frac{3}{2}}\max\left\{(L_kW_k)^{-\alpha_1}, \left(\frac{n}{K} \right)^{-\frac{\beta_1}{\beta_1+2}}\ln n\right\}  + \frac{K^{\frac{3}{2}\vee \gamma}}{\sqrt{n}} +    \delta_{opt}^2.
$$
This completes the proof.   \hfill\(\Box\)\\

\bibliographystyle{chicago}

\end{document}